\pdfoutput=1
\documentclass{article}
\usepackage[final,nonatbib]{neurips_2022}

\usepackage[utf8]{inputenc} %
\usepackage[T1]{fontenc}    %
\usepackage{url}            %
\usepackage{booktabs}       %
\usepackage{amsfonts}       %
\usepackage{nicefrac}       %
\usepackage{microtype}      %
\usepackage{xcolor}         %
    
\usepackage{amsmath}
\usepackage{amssymb}
\usepackage{graphicx}
\usepackage{makecell}
\usepackage{cleveref}
\usepackage{tabularx}
\usepackage[justification=centering,skip=0pt]{subcaption}
\newcolumntype{C}[1]{>{\centering\arraybackslash}p{#1}}
\newcolumntype{L}[1]{>{\raggedright\arraybackslash}p{#1}}
\newcolumntype{Y}{>{\centering\arraybackslash}X}
\usepackage{array}
\usepackage{float}
\usepackage{array,multirow,graphicx}
\graphicspath{ {./images/} }

\usepackage{gensymb}
\usepackage{rotating}

\usepackage{enumerate}

\newcommand{\inner}[1]{\left\langle#1\right\rangle}

\newcommand{\norm}[1]{\left\|#1\right\|}

\usepackage{xspace}
\usepackage[normalem]{ulem}

\newcommand{\madryft}{MNR-RN50\xspace}
\newcommand{\xcitft}{MNR-XCiT\xspace}
\newcommand{\deitft}{MNR-DeiT\xspace}

\newcommand{\mh}[1]{{\color{black}{#1}}}
\newcommand{\mx}[1]{{\color{black}{#1}}}
\newcommand{\vb}[1]{{\color{black}{#1}}}
\newcommand{\vbcancel}[1]{}

\makeatletter
\newcommand{\printfnsymbol}[1]{%
  \textsuperscript{\@fnsymbol{#1}}%
}
\makeatother

\title{Diffusion Visual Counterfactual Explanations}

\author{%
  Maximilian Augustin\thanks{Equal contribution.}\quad\;
  Valentyn Boreiko\printfnsymbol{1}\quad\; Francesco Croce \quad\; Matthias Hein\\
  University of Tübingen\\
}

\begin{document}

\maketitle

\begin{abstract}
Visual Counterfactual Explanations (VCEs) are an important tool to understand the decisions of an image classifier. They are “small” but “realistic” semantic changes of the image changing the 
classifier decision. Current approaches for the generation of VCEs are restricted to adversarially robust models and often contain non-realistic artefacts, or are limited
to image classification 
problems with few classes. In this paper, we overcome this by generating Diffusion Visual 
Counterfactual Explanations (DVCEs) for arbitrary ImageNet classifiers via a diffusion process. Two modifications to the diffusion process are key for our DVCEs: first, an adaptive parameterization, whose hyperparameters generalize across images and models, together with distance regularization and late start of the diffusion process, allow us to generate images with minimal semantic changes to the original ones %
but different classification.   %
Second, our cone regularization via an adversarially robust model ensures that the diffusion process does not converge to trivial non-semantic changes, but instead produces realistic images of the target class which achieve high confidence by the classifier.
Code is available under \url{https://github.com/valentyn1boreiko/DVCEs}.
\end{abstract}

\setlength\tabcolsep{1pt}
    \section{Introduction}
    It can be argued that one of the main problems hindering the widespread use of machine learning and image classification in particular, is the missing possibility to explain the decisions of black-box models such as neural networks. This is not only a problem for decisions affecting humans where the current draft for AI regulation in Europe \cite{EU2021} requires ``transparency'', but it is a pressing problem in all applications of machine learning. The reason, to some extent, is that humans would like to understand but also control if the learning algorithm has captured the ``concepts'' of the underlying classes or if it just predicts well using spurious features, artefacts in the data set, or other sources of error. In this paper, we focus on model-agnostic explanations which can, in principle, be applied to any image classifier and do not rely on the specific structure of the classifier such as decision trees or linear classifiers. In this area, in particular for image classification, sensitivity based explanations \cite{BaeEtAl2010}, explanations based on feature attributions \cite{BacEtAl2015}, saliency maps \cite{simonyan2014deep, Selvaraju_2019, etmann2019connection, wang2020smoothed, SrinivasFleuret2019}, Shapley additive explanations \cite{SHAPley_explanations}, and local fits of interpretable models \cite{Ribeiro-Lime} have been proposed. Moreover, \cite{wachter2018counterfactual} proposed counterfactual explanations (CEs), which are instance-specific explanations. They can be applied to any  classifier  and it has been argued that they are close to the human justification of decisions \cite{Mil2017} using
    counterfactual reasoning: ``I would recognize it as zebra (instead of horse) if it had
    black and white stripes.''  For a given classifier almost all methods for the generation of CEs \cite{wachter2018counterfactual, dhurandhar2018explanations, Mothilal2020counterfactual, Barocas2020, Pawlowski2020counterfactual, verma2020counterfactual, schut2021generating} try to solve the following problem:  ``Given a target class $c$ what is the minimal change $\delta$ of input $x$, such that $x+\delta$ is classified as class $c$ with high probability and is a realistic instance of my data generating distribution?'' 
    From the perspective of ``debugging'' existing machine learning models, CEs are interesting as they construct a concrete input $x+\delta$ with a different classification that allows the developer to test if the model has learned the correct features.

    The main reason why visual counterfactual explanations (VCEs), that is CEs for image classification,  are not widely used is that the tasks of generating CEs and adversarial examples \cite{SzeEtAl2014} %
    are very related. Even imperceivable changes of the image can already change the prediction of the classifier, however, the resulting noise patterns do not show the user if the classifier has picked up the right class-specific features. One possible solution is using adversarially robust models \cite{santurkar2019image,augustin2020,boreiko2022sparse}, which have been shown to produce semantically meaningful VCEs by directly maximizing the probability of the target class in image space. These approaches have the downside that they only can generate VCEs for robust models which is a significant restriction as these models are not competitive in terms of prediction accuracy. 
    The second approach is to restrict the generation of VCEs using a generative model or constraining the set of potential image manipulations \cite{HenEtAl2016,HenEtAl2018,SamEtAl2018%
    , chang2018explaining,GoyEtAl2019,schutte2021using,liu2021more}. However, these approaches are either restricted to datasets with a small number of classes, cannot provide explanations for arbitrary classifiers, or generate VCEs that look realistic but have so little in common with the original image that not much insight can be gained.
    
    Recently, \cite{explaining_in_style} trained a StyleGAN2 \cite{Karras2019stylegan2} model to discover and manipulate class attributes. While their approach yields impressive results on smaller-scale datasets with few similar classes (for example, different bird species), the authors did not demonstrate that the method scales to complex tasks such as ImageNet with hundreds of classes where different classes require different sets of attributes. Another disadvantage is that the StyleGAN model needs to be retrained for every classifier, making it prohibitively expensive to explain multiple large models. Moreover, \cite{khorram2022cycle} have proposed a loss to do VCEs in the latent space of a GAN/VAE. They show promising results for MNIST/FMNIST, but no code is available for ImageNet. %
    Furthermore, to explain a classifier, the conditional information during the generation of explanations should ideally come only from the classifier itself, %
    but \cite{khorram2022cycle} rely on a conditional GAN which might introduce a bias. %

\begin{figure}[t]
     \centering
     \small
     \begin{tabular}{c|cc||c|cc}
     \hline
     Original &
     Class 1 &
     Class 2 &
     Original &
     Class 1 &
     Class 2\\
     \hline

      \begin{subfigure}{0.155\textwidth}\centering
     \caption*{\scriptsize \makecell{chimpanzee\\ 0.88}}
     \includegraphics[width=1\textwidth]{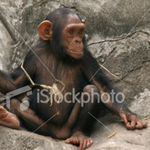}
     \end{subfigure} &

     \begin{subfigure}{0.155\textwidth}\centering
     \caption*{\scriptsize \makecell{orangutan:\\0.99}}
     \includegraphics[width=1\textwidth]{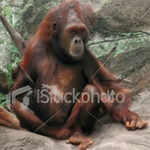}
     \end{subfigure} &
     \begin{subfigure}{0.155\textwidth}\centering
     \caption*{\scriptsize \makecell{gorilla:\\0.98}}
     \includegraphics[width=1\textwidth]{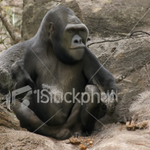}
     \end{subfigure} &

     \begin{subfigure}{0.155\textwidth}\centering
     \caption*{\scriptsize \makecell{Chesapeake Bay\\ret.: 0.97}}
     \includegraphics[width=1\textwidth]{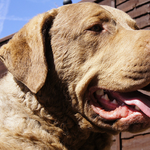}
     \end{subfigure} &
     \begin{subfigure}{0.155\textwidth}\centering
     \caption*{\scriptsize \makecell{golden retriever:\\0.99}}
     \includegraphics[width=1\textwidth]{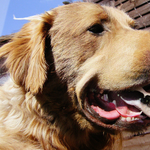}
     \end{subfigure} &
      \begin{subfigure}{0.155\textwidth}\centering
     \caption*{\scriptsize \makecell{labrador retriever:\\0.92}}
     \includegraphics[width=1\textwidth]{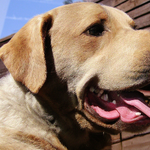}
     \end{subfigure}\\
     \end{tabular}

      \caption{\label{fig:teaser}Our method DVCE creates visual counterfactual explanations (VCEs) that explain any given classifier (here: non-robust ConvNeXt) by changing an image of one class into a similar target class by changing class relevant features while preserving the overall image structure.}
     \end{figure}

    In this paper, we overcome the aforementioned challenges 
     and generate our Diffusion Visual Counterfactual Explanations (DVCEs) for arbitrary ImageNet classifiers (see Fig \ref{fig:teaser}). We use the progress in the generation of realistic images using diffusion processes \cite{sohl2015deep,ho2020denoising,song2020denoising,song2021scorebased},  which recently were able to outperform GANs \cite{dhariwal2021diffusion}. Similar to \cite{nichol2021glide,avrahami2021blended}, we use a classifier and a distance-type regularization to guide the generation of the images. 
 Two modifications to the diffusion process are key elements for our DVCEs: i) a combination of distance regularization and starting point of the diffusion process together with an adaptive reparameterization lets us generate VCEs visually close to the original image in a controlled way so that hyperparameters can be fixed across images and even across models, ii) our cone regularization of the gradient of the classifier via an adversarially robust model ensures that the diffusion process does not converge to trivial non-semantic changes but instead produces realistic images of the target class which achieve high confidence by the classifier. Our approach can be employed for any dataset where a generative diffusion model and an adversarially robust model are available. In a qualitative comparison, user study, and a quantitative evaluation (Sec. \ref{subsec:method_comp}),
 we show that our DVCEs achieve higher realism (according to FID) and have more meaningful features (according to the user study) than both the recent methods of \cite{boreiko2022sparse} and \cite{avrahami2021blended}.%
    
\section{Diffusion models}
Diffusion models \cite{dhariwal2021diffusion,sohl2015deep,song2020denoising,ho2020denoising,nichol2021improved} are generative models that consist of two steps: a \textit{forward diffusion} process (and thus a Markov process) that transforms the data distribution to a prior distribution (which is usually assumed to be a standard normal distribution) and the \textit{reverse diffusion} process that transforms the prior distribution back to the data distribution. 
The existence of the (continuous-time) reverse diffusion process was closer investigated in \cite{song2021scorebased}.\\
In the discrete-time setting, a Markov chain $\{x_1,...,x_T\}$ for any data point $x_0$ in the forward direction is defined via a  Markov process that adds noise to the data point at each timestep:
\begin{equation}\label{eq:forward_diff}
    q(x_t|x_{t-1}) = \mathcal{N}(\sqrt{1-\beta_t}x_{t-1}, \beta_t I),
\end{equation}
where $\{\beta_1,...,\beta_T\}$ is some variance schedule, chosen such that $q(x_T) \sim \mathcal{N}(0, I)$.
Note that given $x_0$, it is possible to sample from $q(x_t|x_0)$ in closed form instead of applying noise $t$ times using:

\begin{align}
    q(x_t|x_0) &= \mathcal{N}( \sqrt{\overline{\alpha}_t} x_0, 1 - \overline{\alpha}_t I), \\
    x_t &= \sqrt{\overline{\alpha}_t} x_0 + \epsilon \sqrt{1 - \overline{\alpha}_t},\quad \epsilon \sim \mathcal{N}(0, I), \label{eq:forward_t}
\end{align}

where $\overline{\alpha}_t := \prod_{k=1}^t (1-\beta_k)$.
In  \cite{sohl2015deep}, the authors have shown that the reverse transitions $q(x_{t-1}|x_t)$ approach diagonal Gaussian distributions as $T \rightarrow \infty$ and thus 
one can use a DNN with parameters $\theta$ to approximate $q(x_{t-1}|x_t)$ as $p_{\theta}(x_{t-1}|x_t)$ by predicting the mean and %
the diagonal covariance matrix $\mu_{\theta}(x_t,t)$ and $\Sigma_{\theta}(x_t,t)$:
\begin{equation}\label{eq:reverse_transitions}
    p_{\theta}(x_{t-1}|x_t) = \mathcal{N}(\mu_{\theta}(x_t,t), \Sigma_{\theta}(x_t,t)).
\end{equation}
To sample from $p_\theta(x_0)$, one samples $x_T$ from $\mathcal{N}(0, I)$ and then follows the reverse process by repeatedly sampling from the transition probabilities $p_{\theta}(x_{t-1}|x_t)$. 

Instead of predicting $\mu_{\theta}(x_t,t)$ directly, it has been shown in \cite{ho2020denoising} that the best performing parameterization uses a neural network $\epsilon_{\theta}(x_t, t)$ to approximate the source noise $\epsilon$ in \eqref{eq:forward_t} and the loss used for training resembles that of a denoising model:
\begin{equation}
    L_\mathrm{simple}(\theta, T, q(x_0)) := \mathbb{E}_{t \sim [1, T], x_0 \sim q(x_0), \epsilon \sim \mathcal{N}(0, I)} \norm{\epsilon -\epsilon_{\theta}(x_t, t)}^2.
\end{equation}
The mean of the reverse step $\mu_{\theta}$ in \eqref{eq:reverse_transitions} can be derived \vbcancel{by plugging \eqref{eq:forward_t} in $\tilde{\mu}_t(x_t, x_0)$}\vb{\cite{luo_2022_understanding_diffusion} using Bayes theorem} as:
\begin{equation}
    \mu_{\theta}(x_t, t)=\frac{1}{\sqrt{1-\beta_t}}\Big(x_t - \frac{\beta_t}{\sqrt{1-\overline{\alpha}_t}} \epsilon_{\theta}(x_t, t)\Big).
\end{equation}

 This is, however, only one of the possible equivalent parameterization of the learning objective as shown in \cite{luo_2022_understanding_diffusion}.
Because the objective $L_\mathrm{simple}$ does not give a learning signal for $\Sigma_{\theta}(x_t,t)$, in practice one combines $L_\mathrm{simple}$ with another loss based on a variational lower bound of the data likelihood \cite{sohl2015deep}, which unlike $L_\mathrm{simple}$ allows us to learn the diagonal covariance matrix $\Sigma_{\theta}(x_t,t)$. Concretely, the network $\epsilon_\theta(x_t, t)$ outputs additionally a vector that is used to parametrize $\Sigma_{\theta}(x_t,t)$.

\subsection{Class conditional sampling}
There exist diffusion models trained with and without knowledge about classes in the dataset \cite{dhariwal2021diffusion,ho2021classifierfree}. 
For our experiments, we only use the class-unconditional diffusion model, such that all class-conditional features are introduced by the classifier that we want to explain. 
For the noise-aware classifier $p_\phi(y|\cdot, \cdot) : \mathbb{R}^d \times \{1, ..., T\} \rightarrow [0, 1]$, with parameters $\phi$, that is trained on noisy images corresponding to the various timesteps \cite{dhariwal2021diffusion}, the reverse process transitions are then of the form: 
\begin{equation}\label{eq:cond_reverse}
    p_{\theta, \phi}(x_{t-1}|x_{t}, y) = Z\,  p_{\theta}(x_{t-1}|x_{t})\, p_\phi(y|x_t, t)
\end{equation}
for a normalization constant $Z$. 
As we would like to explain any classifier and not only noise-aware ones, we follow the approach of \cite{avrahami2021blended}, where a classifier $p_\phi(y|\cdot) : \mathbb{R}^d \rightarrow [0, 1]$ is given as input the denoised sample $\hat{x}_0=f_\mathrm{dn}(x_t)$ of $x_t$, using the mapping:
\begin{equation}\label{eq:denoise_sample}
    f_\mathrm{dn} : \mathbb{R}^d \times \{1, ..., T\} \rightarrow\mathbb{R}^d, \;\;\; (x_t, t) \mapsto \frac{x_t}{\sqrt{\overline{\alpha}_t}} - \frac{\sqrt{1-\overline{\alpha}_t}\epsilon_{\theta}(x_t, t)}{\sqrt{\overline{\alpha}_t}}.
\end{equation}

This mapping, derived from \eqref{eq:forward_t} estimates the noise-free image $x_0$ using the  noise approximated by the model $\epsilon_{\theta}(x_t, t)$ for a given timestep $t$. With this, we can define a timestep-aware posterior $p_\phi(y|x_t, t)$ for any classifier $p_\phi(y|\cdot)$ as 
    $p_\phi(y|x_t, t) := p_\phi(y|f_\mathrm{dn}(x_t, t))$.

The issue is that efficient sampling from the original diffusion model is possible only because the reverse process is made of normal distributions. As we need to sample from $p_{\theta, \phi}(x_{t-1}|x_{t}, y)$ hundreds of times to obtain a single sample from the data distribution, it is not possible to use MCMC-samplers with high complexity to sample from each of the individual transitions. %
In \cite{dhariwal2021diffusion}, they proposed to solve it by approximating $p_{\theta, \phi}(x_{t-1}|x_{t}, y)$ with slightly shifted versions of $p_{\theta}(x_{t-1}|x_{t})$ to make closed-form sampling possible. Such transition kernels are given by:
\begin{align}
    p_{\theta, \phi}(x_{t-1}|x_{t}, y) &= \mathcal{N}(\mu_t, \Sigma_{\theta}(x_{t},t)), \label{eq:transition_kernel} \\
    \mu_t &= \mu_{\theta}(x_{t},t) + \Sigma_{\theta}(x_{t},t) \nabla_{x_t} \log p_{\phi}(y|x_t, t), \label{eq:mean_usual}
\end{align}
which we further adapt for the goal of generating VCEs and use in our experiments.

\section{Diffusion Visual Counterfactual Explanations}\label{sec:DVCEs_properties}


A VCE \vbcancel{$z$}\vb{$x$} for a chosen target class \vbcancel{$c$}\vb{$y$}, a given classifier \vbcancel{$f$}\vb{$p_\phi(y|\cdot)$}, and an input \vbcancel{$x$}\vb{$\hat{x}$} should satisfy the following criteria:
i) \textbf{validity:} the VCE \vbcancel{$z$}\vb{$x$} should be classified by \vbcancel{$f$}\vb{$p_\phi(y|\cdot)$} as the desired target class \vbcancel{$c$}\vb{$y$} with high predicted probability, ii) \textbf{realism:} the VCE should \mh{be as close as possible} to a natural image, iii) \textbf{minimality/closeness:} the difference between the VCE \vbcancel{$z$}\vb{$x$} and the original image \vbcancel{$x$}\vb{$\hat{x}$} should be the minimal semantic modification necessary to change the class, in particular, the generated image \vbcancel{$z$}\vb{$x$} should be close to \vbcancel{$x$}\vb{$\hat{x}$} while being valid and realistic, e.g. by changing the object in the image and leaving the background unchanged.
Note that targeted adversarial examples are valid but do not show meaningful semantic changes in the target class for a non-robust model and are not realistic. 

\mh{The $l_{1.5}$-SVCEs of \cite{boreiko2022sparse} \mh{change the image in order to maximize the predicted probability of the classifier into the target class inside a\vb{n} $l_{1.5}$-ball around the image which is a targeted adversarial example}. Thus this only works for robust classifiers and they use an ImageNet classifier that was trained to be multiple-norm robust (MNR), which we denote
in this paper  as \madryft (see Sec. \ref{subsec:model_comp}). The realism of the $l_{1.5}$-SVCEs comes purely from the generative properties of robust classifiers \cite{santurkar2019image}\vb{,} which can lead to artefacts. In contrast, our Diffusion Visual Counterfactual Explanations (DVCEs) work for any classifier and our DVCEs are more realistic due the better generative properties of diffusion models.
An approach similar to our DVCE framework is \mh{Blended Diffusion} (BD) \cite{avrahami2021blended} which manipulates the image inside a masked region. \mh{One can} adapt BD  for the \mh{generation} of VCEs by using as mask the whole image. DVCE and \mh{BD} share the same diffusion model, but BD
cannot be applied to arbitrary classifiers and requires image-specific hyperparameter tuning, see Fig. \ref{fig:method_ours_blended}.}
\subsection{Adaptive Parameterization}

\begin{figure}[t]
     \centering
     \small
     \begin{tabular}{c|c|ccc|ccc}
     \hline
     \multicolumn{1}{c|}{Original}&
     \multicolumn{1}{c|}{DVCEs}&
     \multicolumn{3}{c|}{BDVCEs $C_c$ 10}&
     \multicolumn{3}{c}{BDVCEs $C_c$  25}\\
     
     & & $C_d$  = 100 & 500 & 1000 & 100 & 500 & 1000\\
     \hline

      \begin{subfigure}{0.118\textwidth}\centering
     \caption*{\scriptsize \makecell{keeshond:\\ 0.38}}
     \includegraphics[width=1\textwidth]{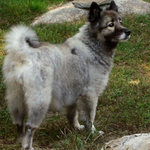}
     \end{subfigure} &

     \begin{subfigure}{0.118\textwidth}\centering
     \caption*{\scriptsize \makecell{ chow:\\ 1.00}}
     \includegraphics[width=1\textwidth]{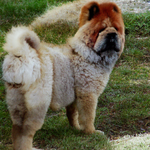}
     \end{subfigure} &
     
     \begin{subfigure}{0.118\textwidth}\centering
     \caption*{\scriptsize \makecell{ chow:\\0.77}}
     \includegraphics[width=1\textwidth]{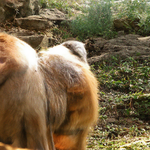}
     \end{subfigure} &
     \begin{subfigure}{0.118\textwidth}\centering
     \caption*{\scriptsize \makecell{ chow:\\0.55}}
     \includegraphics[width=1\textwidth]{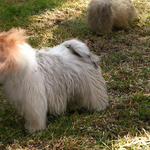}
     \end{subfigure} &
     \begin{subfigure}{0.118\textwidth}\centering
     \caption*{\scriptsize \makecell{ chow:\\0.07}}
     \includegraphics[width=1\textwidth]{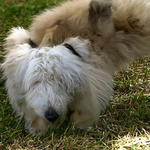}
     \end{subfigure} &
     \begin{subfigure}{0.118\textwidth}\centering
     \caption*{\scriptsize \makecell{ chow:\\0.99}}
     \includegraphics[width=1\textwidth]{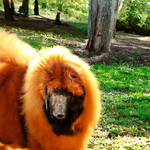}
     \end{subfigure} &
     \begin{subfigure}{0.118\textwidth}\centering
     \caption*{\scriptsize \makecell{ chow:\\0.64}}
     \includegraphics[width=1\textwidth]{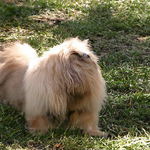}
     \end{subfigure} &
     \begin{subfigure}{0.118\textwidth}\centering
     \caption*{\scriptsize \makecell{ chow:\\0.80}}
     \includegraphics[width=1\textwidth]{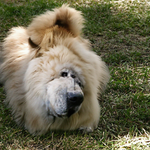}
     \end{subfigure}\\

      \begin{subfigure}{0.118\textwidth}\centering
     \caption*{\scriptsize \makecell{lynx:\\ 0.72}}
     \includegraphics[width=1\textwidth]{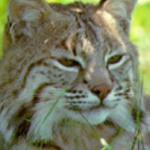}
     \end{subfigure} &

     \begin{subfigure}{0.118\textwidth}\centering
     \caption*{\scriptsize \makecell{ cheetah:\\ 0.99}}
     \includegraphics[width=1\textwidth]{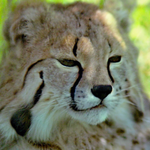}
     \end{subfigure} &
     
     \begin{subfigure}{0.118\textwidth}\centering
     \caption*{\scriptsize \makecell{ cheetah:\\ 0.67}}
     \includegraphics[width=1\textwidth]{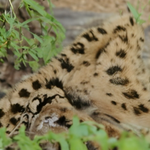}
     \end{subfigure} &
     \begin{subfigure}{0.118\textwidth}\centering
     \caption*{\scriptsize \makecell{ cheetah:\\ 0.38}}
     \includegraphics[width=1\textwidth]{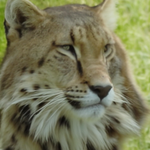}
     \end{subfigure} &
     \begin{subfigure}{0.118\textwidth}\centering
     \caption*{\scriptsize \makecell{ cheetah:\\ 0.77}}
     \includegraphics[width=1\textwidth]{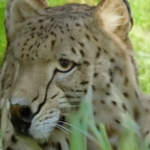}
     \end{subfigure} &
     \begin{subfigure}{0.118\textwidth}\centering
     \caption*{\scriptsize \makecell{ cheetah:\\0.95 }}
     \includegraphics[width=1\textwidth]{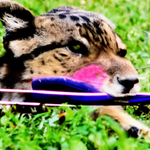}
     \end{subfigure} &
     \begin{subfigure}{0.118\textwidth}\centering
     \caption*{\scriptsize \makecell{ cheetah:\\0.73}}
     \includegraphics[width=1\textwidth]{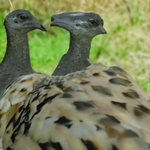}
     \end{subfigure} &
     \begin{subfigure}{0.118\textwidth}\centering
     \caption*{\scriptsize \makecell{ cheetah:\\0.98}}
     \includegraphics[width=1\textwidth]{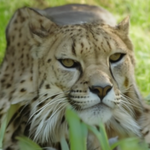}
     \end{subfigure}\\

      \begin{subfigure}{0.118\textwidth}\centering
     \caption*{\scriptsize \makecell{weasel:\\ 0.41}}
     \includegraphics[width=1\textwidth]{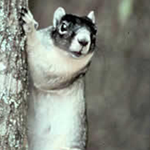}
     \end{subfigure} &

     \begin{subfigure}{0.118\textwidth}\centering
     \caption*{\scriptsize \makecell{guinea pig:\\ 1.00}}
     \includegraphics[width=1\textwidth]{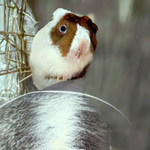}
     \end{subfigure} &
     
     \begin{subfigure}{0.118\textwidth}\centering
     \caption*{\scriptsize \makecell{guinea pig:\\0.86}}
     \includegraphics[width=1\textwidth]{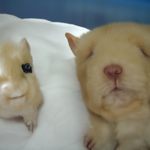}
     \end{subfigure} &
     \begin{subfigure}{0.118\textwidth}\centering
     \caption*{\scriptsize \makecell{guinea pig:\\0.12}}
     \includegraphics[width=1\textwidth]{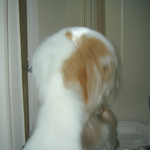}
     \end{subfigure} &
     \begin{subfigure}{0.118\textwidth}\centering
     \caption*{\scriptsize \makecell{guinea pig:\\0.05}}
     \includegraphics[width=1\textwidth]{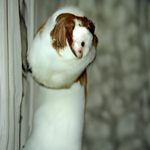}
     \end{subfigure} &
     \begin{subfigure}{0.118\textwidth}\centering
     \caption*{\scriptsize \makecell{guinea pig:\\0.93}}
     \includegraphics[width=1\textwidth]{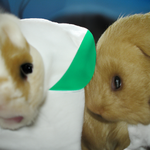}
     \end{subfigure} &
     \begin{subfigure}{0.118\textwidth}\centering
     \caption*{\scriptsize \makecell{guinea pig:\\0.97}}
     \includegraphics[width=1\textwidth]{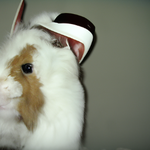}
     \end{subfigure} &
     \begin{subfigure}{0.118\textwidth}\centering
     \caption*{\scriptsize \makecell{guinea pig:\\0.70}}
     \includegraphics[width=1\textwidth]{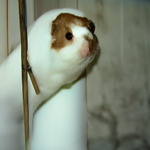}
     \end{subfigure}\\

     \end{tabular}

      \caption{\label{fig:method_ours_blended}DVCEs for the robust \madryft
      model \protect\cite{boreiko2022sparse} in the second column (ours) and using Blended Diffusion (BDVCEs) \protect\cite{avrahami2021blended}, such that regularization is applied to the whole image. Due to our adaptive parameterization, the same parameters ($C_c=0.1, C_d=0.15$) work for our DVCEs  across different images and classifiers (see Fig. \ref{fig:classifier_comp}). For BDVCEs, it is difficult to choose a set of hyperparameters that, even for a single classifier, work for multiple images. For %
      BDVCEs, we report the parameter $C_d$ of the LPIPS weight and the $l_2$-weight is 10 times as large (ratio used in \protect\cite{avrahami2021blended}).} 
     \end{figure}
For \mh{generating} the DVCE of the original image $\hat{x}$, we \mh{do not} want to sample just any image $x$ from 
$p(x|y)$ (high realism and validity)\vb{,} but also to make sure that $d(x, \hat{x})$ is small. 
\vbcancel{To do this}\vb{For this}, we have to condition our diffusion process on $\hat{x}$. Using the \vbcancel{introduced} denoising function $f_\mathrm{dn}$ and analogously to the derivation of \eqref{eq:transition_kernel} and its applications in \cite{dhariwal2021diffusion, liu2021more}, the mean of the transition kernel becomes:

\begin{equation}
\label{eq:new_mean}
\mu_{\theta}(x_{t},t) + \Sigma_{\theta}(x_{t},t) \nabla_{x_t} \Big[C_c \log p_{\phi}\big(y|f_\mathrm{dn}(x_{t}, t)\big) -  C_d \,d \big(\hat{x}, f_\mathrm{dn}(x_{t}, t)\big)\Big],
\end{equation}

where $C_c$ is the coefficient of the classifier, and $C_d$ the one of the distance guiding loss. Intuitively, we take a step in the direction that increases the classifier score while staying close to $\hat{x}$. Note that the last term can be interpreted as the log gradient of a distribution with density $\exp( - C_d\,d_t(\hat{x}, \cdot))$, where 
$d_t(\hat{x}, \cdot) \vb{:=} d(\hat{x}, f_\mathrm{dn}(\cdot, t))$. Thus we introduce a time\vb{step}-aware prior distribution that enforces our output to be similar to $\hat{x}$. In our work, we use the $l_1$-distance as it produces \mh{sparse} changes.
Several works have tried to minimize some distance during the diffusion process, implicitly in \cite{choi2021ilvr} or explicitly, but for the background, in \vbcancel{\cite{avrahami2021blended}}\vb{BD}. However, there is no principled way to choose the coefficient for such a regularization term. It turns out that it is impossible to find a parameter setting for \vbcancel{\mh{Blended Diffusion} \cite{avrahami2021blended}}\vb{BD} that works across images and classifiers. Thus, we propose \mh{a parameterization}  
\begin{equation}\label{eq:g_update}
    g_\mathrm{update} =  C_c \frac{ \nabla_{x_t} \log p_\phi(y|f_\mathrm{dn}(x_t, t)))}{\norm{\nabla_{x_t} \log p_\phi(y|f_\mathrm{dn}(x_t, t)))}_2} - C_d \frac{\nabla_{x_t} d(\hat{x}, f_\mathrm{dn}(x_t, t))}{\norm{\nabla_{x_t} d(\hat{x}, f_\mathrm{dn}(x_t, t))}_2}, \\
\end{equation}
\mh{which adapts} to the predicted mean of the diffusion model that we use to change $\mu_t$ in \eqref{eq:mean_usual} to
\begin{equation}
    \label{eq:adaptive}
    \mu_{t} = \mu_{\theta}(x_{t},t) + \Sigma_{\theta}(x_{t},t) \norm{\mu_{\theta}(x_{t},t)}_2 g_\mathrm{update}.
\end{equation}
\mh{This} adaptive parameterization allows for fine-grained control of the \mh{influence} of the classifier and distance regularization so that now the hyperparameters $C_c$ and $C_d$ have the same influence across images and even classifiers. \mh{It} facilitates 
the generation of DVCEs as otherwise hyperparameter finetuning would be necessary for each image as in \vbcancel{\mh{Blended Diffusion}}\vb{BD}, see Fig. \ref{fig:method_ours_blended} for a comparison. However, even with our adaptive parameterization, it is still not easy to produce semantically meaningful changes close to $\hat{x}$ as can be seen in App. 
\ref{app:starting_T}. Thus, as in \cite{avrahami2021blended}, we vary the starting point \mh{of} the diffusion process and observe in App. 
\ref{app:starting_T}
that starting from step $\frac{T}{2}$ of the forward diffusion process, together with the adaptive parameterization and using as the distance the $l_1$-distance, provides us with sparse but semantically meaningful changes. In our experiments, we set $T=200$.

\subsection{Cone Projection for Classifier Guidance}\label{subsec:cone}
\begin{figure*}[t]
     \centering
     \small
     \begin{tabular}{c|ccc||c|ccc}
     \hline
     \multicolumn{1}{c|}{Original} &
      Non-robust &
     Robust &
     Cone Proj. &
     Original &
     Non-robust &
     Robust &
     Cone Proj.\\
     \hline
      \begin{subfigure}{0.118\textwidth}\centering
     \caption*{\scriptsize \makecell{ ladybug\\ \quad }}
     \includegraphics[width=1\textwidth]{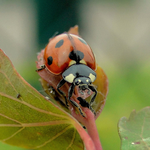}
     \end{subfigure} &
      \begin{subfigure}{0.118\textwidth}\centering
     \caption*{\scriptsize \makecell{ weevil:\\ 0.99}}
     \includegraphics[width=1\textwidth]{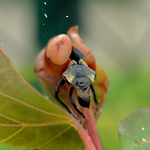}
     \end{subfigure} &
           \begin{subfigure}{0.118\textwidth}\centering
     \caption*{\scriptsize \makecell{ weevil:\\ 1.00}}
     \includegraphics[width=1\textwidth]{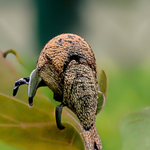}
     \end{subfigure} &
      \begin{subfigure}{0.118\textwidth}\centering
     \caption*{\scriptsize \makecell{ weevil:\\ 0.99}}
     \includegraphics[width=1\textwidth]{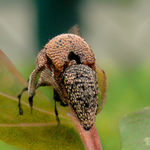}
     \end{subfigure} &
     
      \begin{subfigure}{0.118\textwidth}\centering
     \caption*{\scriptsize \makecell{ ringlet\\ \quad }}
     \includegraphics[width=1\textwidth]{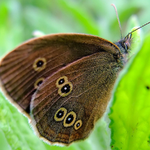}
     \end{subfigure} &
      \begin{subfigure}{0.118\textwidth}\centering
     \caption*{\scriptsize \makecell{ monarch:\\ 0.47}}
     \includegraphics[width=1\textwidth]{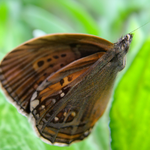}
     \end{subfigure} &
           \begin{subfigure}{0.118\textwidth}\centering
     \caption*{\scriptsize \makecell{ monarch:\\1.00 }}
     \includegraphics[width=1\textwidth]{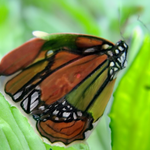}
     \end{subfigure} &
      \begin{subfigure}{0.118\textwidth}\centering
     \caption*{\scriptsize \makecell{monarch:\\0.98 }}
     \includegraphics[width=1\textwidth]{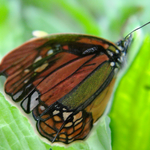}
     \end{subfigure}\\
    \end{tabular}
      \caption{\label{fig:method_cone_projection}DVCEs for the non-robust Swin-TF\protect\cite{liu2021swin}, robust \madryft model, and cone-projected DVCEs for Swin-TF\protect\cite{liu2021swin}. %
      Similar to adversarial examples, guidance by the non-robust  Swin-TF does not yield semantically meaningful changes. In contrast, the DVCEs of the robust \madryft and the DVCE with cone projection for the Swin-TF\protect\cite{liu2021swin} yield valid and realistic VCEs.
      }
     \end{figure*}

\mh{A key objective for} our DVCEs is that they can be applied to any image classifier regardless of whether it is adversarially robust or not. 
\mh{Using} diffusion with the new mean \mx{as in} \eqref{eq:adaptive} does not work \mx{with a non-robust classifier} and leads to very small modifications of the image, which are similar to adversarial examples without meaningful changes. The reason is that the gradients of non-robust classifiers are 
noisier and less semantically meaningful \mh{than those of robust classifiers}. We illustrate this for a non-robust Swin-\vbcancel{Transformer}\vb{TF} \cite{liu2021swin} in Fig.  \ref{fig:method_cone_projection} where hardly any changes are generated.
Note that this is even the case if we 
\mh{first denoise the sample} using the denoising function $f_\mathrm{dn}$ \eqref{eq:denoise_sample}.

As a solution, 
we suggest projecting the gradient of an \mh{adversarially} robust classifier with parameters \vbcancel{$\upsilon$}\vb{$\psi$}, $\nabla_{x_t} \log p_{\text{robust},\psi}(y|f_\mathrm{dn}(x_t, t))$, onto a cone centered at the gradient of the classifier, $\nabla_{x_t} \log p_{\phi}(y|f_\mathrm{dn}(x_t, t))$.  
More precisely, we define
\begin{align*}
    \label{eq:cone-proj}
    g_\mathrm{proj} &= P_{\text{cone}(\alpha, \nabla_{x_t} \log p_\phi(y|f_\mathrm{dn}(x_t, t)))} \Big[\nabla_{x_t} \log p_{\text{robust},\psi}\big(y|f_\mathrm{dn}(x_t, t)\big)\Big],
\end{align*}
where $\text{cone}(\alpha, v) := \{w \in \mathbb{R}^{d} : \angle (v, w) \leq \alpha \}$ is the cone of angle $\alpha$ around vector $v$, and the projection $P_{\text{cone}(\alpha, v)}$ onto $\text{cone}(\alpha, v)$ is given as:
\begin{equation*}
P_{\text{cone}(\alpha, v)} [w] :=\begin{cases}
  \inner{u,w}u, &\text{ $\angle (w, v) > \alpha$}\\
w, &\text{ else},
\end{cases}
\end{equation*}
 where $P_{v^\perp}(w) :=  w - \frac{\langle w, v \rangle}{\langle v, v \rangle} v$ and\vbcancel{we define} $u=\sin (\alpha)\frac{P_{v^\perp}(w)}{\norm{P_{v^\perp}(w)}_2}  + \cos(\alpha)\frac{v}{\norm{v}_2}$ (note $\norm{u}_2=1$). The \mh{motivation} for this projection is to \mx{\vbcancel{drastically }reduce} the noise in the gradient of the non-robust classifier. Note, that the projection of the gradient of the robust classifier onto the cone generated by the non-robust classifier \vb{with angle $\alpha<90\degree$} is always an ascent direction for $\log p_{\phi}(y|f_\mathrm{dn}(x_t, t))$
 \vb{,} \mh{which we would like to maximize}
 (note that $g_\mathrm{proj}$ is not necessarily an ascent direction for $\log p_{\text{robust},\psi}(y|f_\mathrm{dn}(x_t, t))$). Thus, \vbcancel{one can see }the cone projection \vbcancel{as}\vb{is} a form of regularization of $\nabla_{x_t} \log p_{\phi}(y|f_\mathrm{dn}(x_t, t))$\vb{,} which guides the diffusion process \mh{to} semantically meaningful changes \mh{of the image.} 

\begin{figure}[h!]
     \centering
     \small
     \begin{tabular}{c|ccc|ccc}
     \hline
     \multicolumn{1}{c|}{Original}&
     \multicolumn{3}{c|}{Target class 1} & 
     \multicolumn{3}{c|}{Target class 2}\\
     \hline
     & \scriptsize DVCEs (ours) &  \scriptsize $l_{1.5}$-SVCEs \cite{boreiko2022sparse}&
     \scriptsize BDVCEs \cite{avrahami2021blended} &
       \scriptsize DVCEs (ours) &  \scriptsize $l_{1.5}$-SVCEs \cite{boreiko2022sparse}&
       \scriptsize BDVCEs \cite{avrahami2021blended} \\
     \hline

      \begin{subfigure}{0.136\textwidth}\centering
     \caption*{\scriptsize \makecell{ mashed potato:\\ 0.25}}
     \includegraphics[width=1\textwidth]{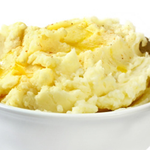}
     \end{subfigure} &

     \begin{subfigure}{0.136\textwidth}\centering
     \caption*{\scriptsize \makecell{ guacamole:\\ 1.00}}
     \includegraphics[width=1\textwidth]{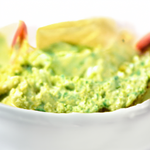}
     \end{subfigure} &
     \begin{subfigure}{0.136\textwidth}\centering
     \caption*{\scriptsize \makecell{ guacamole:\\ 0.91}}
     \includegraphics[width=1\textwidth]{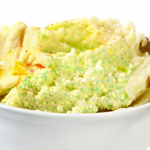}
     \end{subfigure} &
     \begin{subfigure}{0.136\textwidth}\centering
     \caption*{\scriptsize \makecell{guacamole:\\ 0.90}}
     \includegraphics[width=1\textwidth]{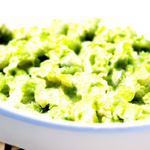}
     \end{subfigure} &

       \begin{subfigure}{0.136\textwidth}\centering
     \caption*{\scriptsize \makecell{ carbonara:\\ 1.00}}
     \includegraphics[width=1\textwidth]{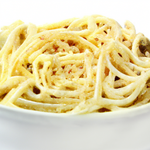}
     \end{subfigure}& 
      \begin{subfigure}{0.136\textwidth}\centering
     \caption*{\scriptsize \makecell{ carbonara:\\ 0.95}}
     \includegraphics[width=1\textwidth]{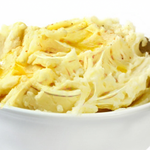}
     \end{subfigure}&
      \begin{subfigure}{0.136\textwidth}\centering
     \caption*{\scriptsize \makecell{ cabronara:\\ 0.99}}
     \includegraphics[width=1\textwidth]{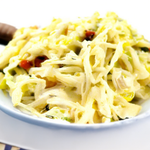}
     \end{subfigure}\\

      \begin{subfigure}{0.136\textwidth}\centering
     \caption*{\scriptsize \makecell{cheetah:\\ 1.00}}
     \includegraphics[width=1\textwidth]{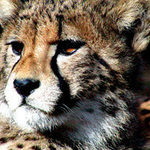}
     \end{subfigure} &

     \begin{subfigure}{0.136\textwidth}\centering
     \caption*{\scriptsize \makecell{leopard:\\ 0.96}}
     \includegraphics[width=1\textwidth]{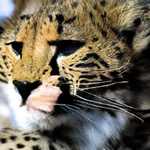}
     \end{subfigure} &
      \begin{subfigure}{0.136\textwidth}\centering
     \caption*{\scriptsize \makecell{leopard:\\ 0.90}}
     \includegraphics[width=1\textwidth]{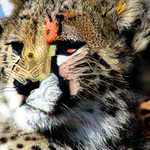}
     \end{subfigure}& 
     \begin{subfigure}{0.136\textwidth}\centering
     \caption*{\scriptsize \makecell{ leopard:\\0.39 }}
     \includegraphics[width=1\textwidth]{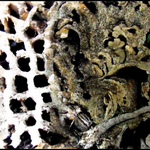}
     \end{subfigure} &

           \begin{subfigure}{0.136\textwidth}\centering
     \caption*{\scriptsize \makecell{tiger:\\ 1.00}}
     \includegraphics[width=1\textwidth]{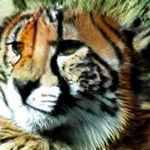}
     \end{subfigure} &
      \begin{subfigure}{0.136\textwidth}\centering
     \caption*{\scriptsize \makecell{tiger:\\ 0.97}}
     \includegraphics[width=1\textwidth]{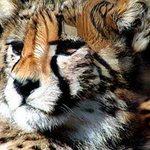}
     \end{subfigure}&
      \begin{subfigure}{0.136\textwidth}\centering
     \caption*{\scriptsize \makecell{ tiger:\\ 0.94}}
     \includegraphics[width=1\textwidth]{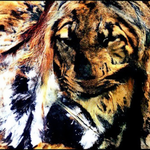}
     \end{subfigure}\\

      \begin{subfigure}{0.136\textwidth}\centering
     \caption*{\scriptsize \makecell{cheesburger:\\ 0.54}}
     \includegraphics[width=1\textwidth]{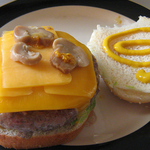}
     \end{subfigure} &

     \begin{subfigure}{0.136\textwidth}\centering
     \caption*{\scriptsize \makecell{pizza:\\ 1.00}}
     \includegraphics[width=1\textwidth]{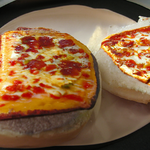}
     \end{subfigure} &
     \begin{subfigure}{0.136\textwidth}\centering
     \caption*{\scriptsize \makecell{pizza:\\ 1.00}}
     \includegraphics[width=1\textwidth]{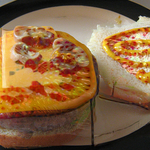}
     \end{subfigure} &
     \begin{subfigure}{0.136\textwidth}\centering
     \caption*{\scriptsize \makecell{pizza:\\ 0.94}}
     \includegraphics[width=1\textwidth]{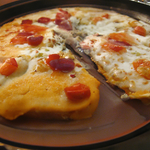}
     \end{subfigure} &

          \begin{subfigure}{0.136\textwidth}\centering
     \caption*{\scriptsize \makecell{potpie:\\ 0.99}}
     \includegraphics[width=1\textwidth]{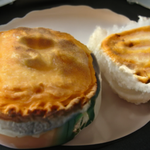}
     \end{subfigure} &
      \begin{subfigure}{0.136\textwidth}\centering
     \caption*{\scriptsize \makecell{potpie:\\ 0.90}}
     \includegraphics[width=1\textwidth]{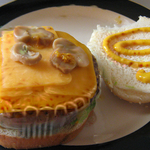}
     \end{subfigure}& 
      \begin{subfigure}{0.136\textwidth}\centering
     \caption*{\scriptsize \makecell{potpie:\\ 0.82}}
     \includegraphics[width=1\textwidth]{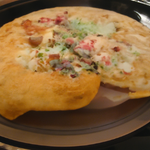}
     \end{subfigure}\\

      \begin{subfigure}{0.136\textwidth}\centering
     \caption*{\scriptsize \makecell{hognose snake:\\ 0.49}}
     \includegraphics[width=1\textwidth]{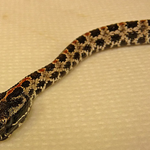}
     \end{subfigure} &

     \begin{subfigure}{0.136\textwidth}\centering
     \caption*{\scriptsize \makecell{king snake:\\ 1.00}}
     \includegraphics[width=1\textwidth]{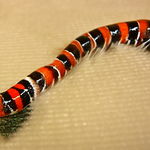}
     \end{subfigure} &
     \begin{subfigure}{0.136\textwidth}\centering
     \caption*{\scriptsize \makecell{king snake:\\ 0.98}}
     \includegraphics[width=1\textwidth]{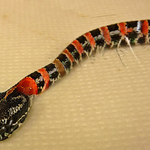}
     \end{subfigure} &
     \begin{subfigure}{0.136\textwidth}\centering
     \caption*{\scriptsize \makecell{king snake:\\ 0.98}}
     \includegraphics[width=1\textwidth]{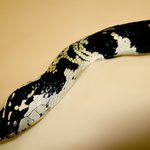}
     \end{subfigure} &

           \begin{subfigure}{0.136\textwidth}\centering
     \caption*{\scriptsize \makecell{night snake:\\ 0.71}}
     \includegraphics[width=1\textwidth]{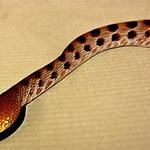}
     \end{subfigure} &
     \begin{subfigure}{0.136\textwidth}\centering
     \caption*{\scriptsize \makecell{night snake:\\ 0.82}}
     \includegraphics[width=1\textwidth]{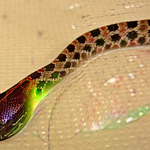}
     \end{subfigure}& 
      \begin{subfigure}{0.136\textwidth}\centering
     \caption*{\scriptsize \makecell{night snake:\\ 0.47}}
     \includegraphics[width=1\textwidth]{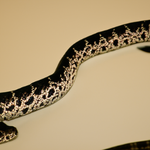}
     \end{subfigure}\\

      \begin{subfigure}{0.136\textwidth}\centering
     \caption*{\scriptsize \makecell{dingo:\\ 0.85}}
     \includegraphics[width=1\textwidth]{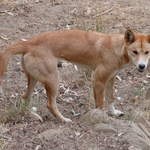}
     \end{subfigure} &

     \begin{subfigure}{0.136\textwidth}\centering
     \caption*{\scriptsize \makecell{timber wolf:\\ 0.98}}
     \includegraphics[width=1\textwidth]{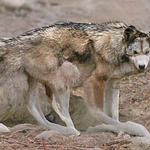}
     \end{subfigure} &\begin{subfigure}{0.136\textwidth}\centering
     \caption*{\scriptsize \makecell{timber wolf:\\ 0.97}}
     \includegraphics[width=1\textwidth]{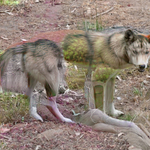}
     \end{subfigure} &
     \begin{subfigure}{0.136\textwidth}\centering
     \caption*{\scriptsize \makecell{timber wolf:\\ 0.77}}
     \includegraphics[width=1\textwidth]{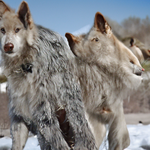}
     \end{subfigure} &

          \begin{subfigure}{0.136\textwidth}\centering
     \caption*{\scriptsize \makecell{white wolf:\\ 0.98}}
     \includegraphics[width=1\textwidth]{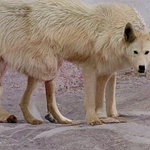}
     \end{subfigure} &
     \begin{subfigure}{0.136\textwidth}\centering
     \caption*{\scriptsize \makecell{white wolf:\\ 0.97}}
     \includegraphics[width=1\textwidth]{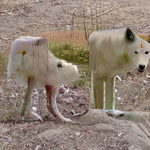}
     \end{subfigure}& 
      \begin{subfigure}{0.136\textwidth}\centering
     \caption*{\scriptsize \makecell{white wolf:\\ 0.92}}
     \includegraphics[width=1\textwidth]{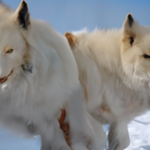}
     \end{subfigure}\\

      \begin{subfigure}{0.136\textwidth}\centering
     \caption*{\scriptsize \makecell{promontory:\\ 0.66}}
     \includegraphics[width=1\textwidth]{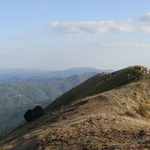}/
     \end{subfigure} &

     \begin{subfigure}{0.136\textwidth}\centering
     \caption*{\scriptsize \makecell{alp:\\ 1.00}}
     \includegraphics[width=1\textwidth]{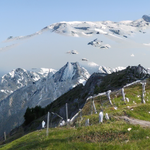}/
     \end{subfigure} &
     \begin{subfigure}{0.136\textwidth}\centering
     \caption*{\scriptsize \makecell{alp:\\ 0.90}}
     \includegraphics[width=1\textwidth]{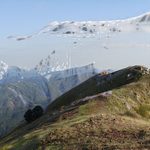}/
     \end{subfigure} &
     \begin{subfigure}{0.136\textwidth}\centering
     \caption*{\scriptsize \makecell{alp:\\ 0.82}}
     \includegraphics[width=1\textwidth]{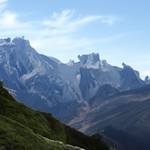}
     \end{subfigure} &

           \begin{subfigure}{0.136\textwidth}\centering
     \caption*{\scriptsize \makecell{volcano:\\ 1.00}}
     \includegraphics[width=1\textwidth]{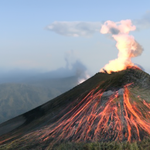}/
     \end{subfigure} &
      \begin{subfigure}{0.136\textwidth}\centering
     \caption*{\scriptsize \makecell{volcano:\\ 0.99}}
     \includegraphics[width=1\textwidth]{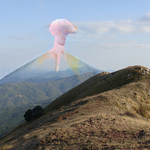}/
     \end{subfigure}& 
      \begin{subfigure}{0.136\textwidth}\centering
     \caption*{\scriptsize \makecell{volcano:\\ 0.96}}
     \includegraphics[width=1\textwidth]{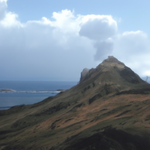}
     \end{subfigure}\\
     \end{tabular}
      \caption{\label{fig:method_comp}Comparison of different VCE-methods for the robust \madryft (taken from \protect\cite{boreiko2022sparse}). We show for each original image (outer left column)
       and for each target class our DVCEs (left column), the $l_{1.5}$-SVCEs of \protect\cite{boreiko2022sparse} (middle), and the adaptation of BDVCEs \protect\cite{avrahami2021blended}  (right).
       Only DVCEs %
      satisfy all desired properties of VCEs. BD fails for leopard and tiger and often produces images far away from the original one (pizza, potpie, timber wolf, white wolf, and alp). $l_{1.5}$-SVCEs show artefacts (white wolf, volcano, night snake) and have often lower image quality.}
     \end{figure}

\subsection{Final Scheme for Diffusion Visual Counterfactuals}\label{subsec:method}
Our solution for a non-adversarially robust classifier $p_\phi(y|\cdot)$ is to use Algorithm 1 of \cite{dhariwal2021diffusion} by replacing the update step with:
\begin{align*}
    \label{eq:guided_regularized}
        g_\mathrm{proj} &= P_{\text{cone}(\alpha, \nabla_{x_t} \log p_\phi(y|f_\mathrm{dn}(x_t, t)))} \Big[\nabla_{x_t} \log p_{\text{robust},\psi}\big(y|f_\mathrm{dn}(x_t, t)\big)\Big], \\
    g_\mathrm{update} &=  C_c \frac{ g_\mathrm{proj}}{\norm{g_\mathrm{proj}}
    _2} - C_d \frac{\nabla_{x_t} d(\hat{x}, f_\mathrm{dn}(x_t, t))}{\norm{\nabla_{x_t} d(\hat{x}, f_\mathrm{dn}(x_t, t))}_2} , \\
    \mu_{t} &= \mu_{\theta}(x_{t},t) + \Sigma_{\theta}(x_{t},t) \norm{\mu_{\theta}(x_{t},t)}_2 g_\mathrm{update}, \\
    p(x_{t-1}|x_{t},\hat{x},y) &= 
    \mathcal{N}(\mu_{t} ,\Sigma_{\theta}(x_{t},t)).
\end{align*}
 For an adversarially robust classifier the cone projection is omitted and one uses $g_\mathrm{proj} = \nabla_{x_t} \log p_{\text{robust},\psi}\big(y|f_\mathrm{dn}(x_t, t)\big)$. In all our experiments we use $C_c = 0.1$, and $C_d = 0.15$ unless we show ablations for one of the parameters. The angle $\alpha$ for the cone projection is fixed to $30^\circ$. In strong contrast to \vbcancel{\mh{Blended Diffusion}}\vb{BD}, these parameters generalize across images and classifiers.


\section{Experiments}
In this section, we evaluate the quality of the %
DVCE. We compare DVCE to existing works in Sec. \ref{subsec:method_comp}. In Sec. \ref{subsec:model_comp}, we compare DVCEs for various state-of-the-art ImageNet models and show how DVCEs can be used to interpret differences between classifiers. For our DVCEs, we use for all experiments the fixed hyperparameters given in Sec. \ref{subsec:method}. The diffusion model used for DVCE is taken from \cite{dhariwal2021diffusion} and has been trained class-unconditionally
on  256x256 ImageNet images using a modified UNet\cite{unet_olaf} architecture.
The user study, discovery of spurious features using VCEs, further experiments, and ablations are in the appendix.

\subsection{Comparison of Methods for VCE Generation}\label{subsec:method_comp}
We compare DVCEs with VCEs produced by BD (BDVCEs) and $l_{1.5}$-SVCEs.
As the latter only works for adversarially robust classifiers, we use the multiple-norm robust ResNet50 from \cite{boreiko2022sparse}, \madryft, as the classifier to create VCEs for all three methods. As this model is robust on its own, we do not use the cone projection for its DVCEs. 

First, in Fig. \ref{fig:method_comp} we present a qualitative evaluation, where we transform one image into two different classes that are close to the true one in the WordNet hierarchy\cite{miller1995wordnet}. %
 The radius of the $l_{1.5}$-ball of \cite{boreiko2022sparse} is chosen as the smallest $r \in \{50, 75, 100, 150\}$ such that the confidence in the target class is larger than $0.90$ per image. For BD, we select the image with the smallest classifier and regularization weight that reaches confidence larger than 0.9 from the set of parameters discussed in Sec. \ref{subsec:method}.  If 0.9 is not reached by any setting for one of the two baselines, we show the image that achieves the highest confidence. 
 As Fig. \ref{fig:method_comp} shows, DVCE is the only method that satisfies all desired properties of VCEs. %
 For example, for ``mashed potato'', DVCE preserves the bowl and only changes the content into either guacamole or carbonara. Our %
 qualitative comparison shows that the same hyperparameter setting of DVCE can handle %
 different classes and transfer between similar classes, such as different snakes or wolf types, as well as different object sizes, e.g. cheetah and snake. %
 
  In contrast, both $l_{1.5}$-SVCEs and BDVCEs require different hyperparameters for different images to achieve high confidence in the target class. 
 Even with the six parameter configurations, BD is not able to always produce  images with high confidence in the target class. %
 More problematic for VCEs is that the resulting images can have high confidence but are neither realistic (cheetah $\rightarrow$ tiger) nor resemble the original image at all (dingo $\rightarrow$ timber wolf/white wolf). 
 Even if the method works with the given parameters, for example, mashed potato $\rightarrow$ guacamole or carbonara, the overall image quality cannot match that of DVCE as often the images contain overly bright colors. In the case of the volcano VCE, DVCE shows class features like lava whereas BDVCE can not clearly be labeled as a volcano.
 For the $l_{1.5}$-SVCEs, one often needs large radii to achieve the desired confidence of 0.90, which often results in %
 images that %
 do not look realistic. 
 
 As noted in \cite{boreiko2022sparse}, a quantitative analysis of VCEs using FID scores is difficult as methods not changing the original image have low FID score. Thus, we have developed a cross-over evaluation scheme, where one partitions the classes into two sets and only analyzes cross-over VCEs (more details are in App. \ref{app:quant-eval}). We show the results in Tab. \ref{tab:quantitative_eval}. In terms of closeness, DVCEs are worse than $l_{1.5}$-SVCEs, which can be expected as they optimize inside an $l_{1.5}$-ball. However, DVCE are  the most realistic ones (FID score) and have similar validity as $l_{1.5}$-SVCEs. BDVCEs are the worst in all categories.

 Moreover, we have conducted a user study ($20$ users), in which participants decided if the changes of the VCE are \textbf{meaningful} or \textbf{subtle} and if the generated image is \textbf{realistic}, see App. \ref{app:user_study} for details. The percentage of total images having the three different properties is (order: DVCE/$l_{1.5}$-SVCE/BD): \textbf{meaningful} - \textbf{62.0\%}, 48.4\%, 38.7\%; \textbf{realism} - 34.7\%, 24.6\%, \textbf{52.2\%}; \textbf{subtle} -  45.0\%, \textbf{50.6\%}, 31.0\%. This confirms that DVCEs generate more meaningful features in the target classes. While the result regarding realism seems to contradict the quantitative evaluation, this is due to fact that realism  means that the user considered the image realistic irrespectively if it shows the target class or not. %
 \begin{figure*}[ht!]
     \centering
     \small
     \begin{tabular}{c|ccc|ccc}
     \hline
     \multicolumn{1}{c|}{Original}&
     \multicolumn{3}{c|}{Target Class 1} &    
     \multicolumn{3}{c}{Target Class 2}\\
     \hline
     & \scriptsize Swin-TF & \scriptsize ConvNeXt & \scriptsize EfficientNet &
       \scriptsize Swin-TF & \scriptsize ConvNeXt & \scriptsize EfficientNet \\
     \hline

      \begin{subfigure}{0.136\textwidth}\centering
     \caption*{\scriptsize \makecell{loggerhead\\ \quad}}
     \includegraphics[width=1\textwidth]{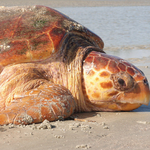}
     \end{subfigure} &

     \begin{subfigure}{0.136\textwidth}\centering
     \caption*{\scriptsize \makecell{leatherback:\\0.93}}
     \includegraphics[width=1\textwidth]{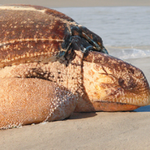}
     \end{subfigure} &
     \begin{subfigure}{0.136\textwidth}\centering
     \caption*{\scriptsize \makecell{leatherback:\\0.99}}
     \includegraphics[width=1\textwidth]{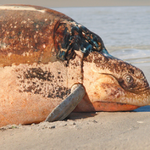}
     \end{subfigure} &
     \begin{subfigure}{0.136\textwidth}\centering
     \caption*{\scriptsize \makecell{leatherback:\\0.99}}
     \includegraphics[width=1\textwidth]{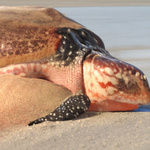}
     \end{subfigure} &

     \begin{subfigure}{0.136\textwidth}\centering
     \caption*{\scriptsize \makecell{box turtle:\\0.97}}
     \includegraphics[width=1\textwidth]{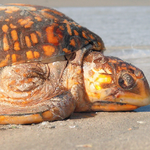}
     \end{subfigure} &
      \begin{subfigure}{0.136\textwidth}\centering
     \caption*{\scriptsize \makecell{box turtle:\\0.99}}
     \includegraphics[width=1\textwidth]{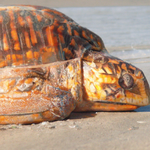}
     \end{subfigure} &
      \begin{subfigure}{0.136\textwidth}\centering
     \caption*{\scriptsize \makecell{box turtle:\\0.99}}
     \includegraphics[width=1\textwidth]{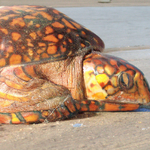}
     \end{subfigure}\\

     \begin{subfigure}{0.136\textwidth}\centering
     \caption*{\scriptsize \makecell{black and golden\\garden spider}}
     \includegraphics[width=1\textwidth]{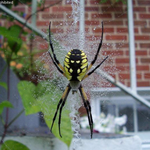}
     \end{subfigure} &

     \begin{subfigure}{0.136\textwidth}\centering
     \caption*{\scriptsize \makecell{black widow:\\0.95}}
     \includegraphics[width=1\textwidth]{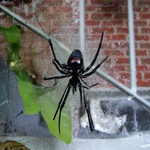}
     \end{subfigure} &

     \begin{subfigure}{0.136\textwidth}\centering
     \caption*{\scriptsize \makecell{black widow:\\0.99}}
     \includegraphics[width=1\textwidth]{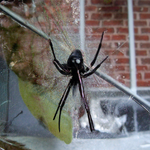}
     \end{subfigure} &

     \begin{subfigure}{0.136\textwidth}\centering
     \caption*{\scriptsize \makecell{black widow:\\1.00 }}
     \includegraphics[width=1\textwidth]{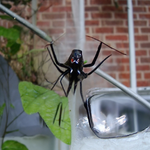}
     \end{subfigure} &
    
      \begin{subfigure}{0.136\textwidth}\centering
     \caption*{\scriptsize \makecell{tarantula:\\0.99}}
     \includegraphics[width=1\textwidth]{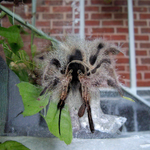}
     \end{subfigure} &
      \begin{subfigure}{0.136\textwidth}\centering
     \caption*{\scriptsize \makecell{tarantula:\\1.00}}
     \includegraphics[width=1\textwidth]{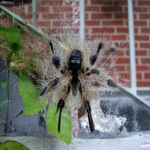}
     \end{subfigure} &
      \begin{subfigure}{0.136\textwidth}\centering
     \caption*{\scriptsize \makecell{tarantula:\\1.00}}
     \includegraphics[width=1\textwidth]{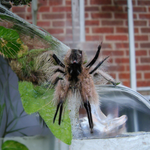}
     \end{subfigure}\\

      \begin{subfigure}{0.136\textwidth}\centering
     \caption*{\scriptsize \makecell{sea urchin\\ \quad}}
     \includegraphics[width=1\textwidth]{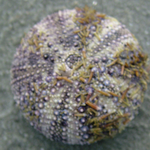}
     \end{subfigure} &

     \begin{subfigure}{0.136\textwidth}\centering
     \caption*{\scriptsize \makecell{sea cucumber:\\0.98}}
     \includegraphics[width=1\textwidth]{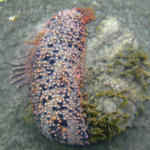}
     \end{subfigure} &
     \begin{subfigure}{0.136\textwidth}\centering
     \caption*{\scriptsize \makecell{sea cucumber:\\0.99}}
     \includegraphics[width=1\textwidth]{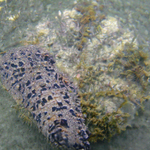}
     \end{subfigure} &

     \begin{subfigure}{0.136\textwidth}\centering
     \caption*{\scriptsize \makecell{sea cucumber:\\0.99}}
     \includegraphics[width=1\textwidth]{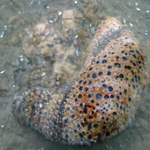}
     \end{subfigure} &

          \begin{subfigure}{0.136\textwidth}\centering
     \caption*{\scriptsize \makecell{starfish:\\0.98}}
     \includegraphics[width=1\textwidth]{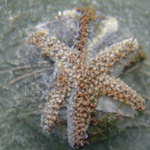}
     \end{subfigure} &
      \begin{subfigure}{0.136\textwidth}\centering
     \caption*{\scriptsize \makecell{starfish:\\0.99}}
     \includegraphics[width=1\textwidth]{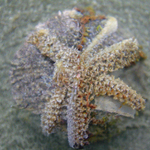}
     \end{subfigure} &
      \begin{subfigure}{0.136\textwidth}\centering
     \caption*{\scriptsize \makecell{starfish:\\0.98}}
     \includegraphics[width=1\textwidth]{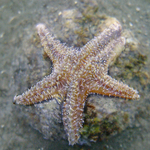}
     \end{subfigure}\\

      \begin{subfigure}{0.136\textwidth}\centering
     \caption*{\scriptsize \makecell{hen of the\\woods}}
     \includegraphics[width=1\textwidth]{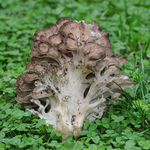}
     \end{subfigure} &

     \begin{subfigure}{0.136\textwidth}\centering
     \caption*{\scriptsize \makecell{agaric:\\0.93}}
     \includegraphics[width=1\textwidth]{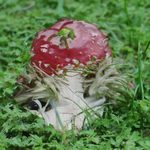}
     \end{subfigure} &

     \begin{subfigure}{0.136\textwidth}\centering
     \caption*{\scriptsize \makecell{agaric:\\1.00}}
     \includegraphics[width=1\textwidth]{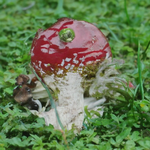}
     \end{subfigure} &

     \begin{subfigure}{0.136\textwidth}\centering
     \caption*{\scriptsize \makecell{agaric:\\0.98}}
     \includegraphics[width=1\textwidth]{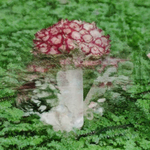}
     \end{subfigure} &

    \begin{subfigure}{0.136\textwidth}\centering
     \caption*{\scriptsize \makecell{gyromitra:\\0.96}}
     \includegraphics[width=1\textwidth]{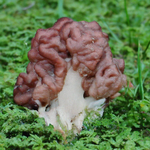}
     \end{subfigure} &
      \begin{subfigure}{0.136\textwidth}\centering
     \caption*{\scriptsize \makecell{gyromitra:\\0.99}}
     \includegraphics[width=1\textwidth]{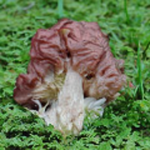}
     \end{subfigure} &
      \begin{subfigure}{0.136\textwidth}\centering
     \caption*{\scriptsize \makecell{gyromitra:\\1.00}}
     \includegraphics[width=1\textwidth]{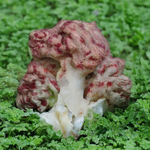}
     \end{subfigure}\\

      \begin{subfigure}{0.136\textwidth}\centering
     \caption*{\scriptsize \makecell{lorikeet\\ \quad}}
     \includegraphics[width=1\textwidth]{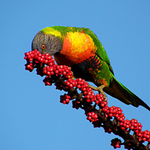}
     \end{subfigure} &

     \begin{subfigure}{0.136\textwidth}\centering
     \caption*{\scriptsize \makecell{macaw:\\0.99}}
     \includegraphics[width=1\textwidth]{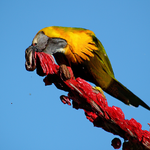}
     \end{subfigure} &

     \begin{subfigure}{0.136\textwidth}\centering
     \caption*{\scriptsize \makecell{macaw:\\0.99}}
     \includegraphics[width=1\textwidth]{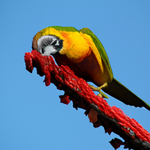}
     \end{subfigure} &

     \begin{subfigure}{0.136\textwidth}\centering
     \caption*{\scriptsize \makecell{macaw:\\0.99}}
     \includegraphics[width=1\textwidth]{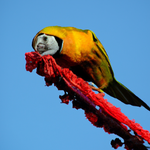}
     \end{subfigure} &
    
          \begin{subfigure}{0.136\textwidth}\centering
     \caption*{\scriptsize \makecell{african grey:\\0.98}}
     \includegraphics[width=1\textwidth]{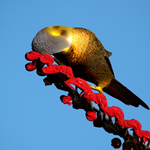}
     \end{subfigure} &
      \begin{subfigure}{0.136\textwidth}\centering
     \caption*{\scriptsize \makecell{african grey:\\0.99}}
     \includegraphics[width=1\textwidth]{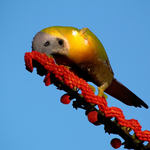}
     \end{subfigure} &
      \begin{subfigure}{0.136\textwidth}\centering
     \caption*{\scriptsize \makecell{african grey:\\0.99}}
     \includegraphics[width=1\textwidth]{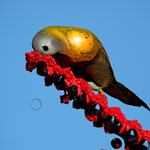}
     \end{subfigure}\\

      \begin{subfigure}{0.136\textwidth}\centering
     \caption*{\scriptsize \makecell{mosque\\ \quad}}
     \includegraphics[width=1\textwidth]{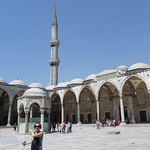}
     \end{subfigure} &

     \begin{subfigure}{0.136\textwidth}\centering
     \caption*{\scriptsize \makecell{stupa:\\0.98}}
     \includegraphics[width=1\textwidth]{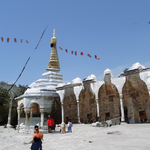}
     \end{subfigure} &

     \begin{subfigure}{0.136\textwidth}\centering
     \caption*{\scriptsize \makecell{stupa:\\0.98}}
     \includegraphics[width=1\textwidth]{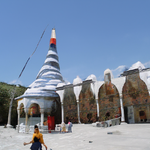}
     \end{subfigure} &

     \begin{subfigure}{0.136\textwidth}\centering
     \caption*{\scriptsize \makecell{stupa:\\1.00}}
     \includegraphics[width=1\textwidth]{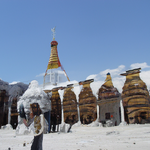}
     \end{subfigure} &

      \begin{subfigure}{0.136\textwidth}\centering
     \caption*{\scriptsize \makecell{church:\\0.99}}
     \includegraphics[width=1\textwidth]{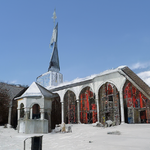}
     \end{subfigure} &
      \begin{subfigure}{0.136\textwidth}\centering
     \caption*{\scriptsize \makecell{church:\\0.99}}
     \includegraphics[width=1\textwidth]{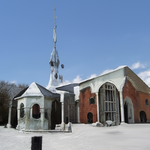}
     \end{subfigure} &
      \begin{subfigure}{0.136\textwidth}\centering
     \caption*{\scriptsize \makecell{church:\\0.98}}
     \includegraphics[width=1\textwidth]{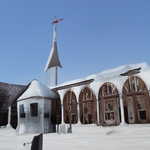}
     \end{subfigure}\\

      \begin{subfigure}{0.136\textwidth}\centering
     \caption*{\scriptsize \makecell{red wolf\\ \quad}}
     \includegraphics[width=1\textwidth]{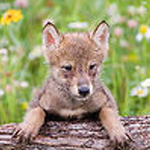}
     \end{subfigure} &

     \begin{subfigure}{0.136\textwidth}\centering
     \caption*{\scriptsize \makecell{coyote:\\0.99}}
     \includegraphics[width=1\textwidth]{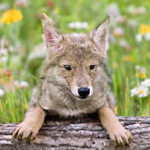}
     \end{subfigure} &

     \begin{subfigure}{0.136\textwidth}\centering
     \caption*{\scriptsize \makecell{coyote:\\1.00}}
     \includegraphics[width=1\textwidth]{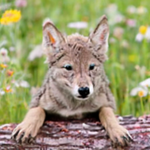}
     \end{subfigure} &

     \begin{subfigure}{0.136\textwidth}\centering
     \caption*{\scriptsize \makecell{coyote:\\1.00}}
     \includegraphics[width=1\textwidth]{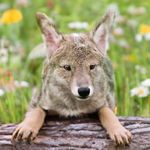}
     \end{subfigure} &

      \begin{subfigure}{0.136\textwidth}\centering
     \caption*{\scriptsize \makecell{timber wolf:\\0.99}}
     \includegraphics[width=1\textwidth]{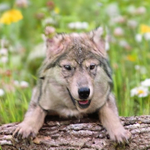}
     \end{subfigure} &
      \begin{subfigure}{0.136\textwidth}\centering
     \caption*{\scriptsize \makecell{timber wolf:\\0.99}}
     \includegraphics[width=1\textwidth]{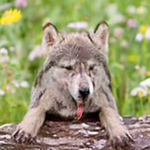}
     \end{subfigure} &
      \begin{subfigure}{0.136\textwidth}\centering
     \caption*{\scriptsize \makecell{timber wolf:\\1.00}}
     \includegraphics[width=1\textwidth]{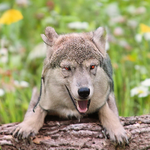}
     \end{subfigure} \\

     \begin{subfigure}{0.136\textwidth}\centering
     \caption*{\scriptsize \makecell{pineapple\\ \quad}}
     \includegraphics[width=1\textwidth]{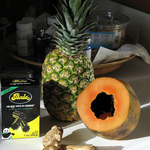}
     \end{subfigure} &
     
     \begin{subfigure}{0.136\textwidth}\centering
     \caption*{\scriptsize \makecell{jackfruit:\\0.99}}
     \includegraphics[width=1\textwidth]{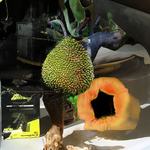}
     \end{subfigure} &
     \begin{subfigure}{0.136\textwidth}\centering
     \caption*{\scriptsize \makecell{jackfruit:\\0.99}}
     \includegraphics[width=1\textwidth]{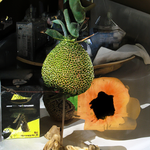}
     \end{subfigure} &
      \begin{subfigure}{0.136\textwidth}\centering
     \caption*{\scriptsize \makecell{jackfruit:\\0.99}}
     \includegraphics[width=1\textwidth]{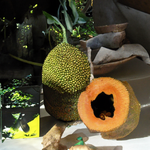}
     \end{subfigure} &

       \begin{subfigure}{0.136\textwidth}\centering
     \caption*{\scriptsize \makecell{custard apple:\\0.96}}
     \includegraphics[width=1\textwidth]{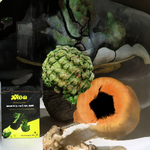}
     \end{subfigure} &
      \begin{subfigure}{0.136\textwidth}\centering
     \caption*{\scriptsize \makecell{custard apple:\\0.99}}
     \includegraphics[width=1\textwidth]{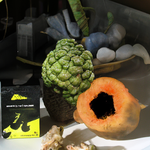}
     \end{subfigure} &
           \begin{subfigure}{0.136\textwidth}\centering
     \caption*{\scriptsize \makecell{custard apple:\\0.99}}
     \includegraphics[width=1\textwidth]{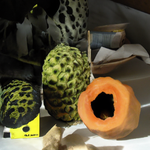}
     \end{subfigure}
     \\

     \end{tabular}

      \caption{\label{fig:classifier_comp}
      DVCEs for three non-robust classifiers: Swin-TF\protect\cite{liu2021swin}, ConvNeXt\protect\cite{liu2022convnet} and EfficientNet\protect\cite{xie2020selftraining}.
      For each original image (outer left column) we show %
      DVCEs into two target classes. Please zoom into the images to see more fine-grained details and differences.}
 \end{figure*}
\begin{table}[hbt!]
    \centering
    \begin{tabular}{|c|c|c|c|c|c|c|c|}
         \cline{2-7}
         \multicolumn{1}{c|}{}& \multicolumn{4}{c|}{\textbf{Closeness}} & \multicolumn{1}{c|}{\textbf{Validity}} & \multicolumn{1}{c|}{\textbf{Realism}} \\
         \hline
         Metric & $l_{1} \downarrow$ &$l_{1.5} \downarrow$ &  $l_{2} \downarrow$ & LPIPS-Alex $\downarrow$ & Mean Conf. $\uparrow$ & Avg. FID $\downarrow$ \\
         \hline
         \hline
         DVCEs (ours)&  $\;$12799$\;$ & $\;$293$\;$ &  $\;$48$\;$ & 0.35 & 0.932 & \textbf{17.6} \\
         $l_{1.5}$-SVCEs\cite{boreiko2022sparse} & \textbf{5139} & \textbf{139} & \textbf{26} & \textbf{0.20} & \textbf{0.945} & 25.6 \\
         Blended Diffusion\cite{avrahami2021blended}& 35678 & 722 & $\;$108$\;$ & 0.58 & 0.825 & 27.9 \\
         \hline
    \end{tabular}\\
    \vspace{0.5em}
    \caption{
    Quantitative evaluation of VCEs. DVCEs outperform BDVCEs \cite{avrahami2021blended} in all metrics. Moreover, they achieve comparable to $l_{1.5}$-SVCEs Mean Conf. (validity), while outperforming them significantly in Avg. FID. (realism) but do larger changes to the image than SVCE (closeness).}\label{tab:quantitative_eval}
\end{table}

\subsection{Model comparison}\label{subsec:model_comp}
\textbf{Non-robust models.} In Fig. \ref{fig:classifier_comp} we show that DVCEs can be generated for various state-of-the-art ImageNet \cite{ILSVRC15} models. We use a Swin-TF\cite{liu2021swin}, a ConvNeXt \cite{liu2022convnet} and a Noisy-Student EfficientNet \cite{xie2020selftraining, tan2019efficientnet}. Both the Swin-TF and the ConvNeXt are pretrained on ImageNet21k \cite{ImageNet21k} whereas the EfficientNet uses noisy-student self-training on a large unlabeled pool. As all models do not yield perceptually aligned gradients, we use the cone projection with $30^\circ$ angle described in Sec. \ref{subsec:cone} with the robust model from the previous section. All other parameters are identical to the previous experiment. We highlight that DVCE is the first method capable of explaining (in the sense of VCEs) arbitrary classifiers on a task as challenging as ImageNet. Overall, DVCEs %
satisfy all desired properties of VCEs.
It also allows us to inspect the most important features of each model and class. %
For the stupa and church classes, for example, it seems like all models use different roof and tower structures as the most prominent feature as they spent most of their budget on changing those.

\begin{figure*}[ht!]
     \centering
     \small
     \begin{tabular}{c|ccc|ccc}
     \hline
     \multicolumn{1}{c|}{Original}&
     \multicolumn{3}{c|}{Target Class 1} &    
     \multicolumn{3}{c}{Target Class 2}\\
     \hline
     & \scriptsize \madryft & \scriptsize \xcitft & \scriptsize \deitft 
          & \scriptsize \madryft & \scriptsize \xcitft & \scriptsize \deitft
          \\
     \hline

      \begin{subfigure}{0.136\textwidth}\centering
     \caption*{\scriptsize \makecell{pirate ship\\ \quad}}
     \includegraphics[width=1\textwidth]{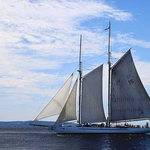}
     \end{subfigure} &

     \begin{subfigure}{0.136\textwidth}\centering
     \caption*{\scriptsize \makecell{liner ship:\\1.00}}
     \includegraphics[width=1\textwidth]{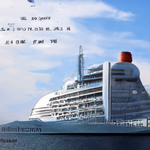}
     \end{subfigure} &
     \begin{subfigure}{0.136\textwidth}\centering
     \caption*{\scriptsize \makecell{liner ship:\\1.00}}
     \includegraphics[width=1\textwidth]{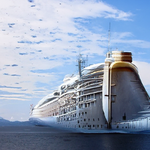}
     \end{subfigure} &
     \begin{subfigure}{0.136\textwidth}\centering
     \caption*{\scriptsize \makecell{liner ship:\\1.00}}
     \includegraphics[width=1\textwidth]{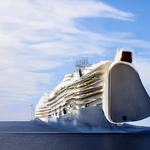}
     \end{subfigure} &

    \begin{subfigure}{0.136\textwidth}\centering
     \caption*{\scriptsize \makecell{container ship:\\1.00}}
     \includegraphics[width=1\textwidth]{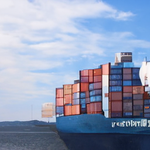}
     \end{subfigure} &
      \begin{subfigure}{0.136\textwidth}\centering
     \caption*{\scriptsize \makecell{container ship:\\1.00}}
     \includegraphics[width=1\textwidth]{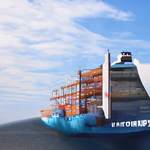}
     \end{subfigure} &
      \begin{subfigure}{0.136\textwidth}\centering
     \caption*{\scriptsize \makecell{container ship:\\1.00}}
     \includegraphics[width=1\textwidth]{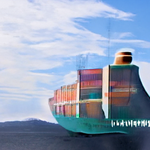}
     \end{subfigure}\\

      \begin{subfigure}{0.136\textwidth}\centering
     \caption*{\scriptsize \makecell{mashed potato\\ \quad}}
     \includegraphics[width=1\textwidth]{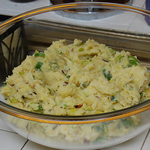}
     \end{subfigure} &

     \begin{subfigure}{0.136\textwidth}\centering
     \caption*{\scriptsize \makecell{dough:\\1.00}}
     \includegraphics[width=1\textwidth]{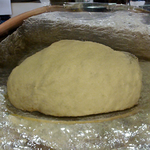}
     \end{subfigure} &

     \begin{subfigure}{0.136\textwidth}\centering
     \caption*{\scriptsize \makecell{dough:\\1.00}}
     \includegraphics[width=1\textwidth]{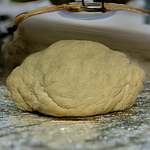}
     \end{subfigure} &

     \begin{subfigure}{0.136\textwidth}\centering
     \caption*{\scriptsize \makecell{dough:\\1.00}}
     \includegraphics[width=1\textwidth]{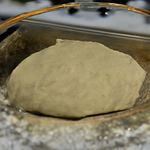}
     \end{subfigure} &
    
          \begin{subfigure}{0.136\textwidth}\centering
     \caption*{\scriptsize \makecell{carbonara:\\1.00}}
     \includegraphics[width=1\textwidth]{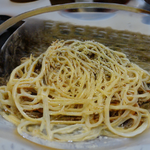}
     \end{subfigure} &
      \begin{subfigure}{0.136\textwidth}\centering
     \caption*{\scriptsize \makecell{carbonara:\\1.00}}
     \includegraphics[width=1\textwidth]{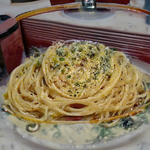}
     \end{subfigure} &
      \begin{subfigure}{0.136\textwidth}\centering
     \caption*{\scriptsize \makecell{carbonara:\\1.00}}
     \includegraphics[width=1\textwidth]{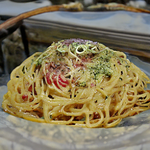}
     \end{subfigure}\\
    
     \end{tabular}

      \caption{\label{fig:app_robust_comp_small}
      We compare DVCEs for three different robust models (no cone projection) which are all fine-tuned to be multiple-norm adversarially robust \protect\cite{croce2021adversarial}, that is against $l_1$, $l_2$ and $l_\infty$-perturbations. More examples can be found in Fig. \ref{fig:app_robust_comp}.}
 \end{figure*}

\textbf{Robust models.} 
Here, we evaluate the DVCEs of different robust models  trained to be multiple-norm adversarially robust. They are generated by multiple norm-finetuning  \cite{croce2021adversarial} an initially $l_p$-robust model to become robust with respect to $l_1$-, $l_2$- and $l_\infty$-threat model, specifically an $l_2$-robust ResNet50 \cite{robustness} resulting in \madryft used in the revious sections, an  $l_\infty$-robust XCiT-transformer model \cite{debenedetti2022adversarially} called \xcitft in the following and an $l_\infty$-robust DeiT-transformer \cite{bai2021transformers} called \deitft.
Their multiple-norm robust accuracies can be found in \cite{croce2021adversarial}. 
In \cite{boreiko2022sparse}, different $l_p$-ball adversarially robust models were compared for the generation of VCEs and they showed that for their $l_{1.5}$-SVCEs the multiple norm robust model was better (both in terms of FID and qualitatively) than individual $l_p$-norm robust classifiers. This is the reason why we also use a multiple-norm robust model for the cone-projection of a non-robust classifier.  %
In Fig. \ref{fig:app_robust_comp_small}, we show DVCEs of the three different classes for two examples images and two target classes each. All of the multiple-norm robust models have similarly good DVCEs showing classifier-specific variations in the semantic changes. This proves again the generative properties of adversarially robust models, in particular of robust transformers. More examples and further experiments are in App. \ref{app:different_robust_models}.

\section{Limitations, Future Work and Societal impact}\label{sec:limitations}
In comparison to GANs, diffusion-based approaches can be expensive to evaluate as they require multiple iterations of the reverse process to create one sample. This can be an issue when creating a large amount of VCEs and makes deployment in a time-sensitive setting challenging.\\
We also rely on a robust model during the creation of VCEs for the cone projection. While we show that standard classifiers do not yield the desired gradients, training robust models can be challenging. 
An interesting direction for future research is thus to replace the cone projection with another ``denoising'' procedure for the gradient of a non-robust model.\\
From a theoretical standpoint, using conditional sampling with reverse transitions of the form \eqref{eq:cond_reverse} is justified, in practice, however, we have to approximate the reverse transitions with shifted normal distributions of the form \eqref{eq:transition_kernel}. This approximation relies on the conditioning function having low curvature, and it can be hard to verify this once we start adding a classifier, distance and other possible terms and it is unclear how this influences the outcome of the diffusion process.\\
The quantitative evaluation of VCEs is difficult, as the standard FID metric compares the distribution of features of a classifier over a test and a generated dataset. However, for VCEs it is not only important to generate realistic images, but also to achieve high confidence and to create meaningful changes. Moreover, metrics such as IM1, IM2 \cite{looveren2019im1im2} for VCEs rely on a well-trained (V)AE for every class, which is difficult to achieve for a dataset with 1000 classes and high-resolution images. Future research should therefore try to develop metrics for the quantitative evaluation of VCEs and we think that the evaluation in Tab. \ref{tab:quantitative_eval} is the first step in that direction.\\
DVCEs and VCEs in general help to discover biases of the classifiers and thus have a positive societal impact, however one can abuse them for unintended purposes as any conditional generative model.

\section{Conclusion}

We have proposed DVCEs, a novel way to create VCEs for any state-of-the-art ImageNet classifier using diffusion models and our cone projection. DVCEs can handle vastly different image configurations, object sizes, and classes and satisfy all desired properties of VCEs. 
\section*{Acknowledgement}
    The authors acknowledge support by the DFG Excellence Cluster Machine Learning - New Perspectives for Science, EXC 2064/1, Project number 390727645 and the German Federal Ministry of Education and Research (BMBF): Tübingen AI Center, FKZ: 01IS18039A as well as the DFG grant 389792660 as part of TRR 248. 

\clearpage
\bibliography{arxiv}

\newpage
\section*{Checklist}

\begin{enumerate}

\item For all authors...
\begin{enumerate}
  \item Do the main claims made in the abstract and introduction accurately reflect the paper's contributions and scope?
    \answerYes{}
  \item Did you describe the limitations of your work?
    \answerYes{See Sec. \ref{sec:limitations}.}
  \item Did you discuss any potential negative societal impacts of your work?
    \answerYes{See Sec. \ref{sec:limitations}.}
  \item Have you read the ethics review guidelines and ensured that your paper conforms to them?
    \answerYes{}
\end{enumerate}

\item If you are including theoretical results...
\begin{enumerate}
  \item Did you state the full set of assumptions of all theoretical results?
     \answerNA{}
        \item Did you include complete proofs of all theoretical results?
     \answerNA{}
\end{enumerate}

\item If you ran experiments...
\begin{enumerate}
  \item Did you include the code, data, and instructions needed to reproduce the main experimental results (either in the supplemental material or as a URL)?
    \answerYes{See supplemental material.}
  \item Did you specify all the training details (e.g., data splits, hyperparameters, how they were chosen)?
    \answerNA{We didn't train models.}
        \item Did you report error bars (e.g., with respect to the random seed after running experiments multiple times)?
    \answerNA{We have conducted qualitative experiments in the main paper for the same seed and show the diversity of VCEs across seeds in App. \ref{app:diversity}.}
        \item Did you include the total amount of compute and the type of resources used (e.g., type of GPUs, internal cluster, or cloud provider)?
    \answerYes{See App. \ref{app:hardware}}
\end{enumerate}

\item If you are using existing assets (e.g., code, data, models) or curating/releasing new assets...
\begin{enumerate}
  \item If your work uses existing assets, did you cite the creators?
    \answerYes{}
  \item Did you mention the license of the assets?
    \answerYes{Yes, directly in the citation, when applicable.}
  \item Did you include any new assets either in the supplemental material or as a URL?
    \answerYes{We include our code in the supplemental material.}
  \item Did you discuss whether and how consent was obtained from people whose data you're using/curating?
    \answerNA{ImangeNet and ImageNet21k are public datasets.}
  \item Did you discuss whether the data you are using/curating contains personally identifiable information or offensive content?
    \answerNA{}
\end{enumerate}

\item If you used crowdsourcing or conducted research with human subjects...
\begin{enumerate}
  \item Did you include the full text of instructions given to participants and screenshots, if applicable?
    \answerYes{We include the results from the user study together with the screenshots of instructions in App. \ref{app:user_study}.}
  \item Did you describe any potential participant risks, with links to Institutional Review Board (IRB) approvals, if applicable?
    \answerNA{Not applicable.}
  \item Did you include the estimated hourly wage paid to participants and the total amount spent on participant compensation?
    \answerYes{See App. \ref{app:user_study}.}
\end{enumerate}

\end{enumerate}

\clearpage

\appendix

\section{Appendix}
\section*{Overview of Appendix}
In the following, we provide a brief overview of the additional experiments reported in the Appendix.

\begin{itemize}
    \item In App. \ref{app:ablation}, we show how varying $l_1$ regularization weight (\ref{app:regularization}), the starting point of the diffusion process $T$ (\ref{app:starting_T}), diversity (\ref{app:diversity}), and different levels of robustness (\ref{app:different_robust_models}) influence the DVCEs. In \ref{app:cone-projection-angle},  we add the ablation study of the angle used in the cone projection.
    \item In App. \ref{app:user_study}, we show the results of our user study which shows that both quantitatively and qualitatively DVCEs generate more meaningful features compared to $l_{1.5}$-SVCEs \cite{boreiko2022sparse} and BDVCEs \cite{avrahami2021blended}.
    \item In App. \ref{app:hardware}, we describe the hardware and resources used.
    \item In App. \ref{app:quant-eval}, we explain the quantitative evaluation of the realism, validity, and closeness of our DVCEs.
    \item In App. \ref{app:spurious-features}, we show, how our DVCEs can help to uncover spurious features.
    \item In App. \ref{app:failure-cases}, we show some of the failure cases of our method.
\end{itemize}
\section{Ablation study of DVCEs}\label{app:ablation}
\subsection{Regularization}\label{app:regularization}
\begin{figure*}[h]
     \centering
     \small
     \begin{tabular}{c|ccc||c|ccc}
     \hline
     \multicolumn{1}{c|}{Original}&

     $C_d=0.05$ & $C_d=0.15$ & $C_d=0.25$ &
     \multicolumn{1}{c|}{Original}&
     $C_d=0.05$ & $C_d=0.15$ & $C_d=0.25$\\
     \hline
     
     \begin{subfigure}{0.121\textwidth}\centering
     \caption*{\scriptsize \makecell{kite:\\ 0.88}}
     \includegraphics[width=1\textwidth]{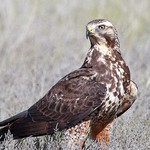}
     \end{subfigure} &
      \begin{subfigure}{0.121\textwidth}\centering
       \caption*{\scriptsize \makecell{bald eagle:\\ 0.98}}
     \includegraphics[width=1\textwidth]{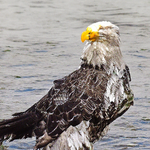}
     \end{subfigure} &
      \begin{subfigure}{0.121\textwidth}\centering
       \caption*{\scriptsize \makecell{bald eagle:\\ 0.98}}
     \includegraphics[width=1\textwidth]{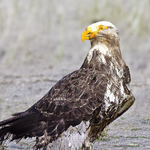}
     \end{subfigure} &
      \begin{subfigure}{0.121\textwidth}\centering
       \caption*{\scriptsize \makecell{bald eagle:\\ 0.98}}
     \includegraphics[width=1\textwidth]{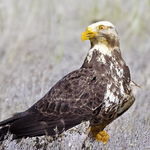}
     \end{subfigure} &

     \begin{subfigure}{0.121\textwidth}\centering
       \caption*{\scriptsize \makecell{jellyfish:\\ 0.96}}
     \includegraphics[width=1\textwidth]{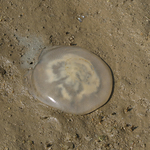}
     \end{subfigure} &
      \begin{subfigure}{0.121\textwidth}\centering
       \caption*{\scriptsize \makecell{sea anemone:\\ 1.00}}
     \includegraphics[width=1\textwidth]{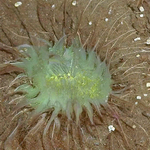}
     \end{subfigure} &
      \begin{subfigure}{0.121\textwidth}\centering
       \caption*{\scriptsize \makecell{sea anemone:\\ 0.98}}
     \includegraphics[width=1\textwidth]{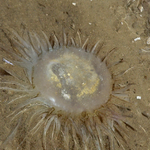}
     \end{subfigure} &
      \begin{subfigure}{0.121\textwidth}\centering
       \caption*{\scriptsize \makecell{sea anemone:\\ 0.98}}
     \includegraphics[width=1\textwidth]{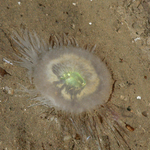}
     \end{subfigure}\\
     \hline
     \begin{subfigure}{0.121\textwidth}\centering
       \caption*{\scriptsize \makecell{Siamese cat:\\ 0.95}}
     \includegraphics[width=1\textwidth]{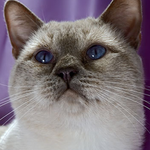}
     \end{subfigure} &
      \begin{subfigure}{0.121\textwidth}\centering
       \caption*{\scriptsize \makecell{Egyptian cat:\\ 0.99}}
     \includegraphics[width=1\textwidth]{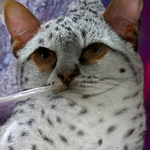}
     \end{subfigure} &
      \begin{subfigure}{0.121\textwidth}\centering
       \caption*{\scriptsize \makecell{Egyptian cat:\\ 0.99}}
     \includegraphics[width=1\textwidth]{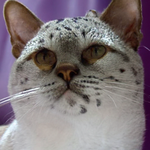}
     \end{subfigure} &
      \begin{subfigure}{0.121\textwidth}\centering
       \caption*{\scriptsize \makecell{Egyptian cat:\\ 0.97}}
     \includegraphics[width=1\textwidth]{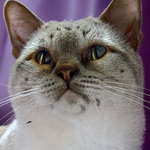}
     \end{subfigure} &

     \begin{subfigure}{0.121\textwidth}\centering
       \caption*{\scriptsize \makecell{house finch:\\ 0.96}}
     \includegraphics[width=1\textwidth]{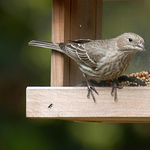}
     \end{subfigure} &
      \begin{subfigure}{0.121\textwidth}\centering
       \caption*{\scriptsize \makecell{junco:\\ 0.98}}
     \includegraphics[width=1\textwidth]{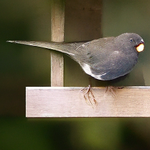}
     \end{subfigure} &
      \begin{subfigure}{0.121\textwidth}\centering
       \caption*{\scriptsize \makecell{junco:\\ 0.98}}
     \includegraphics[width=1\textwidth]{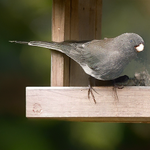}
     \end{subfigure} &
      \begin{subfigure}{0.121\textwidth}\centering
       \caption*{\scriptsize \makecell{junco:\\ 0.97}}
     \includegraphics[width=1\textwidth]{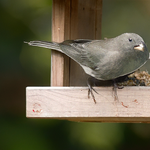}
     \end{subfigure}\\

     \end{tabular}

      \caption{\label{fig:reg_comp}
      Ablation study for varying weights $C_d$ for the $l_1$-distance term in our DVCEs for the non-robust Swin-TF using the cone projection introduced in Sec. \ref{subsec:cone}.  
    }
     \end{figure*}

In this section, we start by evaluating the impact of the distance regularization term on the diffusion process. We want VCEs to resemble the original image in the overall appearance and only change class-specific features to transfer the image into the target class. To achieve sparse changes, we use $l_1$-distance regularization. In Fig. \ref{fig:reg_comp}, we vary the regularization strengths from $0.05$ to $0.25$. As can be seen, all regularization strengths generate meaningful target class-specific features, however, regularization $C_d$ of $0.15$ gives a good trade-off between being close to the original image as well as being realistic.
For the rest of the evaluation, we, therefore, chose a weight of $0.15$. 

\subsection{Starting $T_\mathrm{start}$}\label{app:starting_T}
Next, we show how the starting time of the diffusion process influences the DVCEs. Diffusion-based sampling that is not conditioned on an original image starts the process at standard normally distributed noise. However, we show that it is easier to obtain high-quality VCEs that are similar to the original image by instead starting in the middle of the diffusion process by sampling from the closed-form probability distribution corresponding to timestep  $T_\mathrm{start}$ \eqref{eq:forward_t}. For this, we compare three different settings and define $\eta := \frac{T_\mathrm{start}}{T}$ where we choose $\eta \in \{0.25, 0.5, 0.75\}$. We observe again that $\eta = 0.5$ gives a good trade-off between closeness, realism, and showing class-specific features. Thus, in all our experiments in this paper, we chose $\eta = 0.5$.
\begin{figure*}[h]
     \centering
     \small
     \begin{tabular}{c|ccc||c|ccc}
     \hline
     \multicolumn{1}{c|}{Original}&
     $\eta = 0.75$ & $\eta = 0.5$ & $\eta = 0.25$ &
     \multicolumn{1}{c|}{Original}&
     $\eta = 0.75$ & $\eta = 0.5$ & $\eta = 0.25$ \\
     \hline
     
     \begin{subfigure}{0.121\textwidth}\centering
     \caption*{\scriptsize \makecell{kite:\\ 0.88}}
     \includegraphics[width=1\textwidth]{images/swin/bird of prey raptor raptorial bird_kite_5_1096/original.png}
     \end{subfigure} &
      \begin{subfigure}{0.121\textwidth}\centering
       \caption*{\scriptsize \makecell{bald eagle:\\0.98 }}
     \includegraphics[width=1\textwidth]{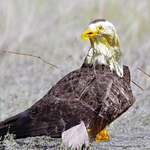}
     \end{subfigure} &
      \begin{subfigure}{0.121\textwidth}\centering
       \caption*{\scriptsize \makecell{bald eagle:\\ 0.98}}
     \includegraphics[width=1\textwidth]{images/swin_seed_1_0_15/0.png}
     \end{subfigure} &
      \begin{subfigure}{0.121\textwidth}\centering
       \caption*{\scriptsize \makecell{bald eagle:\\0.90}}
     \includegraphics[width=1\textwidth]{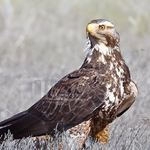}
     \end{subfigure} &

     \begin{subfigure}{0.121\textwidth}\centering
       \caption*{\scriptsize \makecell{jellyfish:\\ 0.96}}
     \includegraphics[width=1\textwidth]{images/swin/coelenterate cnidarian_jellyfish_6_5391/original.png}
     \end{subfigure} &
      \begin{subfigure}{0.121\textwidth}\centering
       \caption*{\scriptsize \makecell{sea anemone:\\0.97 }}
     \includegraphics[width=1\textwidth]{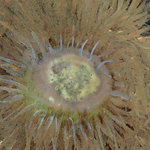}
     \end{subfigure} &
      \begin{subfigure}{0.121\textwidth}\centering
       \caption*{\scriptsize \makecell{sea anemone:\\ 0.98}}
     \includegraphics[width=1\textwidth]{images/swin_seed_1_0_15/1.png}
     \end{subfigure} &
      \begin{subfigure}{0.121\textwidth}\centering
       \caption*{\scriptsize \makecell{sea anemone:\\0.99}}
     \includegraphics[width=1\textwidth]{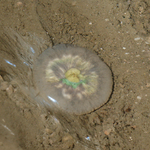}
     \end{subfigure}\\
     \hline
     \begin{subfigure}{0.121\textwidth}\centering
       \caption*{\scriptsize \makecell{Siamese cat:\\ 0.95}}
     \includegraphics[width=1\textwidth]{images/swin/domestic cat house cat Felis domesticus Felis catus_Siamese cat_10_14201/original.png}
     \end{subfigure} &
      \begin{subfigure}{0.121\textwidth}\centering
       \caption*{\scriptsize \makecell{Egyptian cat:\\0.99 }}
     \includegraphics[width=1\textwidth]{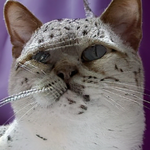}
     \end{subfigure} &
      \begin{subfigure}{0.121\textwidth}\centering
       \caption*{\scriptsize \makecell{Egyptian cat:\\ 0.99}}
     \includegraphics[width=1\textwidth]{images/swin_seed_1_0_15/2.png}
     \end{subfigure} &
      \begin{subfigure}{0.121\textwidth}\centering
       \caption*{\scriptsize \makecell{Egyptian cat:\\0.76}}
     \includegraphics[width=1\textwidth]{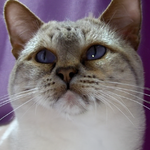}
     \end{subfigure} &

     \begin{subfigure}{0.121\textwidth}\centering
       \caption*{\scriptsize \makecell{house finch:\\ 0.96}}
     \includegraphics[width=1\textwidth]{images/swin/finch_house finch_12_649/original.png}
     \end{subfigure} &
      \begin{subfigure}{0.121\textwidth}\centering
       \caption*{\scriptsize \makecell{junco:\\0.97 }}
     \includegraphics[width=1\textwidth]{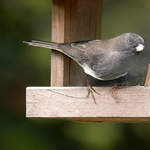}
     \end{subfigure} &
      \begin{subfigure}{0.121\textwidth}\centering
       \caption*{\scriptsize \makecell{junco:\\ 0.98}}
     \includegraphics[width=1\textwidth]{images/swin_seed_1_0_15/3.png}
     \end{subfigure} &
      \begin{subfigure}{0.121\textwidth}\centering
       \caption*{\scriptsize \makecell{junco:\\0.96 }}
     \includegraphics[width=1\textwidth]{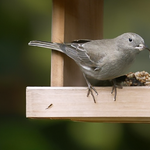}
     \end{subfigure}\\

     \end{tabular}

      \caption{\label{fig:start_t_comp}
      Comparison of different starting times $\eta := \frac{T_\mathrm{start}}{T}$ of the diffusion process for our DVCEs for the non-robust Swin-TF using the cone projection.%
    }
     \end{figure*}

\subsection{Diversity}\label{app:diversity}
\begin{figure*}[h]
     \centering
     \small
     \begin{tabular}{c|ccc||c|ccc}
     \hline
     \multicolumn{1}{c|}{Original}&
      Seed 1 & Seed 2 & Seed 3 &
     \multicolumn{1}{c|}{Original}&
     Seed 1 & Seed 2 & Seed 3\\
     \hline
     
     \begin{subfigure}{0.118\textwidth}\centering
     \caption*{\scriptsize \makecell{kite:\\ 0.88}}
     \includegraphics[width=1\textwidth]{images/swin/bird of prey raptor raptorial bird_kite_5_1096/original.png}
     \end{subfigure} &
      \begin{subfigure}{0.118\textwidth}\centering
      \caption*{\scriptsize \makecell{bald eagle:\\0.98}}
     \includegraphics[width=1\textwidth]{images/swin_seed_1_0_15/0.png}
     \end{subfigure} &
      \begin{subfigure}{0.118\textwidth}\centering
      \caption*{\scriptsize \makecell{bald eagle:\\0.99}}
     \includegraphics[width=1\textwidth]{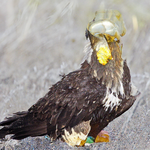}
     \end{subfigure} &
      \begin{subfigure}{0.118\textwidth}\centering
      \caption*{\scriptsize \makecell{bald eagle:\\0.97}}
     \includegraphics[width=1\textwidth]{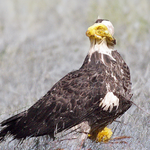}
     \end{subfigure} &

     \begin{subfigure}{0.118\textwidth}\centering
     \caption*{\scriptsize \makecell{jellyfish:\\ 0.96}}
     \includegraphics[width=1\textwidth]{images/swin/coelenterate cnidarian_jellyfish_6_5391/original.png}
     \end{subfigure} &
      \begin{subfigure}{0.118\textwidth}\centering
      \caption*{\scriptsize \makecell{sea anemone:\\ 0.98}}
       \includegraphics[width=1\textwidth]{images/swin_seed_1_0_15/1.png}
     \end{subfigure} &
      \begin{subfigure}{0.118\textwidth}\centering
      \caption*{\scriptsize \makecell{sea anemone:\\ 0.99}}
       \includegraphics[width=1\textwidth]{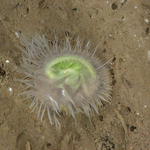}
     \end{subfigure} &
      \begin{subfigure}{0.118\textwidth}\centering
      \caption*{\scriptsize \makecell{sea anemone:\\ 0.98}}
       \includegraphics[width=1\textwidth]{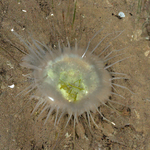}
     \end{subfigure}\\
     \hline
     \begin{subfigure}{0.118\textwidth}\centering
      \caption*{\scriptsize \makecell{Siamese cat:\\ 0.95}}
     \includegraphics[width=1\textwidth]{images/swin/domestic cat house cat Felis domesticus Felis catus_Siamese cat_10_14201/original.png}
     \end{subfigure} &
      \begin{subfigure}{0.118\textwidth}\centering
      \caption*{\scriptsize \makecell{Egyptian cat:\\ 0.99}}
      \includegraphics[width=1\textwidth]{images/swin_seed_1_0_15/2.png}
     \end{subfigure} &
      \begin{subfigure}{0.118\textwidth}\centering
      \caption*{\scriptsize \makecell{Egyptian cat:\\ 0.96}}
      \includegraphics[width=1\textwidth]{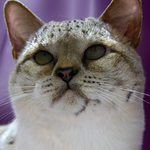}
     \end{subfigure} &
     \begin{subfigure}{0.118\textwidth}\centering
     \caption*{\scriptsize \makecell{Egyptian cat:\\ 0.98}}
     \includegraphics[width=1\textwidth]{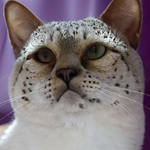}
     \end{subfigure} &

     \begin{subfigure}{0.118\textwidth}\centering
      \caption*{\scriptsize \makecell{house finch:\\ 0.96}}
     \includegraphics[width=1\textwidth]{images/swin/finch_house finch_12_649/original.png}
     \end{subfigure} &
      \begin{subfigure}{0.118\textwidth}\centering
      \caption*{\scriptsize \makecell{junco:\\ 0.98}}
       \includegraphics[width=1\textwidth]{images/swin_seed_1_0_15/3.png}
     \end{subfigure} &
      \begin{subfigure}{0.118\textwidth}\centering
      \caption*{\scriptsize \makecell{junco:\\ 0.97}}
       \includegraphics[width=1\textwidth]{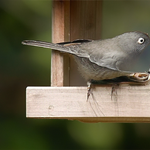}
    \end{subfigure} &
   \begin{subfigure}{0.118\textwidth}\centering
   \caption*{\scriptsize \makecell{junco:\\ 0.98}}
    \includegraphics[width=1\textwidth]{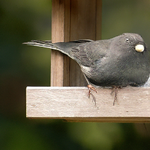}
     \end{subfigure}\\

     \end{tabular}

      \caption{\label{fig:seed_comp}DVCEs for the non-robust Swin-TF using the cone projection %
       across $3$ different seeds for the standard parameters used in the paper. The DVCEs for different seeds show subtle variations of the generated images and satisfy the desired properties of VCEs introduced in Sec. \ref{sec:DVCEs_properties}.
    }
     \end{figure*}

Note that the diffusion process is by design stochastic. This means that, unlike optimization-based VCEs, it is possible to generate a diverse set of VCEs from the same starting image. In Fig. \ref{fig:seed_comp}, we visualize different DVCEs obtained for the non-robust Swin-TF using the cone projection.%

\subsection{Different robust models}\label{app:different_robust_models}
\begin{figure*}[ht!]
     \centering
     \small
     \begin{tabular}{c|ccc|ccc}
     \hline
     \multicolumn{1}{c|}{Original}&
     \multicolumn{3}{c|}{Target Class 1} &    
     \multicolumn{3}{c}{Target Class 2}\\
     \hline
     & \scriptsize \madryft & \scriptsize \xcitft & \scriptsize \deitft &
       \scriptsize \madryft & \scriptsize \xcitft & \scriptsize \deitft\\
     \hline

      \begin{subfigure}{0.136\textwidth}\centering
     \caption*{\scriptsize \makecell{goldfish\\ \quad}}
     \includegraphics[width=1\textwidth]{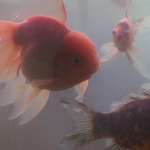}
     \end{subfigure} &

     \begin{subfigure}{0.136\textwidth}\centering
     \caption*{\scriptsize \makecell{lion fish:\\1.00}}
     \includegraphics[width=1\textwidth]{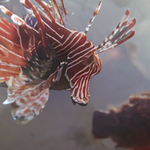}
     \end{subfigure} &
     \begin{subfigure}{0.136\textwidth}\centering
     \caption*{\scriptsize \makecell{lion fish:\\1.00}}
     \includegraphics[width=1\textwidth]{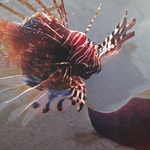}
     \end{subfigure} &
     \begin{subfigure}{0.136\textwidth}\centering
     \caption*{\scriptsize \makecell{lion fish:\\1.00}}
     \includegraphics[width=1\textwidth]{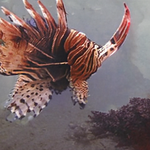}
     \end{subfigure} &

     \begin{subfigure}{0.136\textwidth}\centering
     \caption*{\scriptsize \makecell{anemone fish:\\1.00}}
     \includegraphics[width=1\textwidth]{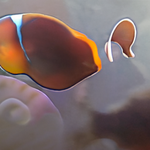}
     \end{subfigure} &
      \begin{subfigure}{0.136\textwidth}\centering
     \caption*{\scriptsize \makecell{anemone fish:\\1.00}}
     \includegraphics[width=1\textwidth]{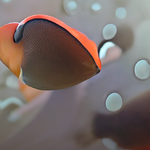}
     \end{subfigure} &
      \begin{subfigure}{0.136\textwidth}\centering
     \caption*{\scriptsize \makecell{anemone fish:\\1.00}}
     \includegraphics[width=1\textwidth]{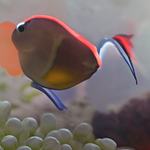}
     \end{subfigure}\\

     \begin{subfigure}{0.136\textwidth}\centering
     \caption*{\scriptsize \makecell{bullfrog\\ \quad}}
     \includegraphics[width=1\textwidth]{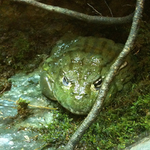}
     \end{subfigure} &

     \begin{subfigure}{0.136\textwidth}\centering
     \caption*{\scriptsize \makecell{tailed frog:\\1.00}}
     \includegraphics[width=1\textwidth]{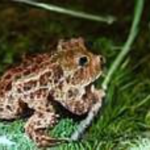}
     \end{subfigure} &

     \begin{subfigure}{0.136\textwidth}\centering
     \caption*{\scriptsize \makecell{tailed frog:\\1.00}}
     \includegraphics[width=1\textwidth]{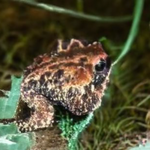}
     \end{subfigure} &

     \begin{subfigure}{0.136\textwidth}\centering
     \caption*{\scriptsize \makecell{tailed frog:\\1.00}}
     \includegraphics[width=1\textwidth]{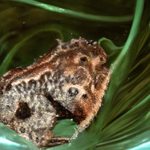}
     \end{subfigure} &
    
      \begin{subfigure}{0.136\textwidth}\centering
     \caption*{\scriptsize \makecell{axolotl:\\1.00}}
     \includegraphics[width=1\textwidth]{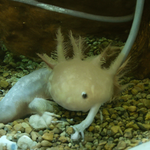}
     \end{subfigure} &
      \begin{subfigure}{0.136\textwidth}\centering
     \caption*{\scriptsize \makecell{axolotl:\\1.00}}
     \includegraphics[width=1\textwidth]{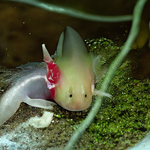}
     \end{subfigure} &
      \begin{subfigure}{0.136\textwidth}\centering
     \caption*{\scriptsize \makecell{axolotl:\\1.00}}
     \includegraphics[width=1\textwidth]{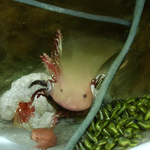}
     \end{subfigure}\\

      \begin{subfigure}{0.136\textwidth}\centering
     \caption*{\scriptsize \makecell{red wolf\\ \quad}}
     \includegraphics[width=1\textwidth]{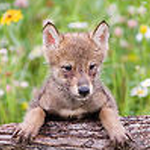}
     \end{subfigure} &

     \begin{subfigure}{0.136\textwidth}\centering
     \caption*{\scriptsize \makecell{coyote:\\1.00}}
     \includegraphics[width=1\textwidth]{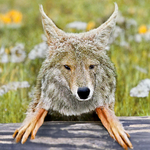}
     \end{subfigure} &
     \begin{subfigure}{0.136\textwidth}\centering
     \caption*{\scriptsize \makecell{coyote:\\1.00}}
     \includegraphics[width=1\textwidth]{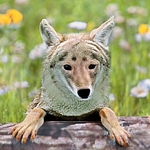}
     \end{subfigure} &

     \begin{subfigure}{0.136\textwidth}\centering
     \caption*{\scriptsize \makecell{coyote:\\1.00}}
     \includegraphics[width=1\textwidth]{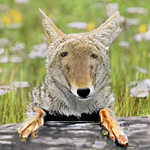}
     \end{subfigure} &

          \begin{subfigure}{0.136\textwidth}\centering
     \caption*{\scriptsize \makecell{timber wolf:\\0.99}}
     \includegraphics[width=1\textwidth]{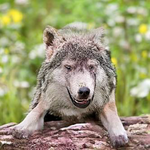}
     \end{subfigure} &
      \begin{subfigure}{0.136\textwidth}\centering
     \caption*{\scriptsize \makecell{timber wolf:\\0.98}}
     \includegraphics[width=1\textwidth]{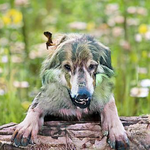}
     \end{subfigure} &
      \begin{subfigure}{0.136\textwidth}\centering
     \caption*{\scriptsize \makecell{timber wolf:\\0.98}}
     \includegraphics[width=1\textwidth]{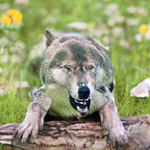}
     \end{subfigure}\\
     
      \begin{subfigure}{0.136\textwidth}\centering
     \caption*{\scriptsize \makecell{pirate ship\\ \quad}}
     \includegraphics[width=1\textwidth]{images/robust_model_comp/madry/36200/original.png}
     \end{subfigure} &

     \begin{subfigure}{0.136\textwidth}\centering
     \caption*{\scriptsize \makecell{liner ship:\\1.00}}
     \includegraphics[width=1\textwidth]{images/robust_model_comp/madry/36200/target_0_radius_0.png}
     \end{subfigure} &
     \begin{subfigure}{0.136\textwidth}\centering
     \caption*{\scriptsize \makecell{liner ship:\\1.00}}
     \includegraphics[width=1\textwidth]{images/robust_model_comp/xcit/36200/target_0_radius_0.png}
     \end{subfigure} &
     \begin{subfigure}{0.136\textwidth}\centering
     \caption*{\scriptsize \makecell{liner ship:\\1.00}}
     \includegraphics[width=1\textwidth]{images/robust_model_comp/deit/36200/target_0_radius_0.png}
     \end{subfigure} &

    \begin{subfigure}{0.136\textwidth}\centering
     \caption*{\scriptsize \makecell{container ship:\\1.00}}
     \includegraphics[width=1\textwidth]{images/robust_model_comp/madry/36200/target_1_radius_0.png}
     \end{subfigure} &
      \begin{subfigure}{0.136\textwidth}\centering
     \caption*{\scriptsize \makecell{container ship:\\1.00}}
     \includegraphics[width=1\textwidth]{images/robust_model_comp/xcit/36200/target_1_radius_0.png}
     \end{subfigure} &
      \begin{subfigure}{0.136\textwidth}\centering
     \caption*{\scriptsize \makecell{container ship:\\1.00}}
     \includegraphics[width=1\textwidth]{images/robust_model_comp/deit/36200/target_1_radius_0.png}
     \end{subfigure}\\

      \begin{subfigure}{0.136\textwidth}\centering
     \caption*{\scriptsize \makecell{mashed potato\\ \quad}}
     \includegraphics[width=1\textwidth]{images/robust_model_comp/madry/46795/original.png}
     \end{subfigure} &

     \begin{subfigure}{0.136\textwidth}\centering
     \caption*{\scriptsize \makecell{dough:\\1.00}}
     \includegraphics[width=1\textwidth]{images/robust_model_comp/madry/46795/target_0_radius_0.png}
     \end{subfigure} &

     \begin{subfigure}{0.136\textwidth}\centering
     \caption*{\scriptsize \makecell{dough:\\1.00}}
     \includegraphics[width=1\textwidth]{images/robust_model_comp/xcit/46795/target_0_radius_0.png}
     \end{subfigure} &

     \begin{subfigure}{0.136\textwidth}\centering
     \caption*{\scriptsize \makecell{dough:\\1.00}}
     \includegraphics[width=1\textwidth]{images/robust_model_comp/deit/46795/target_0_radius_0.png}
     \end{subfigure} &
    
          \begin{subfigure}{0.136\textwidth}\centering
     \caption*{\scriptsize \makecell{carbonara:\\1.00}}
     \includegraphics[width=1\textwidth]{images/robust_model_comp/madry/46795/target_1_radius_0.png}
     \end{subfigure} &
      \begin{subfigure}{0.136\textwidth}\centering
     \caption*{\scriptsize \makecell{carbonara:\\1.00}}
     \includegraphics[width=1\textwidth]{images/robust_model_comp/xcit/46795/target_1_radius_0.png}
     \end{subfigure} &
      \begin{subfigure}{0.136\textwidth}\centering
     \caption*{\scriptsize \makecell{carbonara:\\1.00}}
     \includegraphics[width=1\textwidth]{images/robust_model_comp/deit/46795/target_1_radius_0.png}
     \end{subfigure}\\
     
      \begin{subfigure}{0.136\textwidth}\centering
     \caption*{\scriptsize \makecell{head cabbage\\ \quad}}
     \includegraphics[width=1\textwidth]{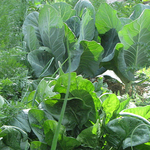}
     \end{subfigure} &

     \begin{subfigure}{0.136\textwidth}\centering
     \caption*{\scriptsize \makecell{broccoli:\\1.00}}
     \includegraphics[width=1\textwidth]{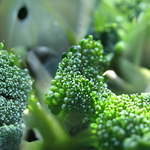}
     \end{subfigure} &

     \begin{subfigure}{0.136\textwidth}\centering
     \caption*{\scriptsize \makecell{broccoli:\\1.00}}
     \includegraphics[width=1\textwidth]{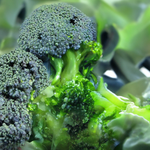}
     \end{subfigure} &

     \begin{subfigure}{0.136\textwidth}\centering
     \caption*{\scriptsize \makecell{broccoli:\\1.00}}
     \includegraphics[width=1\textwidth]{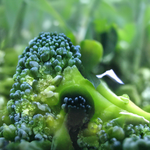}
     \end{subfigure} &

      \begin{subfigure}{0.136\textwidth}\centering
     \caption*{\scriptsize \makecell{cauliflower:\\1.00}}
     \includegraphics[width=1\textwidth]{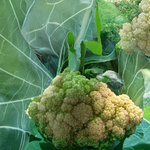}
     \end{subfigure} &
      \begin{subfigure}{0.136\textwidth}\centering
     \caption*{\scriptsize \makecell{cauliflower:\\1.00}}
     \includegraphics[width=1\textwidth]{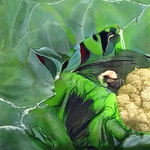}
     \end{subfigure} &
      \begin{subfigure}{0.136\textwidth}\centering
     \caption*{\scriptsize \makecell{cauliflower:\\1.00}}
     \includegraphics[width=1\textwidth]{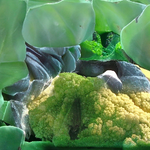}
     \end{subfigure}\\

      \begin{subfigure}{0.136\textwidth}\centering
     \caption*{\scriptsize \makecell{burrito\\ \quad}}
     \includegraphics[width=1\textwidth]{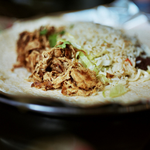}
     \end{subfigure} &

     \begin{subfigure}{0.136\textwidth}\centering
     \caption*{\scriptsize \makecell{pizza:\\1.00}}
     \includegraphics[width=1\textwidth]{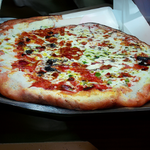}
     \end{subfigure} &

     \begin{subfigure}{0.136\textwidth}\centering
     \caption*{\scriptsize \makecell{pizza:\\1.00}}
     \includegraphics[width=1\textwidth]{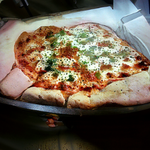}
     \end{subfigure} &

     \begin{subfigure}{0.136\textwidth}\centering
     \caption*{\scriptsize \makecell{pizza:\\1.00}}
     \includegraphics[width=1\textwidth]{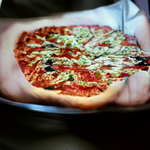}
     \end{subfigure} &

      \begin{subfigure}{0.136\textwidth}\centering
     \caption*{\scriptsize \makecell{cheeseburger:\\1.00}}
     \includegraphics[width=1\textwidth]{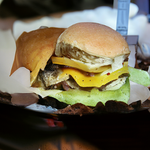}
     \end{subfigure} &
      \begin{subfigure}{0.136\textwidth}\centering
     \caption*{\scriptsize \makecell{cheeseburger:\\1.00}}
     \includegraphics[width=1\textwidth]{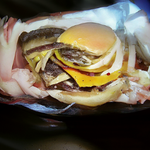}
     \end{subfigure} &
      \begin{subfigure}{0.136\textwidth}\centering
     \caption*{\scriptsize \makecell{cheeseburger:\\1.00}}
     \includegraphics[width=1\textwidth]{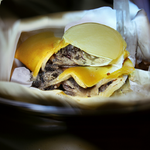}
     \end{subfigure} \\

     \begin{subfigure}{0.136\textwidth}\centering
     \caption*{\scriptsize \makecell{coral reef\\ \quad}}
     \includegraphics[width=1\textwidth]{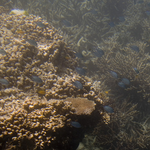}
     \end{subfigure} &
     
     \begin{subfigure}{0.136\textwidth}\centering
     \caption*{\scriptsize \makecell{alp:\\1.00}}
     \includegraphics[width=1\textwidth]{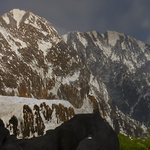}
     \end{subfigure} &
     \begin{subfigure}{0.136\textwidth}\centering
     \caption*{\scriptsize \makecell{alp:\\0.99}}
     \includegraphics[width=1\textwidth]{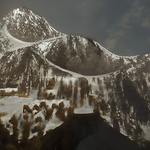}
     \end{subfigure} &
      \begin{subfigure}{0.136\textwidth}\centering
     \caption*{\scriptsize \makecell{alp:\\1.00}}
     \includegraphics[width=1\textwidth]{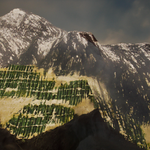}
     \end{subfigure} &

       \begin{subfigure}{0.136\textwidth}\centering
     \caption*{\scriptsize \makecell{cliff:\\0.99}}
     \includegraphics[width=1\textwidth]{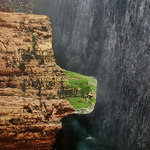}
     \end{subfigure} &
      \begin{subfigure}{0.136\textwidth}\centering
     \caption*{\scriptsize \makecell{cliff:\\0.98}}
     \includegraphics[width=1\textwidth]{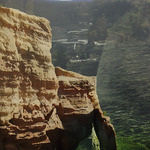}
     \end{subfigure} &
           \begin{subfigure}{0.136\textwidth}\centering
     \caption*{\scriptsize \makecell{cliff:\\0.99}}
     \includegraphics[width=1\textwidth]{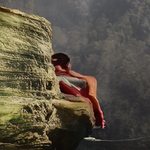}
     \end{subfigure}

     \\

     \end{tabular}

      \caption{\label{fig:app_robust_comp}
      We compare DVCEs for three different robust models (no cone projection) which are all fine-tuned to be multiple-norm adversarially robust \protect\cite{croce2021adversarial}, that is against$l_1$, $l_2$ and $l_\infty$-perturbations.}
 \end{figure*}

In this section we first show more examples in Fig. \ref{fig:app_robust_comp} for the qualitative comparison of different robust models described in Sec. \ref{subsec:model_comp}. 

\begin{figure*}[ht!]
     \centering
     \small
     \begin{tabular}{c|c|ccc}
     \hline
     Original & \quad \quad & \scriptsize \madryft & \scriptsize \xcitft & \scriptsize \deitft\\
     \hline

      \begin{subfigure}{0.201\textwidth}\centering
     \caption*{\scriptsize \makecell{loggerhead\\ \quad}}
     \includegraphics[width=1\textwidth]{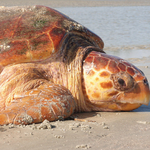}
     \end{subfigure} & 
     \rotatebox[origin=c]{90}{Swin-TF} &

     \begin{subfigure}{0.201\textwidth}\centering
     \caption*{\scriptsize \makecell{box turtle:\\0.98}}
     \includegraphics[width=1\textwidth]{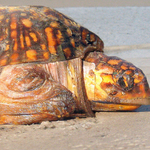}
     \end{subfigure} &
     \begin{subfigure}{0.201\textwidth}\centering
     \caption*{\scriptsize \makecell{box turtle:\\0.99}}
     \includegraphics[width=1\textwidth]{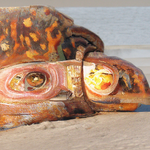}
     \end{subfigure} &
     \begin{subfigure}{0.201\textwidth}\centering
     \caption*{\scriptsize \makecell{box turtle:\\0.99}}
     \includegraphics[width=1\textwidth]{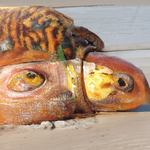}
     \end{subfigure}\\
     
     &
     \rotatebox[origin=c]{90}{ConvNeXt}
     &
     \begin{subfigure}{0.201\textwidth}\centering
     \caption*{\scriptsize \makecell{box turtle:\\0.98}}
     \includegraphics[width=1\textwidth]{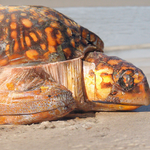}
     \end{subfigure} &
     \begin{subfigure}{0.201\textwidth}\centering
     \caption*{\scriptsize \makecell{box turtle:\\0.99}}
     \includegraphics[width=1\textwidth]{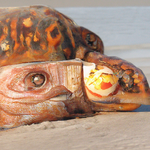}
     \end{subfigure} &
     \begin{subfigure}{0.201\textwidth}\centering
     \caption*{\scriptsize \makecell{box turtle:\\0.99}}
     \includegraphics[width=1\textwidth]{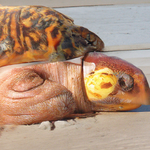}
     \end{subfigure}\\

     &
     \rotatebox[origin=c]{90}{EfficientNet}
     &
     \begin{subfigure}{0.201\textwidth}\centering
     \caption*{\scriptsize \makecell{box turtle:\\0.98}}
     \includegraphics[width=1\textwidth]{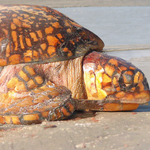}
     \end{subfigure} &
     \begin{subfigure}{0.201\textwidth}\centering
     \caption*{\scriptsize \makecell{box turtle:\\1.00}}
     \includegraphics[width=1\textwidth]{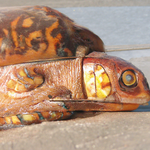}
     \end{subfigure} &
     \begin{subfigure}{0.201\textwidth}\centering
     \caption*{\scriptsize \makecell{box turtle:\\1.00}}
     \includegraphics[width=1\textwidth]{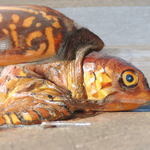}
     \end{subfigure}\\
     
     \hline
     Original & & \scriptsize \madryft & \scriptsize \xcitft & \scriptsize \deitft\\
     \hline

      \begin{subfigure}{0.201\textwidth}\centering
     \caption*{\scriptsize \makecell{goldfish\\ \quad}}
     \includegraphics[width=1\textwidth]{images/robust_model_comp/madry/82/original.png}
     \end{subfigure} & 
     \rotatebox[origin=c]{90}{Swin-TF} &

     \begin{subfigure}{0.201\textwidth}\centering
     \caption*{\scriptsize \makecell{lionfish:\\0.92}}
     \includegraphics[width=1\textwidth]{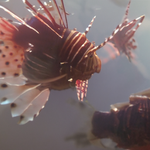}
     \end{subfigure} &
     \begin{subfigure}{0.201\textwidth}\centering
     \caption*{\scriptsize \makecell{lionfish:\\0.95}}
     \includegraphics[width=1\textwidth]{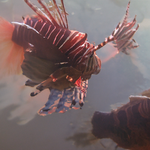}
     \end{subfigure} &
     \begin{subfigure}{0.201\textwidth}\centering
     \caption*{\scriptsize \makecell{lionfish:\\0.93}}
     \includegraphics[width=1\textwidth]{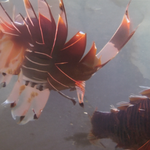}
     \end{subfigure}\\
     
     &
     \rotatebox[origin=c]{90}{ConvNeXt}
     &
     \begin{subfigure}{0.201\textwidth}\centering
     \caption*{\scriptsize \makecell{lionfish:\\0.98}}
     \includegraphics[width=1\textwidth]{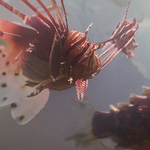}
     \end{subfigure} &
     \begin{subfigure}{0.201\textwidth}\centering
     \caption*{\scriptsize \makecell{lionfish:\\0.99}}
     \includegraphics[width=1\textwidth]{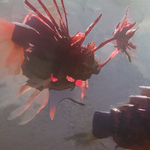}
     \end{subfigure} &
     \begin{subfigure}{0.201\textwidth}\centering
     \caption*{\scriptsize \makecell{lionfish:\\0.98}}
     \includegraphics[width=1\textwidth]{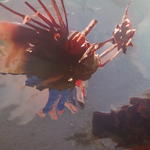}
     \end{subfigure}\\

     &
     \rotatebox[origin=c]{90}{EfficientNet}
     &
     \begin{subfigure}{0.201\textwidth}\centering
     \caption*{\scriptsize \makecell{lionfish:\\0.94}}
     \includegraphics[width=1\textwidth]{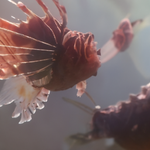}
     \end{subfigure} &
     \begin{subfigure}{0.201\textwidth}\centering
     \caption*{\scriptsize \makecell{lionfish:\\0.98}}
     \includegraphics[width=1\textwidth]{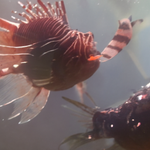}
     \end{subfigure} &
     \begin{subfigure}{0.201\textwidth}\centering
     \caption*{\scriptsize \makecell{lionfish:\\0.98}}
     \includegraphics[width=1\textwidth]{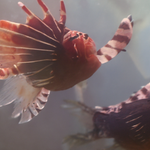}
     \end{subfigure}\\

     \end{tabular}

      \caption{\label{fig:app_non_robust_comp}
      We compare our DVCEs
      for the 
      non-robust classifiers used in Fig. \ref{fig:classifier_comp} (rows)
      where we now vary additionally the adversarially robust model which is used for regularization (same as in  Fig. \ref{fig:app_robust_comp} (columns)) for two different starting images. All non-robust classifiers show high confidence in the corresponding target classes and show meaningful target-specific changes. The variation of the DVCEs for the same classifier across different robust models used for the cone projection is small and on the level of the result of different seeds.}
 \end{figure*}

Further, in Fig. \ref{fig:app_non_robust_comp}, we compare cone projection of $3$ robust and $3$ non-robust models.


\clearpage
\subsection{Cone projection}\label{app:cone-projection-angle}
\begin{figure*}[ht!]
     \centering
     \small
     \begin{tabular}{c|ccc|c|cc}
     \hline
     \multicolumn{1}{c|}{Original}&
     \multicolumn{6}{c|}{Angles}\\
     \hline
     & 1\degree & 5\degree & 15\degree & \textbf{30}\degree & 40\degree & 50\degree  \\
     \hline

      \begin{subfigure}{0.136\textwidth}\centering
     \caption*{\scriptsize \makecell{chimpanzee\\ \quad}}
     \includegraphics[width=1\textwidth]{images/convnext/ape_18361/original.png}
     \end{subfigure} &

     \begin{subfigure}{0.136\textwidth}\centering
     \caption*{\scriptsize \makecell{orangutan:\\0.95}}
     \includegraphics[width=1\textwidth]{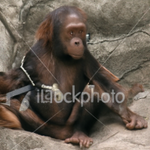}
     \end{subfigure} &
     \begin{subfigure}{0.136\textwidth}\centering
     \caption*{\scriptsize \makecell{orangutan:\\0.96}}
     \includegraphics[width=1\textwidth]{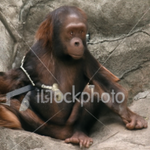}
     \end{subfigure} &
     \begin{subfigure}{0.136\textwidth}\centering
     \caption*{\scriptsize \makecell{orangutan:\\0.97}}
     \includegraphics[width=1\textwidth]{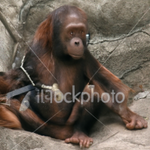}
     \end{subfigure} &
     \begin{subfigure}{0.136\textwidth}\centering
     \caption*{\scriptsize \makecell{orangutan:\\0.97}}
     \includegraphics[width=1\textwidth]{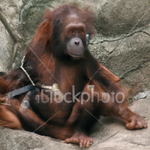}
     \end{subfigure} &
      \begin{subfigure}{0.136\textwidth}\centering
     \caption*{\scriptsize \makecell{orangutan:\\0.97}}
     \includegraphics[width=1\textwidth]{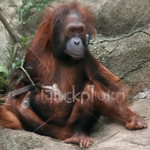}
     \end{subfigure} &
      \begin{subfigure}{0.136\textwidth}\centering
     \caption*{\scriptsize \makecell{orangutan:\\0.97}}
     \includegraphics[width=1\textwidth]{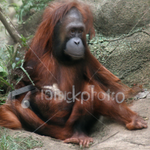}
     \end{subfigure}\\

      \begin{subfigure}{0.136\textwidth}\centering
     \caption*{\scriptsize \makecell{chesapeake bay\\retriever}}
     \includegraphics[width=1\textwidth]{images/convnext/retriever_10452/original.png}
     \end{subfigure} &

     \begin{subfigure}{0.136\textwidth}\centering
     \caption*{\scriptsize \makecell{golden retriever:\\0.01}}
     \includegraphics[width=1\textwidth]{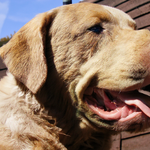}
     \end{subfigure} &
     \begin{subfigure}{0.136\textwidth}\centering
     \caption*{\scriptsize \makecell{golden retriever:\\0.02}}
     \includegraphics[width=1\textwidth]{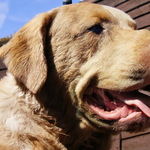}
     \end{subfigure} &
     \begin{subfigure}{0.136\textwidth}\centering
     \caption*{\scriptsize \makecell{golden retriever:\\0.39}}
     \includegraphics[width=1\textwidth]{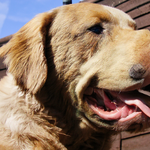}
     \end{subfigure} &
     \begin{subfigure}{0.136\textwidth}\centering
     \caption*{\scriptsize \makecell{golden retriever:\\0.84}}
     \includegraphics[width=1\textwidth]{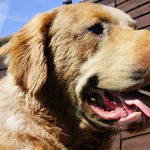}
     \end{subfigure} &
      \begin{subfigure}{0.136\textwidth}\centering
     \caption*{\scriptsize \makecell{golden retriever:\\0.93}}
     \includegraphics[width=1\textwidth]{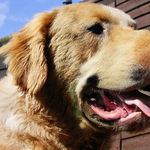}
     \end{subfigure} &
      \begin{subfigure}{0.136\textwidth}\centering
     \caption*{\scriptsize \makecell{golden retriever:\\0.96}}
     \includegraphics[width=1\textwidth]{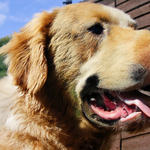}
     \end{subfigure}\\

      \begin{subfigure}{0.136\textwidth}\centering
     \caption*{\scriptsize \makecell{keeshond\\ \quad}}
     \includegraphics[width=1\textwidth]{images/madry/spitz_13068/original.png}
     \end{subfigure} &

     \begin{subfigure}{0.136\textwidth}\centering
     \caption*{\scriptsize \makecell{chow:\\0.98}}
     \includegraphics[width=1\textwidth]{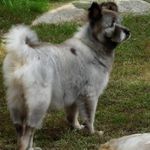}
     \end{subfigure} &
     \begin{subfigure}{0.136\textwidth}\centering
     \caption*{\scriptsize \makecell{chow:\\0.98}}
     \includegraphics[width=1\textwidth]{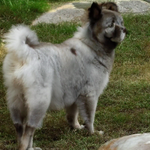}
     \end{subfigure} &
     \begin{subfigure}{0.136\textwidth}\centering
     \caption*{\scriptsize \makecell{chow:\\1.00}}
     \includegraphics[width=1\textwidth]{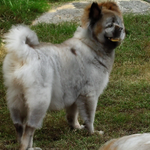}
     \end{subfigure} &
     \begin{subfigure}{0.136\textwidth}\centering
     \caption*{\scriptsize \makecell{chow:\\1.00}}
     \includegraphics[width=1\textwidth]{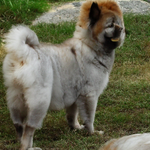}
     \end{subfigure} &
      \begin{subfigure}{0.136\textwidth}\centering
     \caption*{\scriptsize \makecell{chow:\\0.99}}
     \includegraphics[width=1\textwidth]{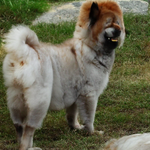}
     \end{subfigure} &
      \begin{subfigure}{0.136\textwidth}\centering
     \caption*{\scriptsize \makecell{chow:\\0.99}}
     \includegraphics[width=1\textwidth]{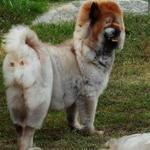}
     \end{subfigure}\\
     
      \begin{subfigure}{0.136\textwidth}\centering
     \caption*{\scriptsize \makecell{lynx\\ \quad}}
     \includegraphics[width=1\textwidth]{images/madry/feline_14352/original.png}
     \end{subfigure} &

     \begin{subfigure}{0.136\textwidth}\centering
     \caption*{\scriptsize \makecell{cheetah:\\0.97}}
     \includegraphics[width=1\textwidth]{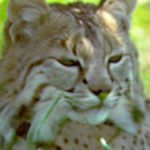}
     \end{subfigure} &
     \begin{subfigure}{0.136\textwidth}\centering
     \caption*{\scriptsize \makecell{cheetah:\\0.98}}
     \includegraphics[width=1\textwidth]{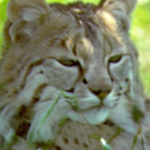}
     \end{subfigure} &
     \begin{subfigure}{0.136\textwidth}\centering
     \caption*{\scriptsize \makecell{cheetah:\\0.98}}
     \includegraphics[width=1\textwidth]{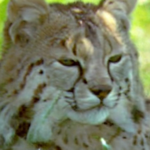}
     \end{subfigure} &
     \begin{subfigure}{0.136\textwidth}\centering
     \caption*{\scriptsize \makecell{cheetah:\\0.97}}
     \includegraphics[width=1\textwidth]{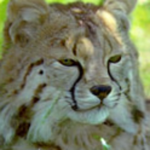}
     \end{subfigure} &
      \begin{subfigure}{0.136\textwidth}\centering
     \caption*{\scriptsize \makecell{cheetah:\\0.97}}
     \includegraphics[width=1\textwidth]{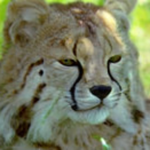}
     \end{subfigure} &
      \begin{subfigure}{0.136\textwidth}\centering
     \caption*{\scriptsize \makecell{cheetah:\\0.96}}
     \includegraphics[width=1\textwidth]{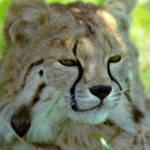}
     \end{subfigure}\\

      \begin{subfigure}{0.136\textwidth}\centering
     \caption*{\scriptsize \makecell{ladybug\\ \quad}}
     \includegraphics[width=1\textwidth]{images/madry/beetle_15069/original.png}
     \end{subfigure} &

     \begin{subfigure}{0.136\textwidth}\centering
     \caption*{\scriptsize \makecell{weevil:\\0.98}}
     \includegraphics[width=1\textwidth]{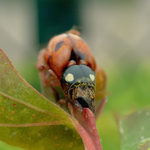}
     \end{subfigure} &
     \begin{subfigure}{0.136\textwidth}\centering
     \caption*{\scriptsize \makecell{weevil:\\0.98}}
     \includegraphics[width=1\textwidth]{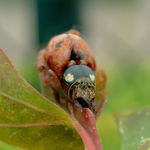}
     \end{subfigure} &
     \begin{subfigure}{0.136\textwidth}\centering
     \caption*{\scriptsize \makecell{weevil:\\0.98}}
     \includegraphics[width=1\textwidth]{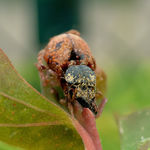}
     \end{subfigure} &
     \begin{subfigure}{0.136\textwidth}\centering
     \caption*{\scriptsize \makecell{weevil:\\0.98}}
     \includegraphics[width=1\textwidth]{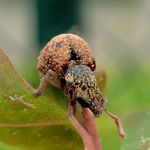}
     \end{subfigure} &
      \begin{subfigure}{0.136\textwidth}\centering
     \caption*{\scriptsize \makecell{weevil:\\0.98}}
     \includegraphics[width=1\textwidth]{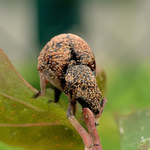}
     \end{subfigure} &
      \begin{subfigure}{0.136\textwidth}\centering
     \caption*{\scriptsize \makecell{weevil:\\0.97}}
     \includegraphics[width=1\textwidth]{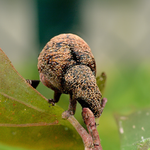}
     \end{subfigure}\\

      \begin{subfigure}{0.136\textwidth}\centering
     \caption*{\scriptsize \makecell{ringlet\\ \quad}}
     \includegraphics[width=1\textwidth]{images/madry/lepidopterous_16137/original.png}
     \end{subfigure} &

     \begin{subfigure}{0.136\textwidth}\centering
     \caption*{\scriptsize \makecell{monarch:\\0.71}}
     \includegraphics[width=1\textwidth]{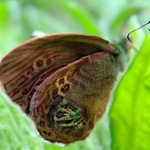}
     \end{subfigure} &
     \begin{subfigure}{0.136\textwidth}\centering
     \caption*{\scriptsize \makecell{monarch:\\0.48}}
     \includegraphics[width=1\textwidth]{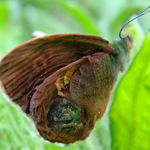}
     \end{subfigure} &
     \begin{subfigure}{0.136\textwidth}\centering
     \caption*{\scriptsize \makecell{monarch:\\0.97}}
     \includegraphics[width=1\textwidth]{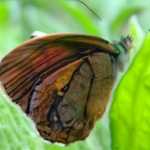}
     \end{subfigure} &
     \begin{subfigure}{0.136\textwidth}\centering
     \caption*{\scriptsize \makecell{monarch:\\0.96}}
     \includegraphics[width=1\textwidth]{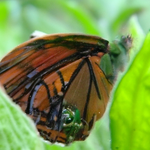}
     \end{subfigure} &
      \begin{subfigure}{0.136\textwidth}\centering
     \caption*{\scriptsize \makecell{monarch:\\0.97}}
     \includegraphics[width=1\textwidth]{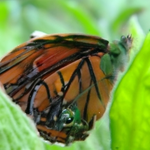}
     \end{subfigure} &
      \begin{subfigure}{0.136\textwidth}\centering
     \caption*{\scriptsize \makecell{monarch:\\0.97}}
     \includegraphics[width=1\textwidth]{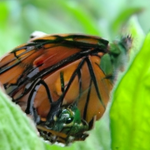}
     \end{subfigure}\\

      \begin{subfigure}{0.136\textwidth}\centering
      \caption*{\scriptsize \makecell{mashed potato\\ \quad}}
     \includegraphics[width=1\textwidth]{images/madry/foodstuff food product_mashed potato_2_46751/original.png}
     \end{subfigure} &

     \begin{subfigure}{0.136\textwidth}\centering
      \caption*{\scriptsize \makecell{carbonara\\0.99}}
     \includegraphics[width=1\textwidth]{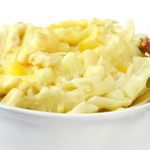}
     \end{subfigure} &
     \begin{subfigure}{0.136\textwidth}\centering
      \caption*{\scriptsize \makecell{carbonara\\0.99}}
     \includegraphics[width=1\textwidth]{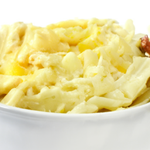}
     \end{subfigure} &
     \begin{subfigure}{0.136\textwidth}\centering
      \caption*{\scriptsize \makecell{carbonara\\0.99}}
     \includegraphics[width=1\textwidth]{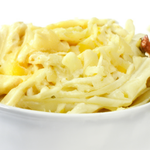}
     \end{subfigure} &
     \begin{subfigure}{0.136\textwidth}\centering
      \caption*{\scriptsize \makecell{carbonara\\0.98}}
     \includegraphics[width=1\textwidth]{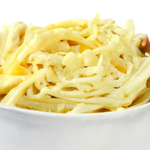}
     \end{subfigure} &
      \begin{subfigure}{0.136\textwidth}\centering
      \caption*{\scriptsize \makecell{carbonara\\0.98}}
     \includegraphics[width=1\textwidth]{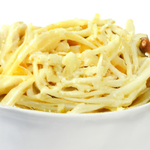}
     \end{subfigure} &
      \begin{subfigure}{0.136\textwidth}\centering
      \caption*{\scriptsize \makecell{carbonara\\0.98}}
     \includegraphics[width=1\textwidth]{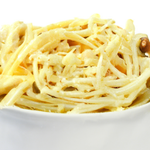}
     \end{subfigure}%

          \end{tabular}

      \caption{\label{fig:app_cone_angles}
      We compare different angles for the cone projection using the ConvNeXt model in combination with the \madryft robust model. For all images one observes that class-specific features (meaningful change) and high confidence (validity) are reached for more than $30\degree$. As the goal is to stick to the gradient of the non-robust classifier as much as possible we have fixed in the rest of the paper the angle of of the cone to $30\degree$.}
 \end{figure*}

In this section, we are going to compare different parameters for the cone projection from Sec. \ref{subsec:cone}, which allows us to explain non-robust classifiers. Remember that we project the gradient of the robust classifier onto a cone centered around the gradient of the target classifier. Thus, larger angles for the cone allow the method to deviate more and more from the target model gradient. In Fig. \ref{fig:app_cone_angles}, we show the resulting DVCEs for various angles between $1\degree$ and $50\degree$. As shown in Fig. \ref{fig:method_cone_projection}, the pure gradient of the target model is unable to visually guide the diffusion process, thus angles smaller than or equal to $15\degree$  often do not produce class-specific features in the resulting images. For angles of at least $30\degree$, we can see class-specific features in all images. Although it is possible to use even larger angles, we use $30\degree$ throughout the entire paper because this keeps the direction close to the target model's gradient, which is important as we want to explain the target model and not the robust one. 

\clearpage

\section{User study}\label{app:user_study}
In this section, we discuss a user study that we performed to compare the $l_{1.5}$-SVCEs \cite{boreiko2022sparse}, BDVCEs \cite{avrahami2021blended}, and our DVCEs.

Whereas in Fig. \ref{fig:method_comp} we provide a qualitative comparison showing the generated VCEs, the user study provides us also with a quantitative comparison. %
The images for DVCEs were generated the same as for Fig. \ref{fig:method_comp}, see Sec. \ref{subsec:method_comp} for details. However, for BDVCEs, maximizing confidence leads to images far away from the original one (images such as leopard, tiger, timber wolf, and white wolf in Fig. \ref{fig:method_comp}), thus we have selected one of the settings of the hyperparameters discussed in Sec. \ref{subsec:method_comp}, where the changes were looking the closest on average while introducing the meaningful features of the target classes. For $l_{1.5}$-SVCEs we have selected the highest radius, $r=150$, as this still leads to smaller changes (see Tab. \ref{tab:quantitative_eval}) in all the metrics, compared to DVCEs and BDVCEs, while maximizes the confidence.
To generate the images, we randomly selected images from the ImageNet test set and then generated VCEs for two target classes (different from the original class of the image) which belong same WordNet category as the original class. 
 
\begin{figure}[hbt!]
     \centering
     \includegraphics[width=1\textwidth]{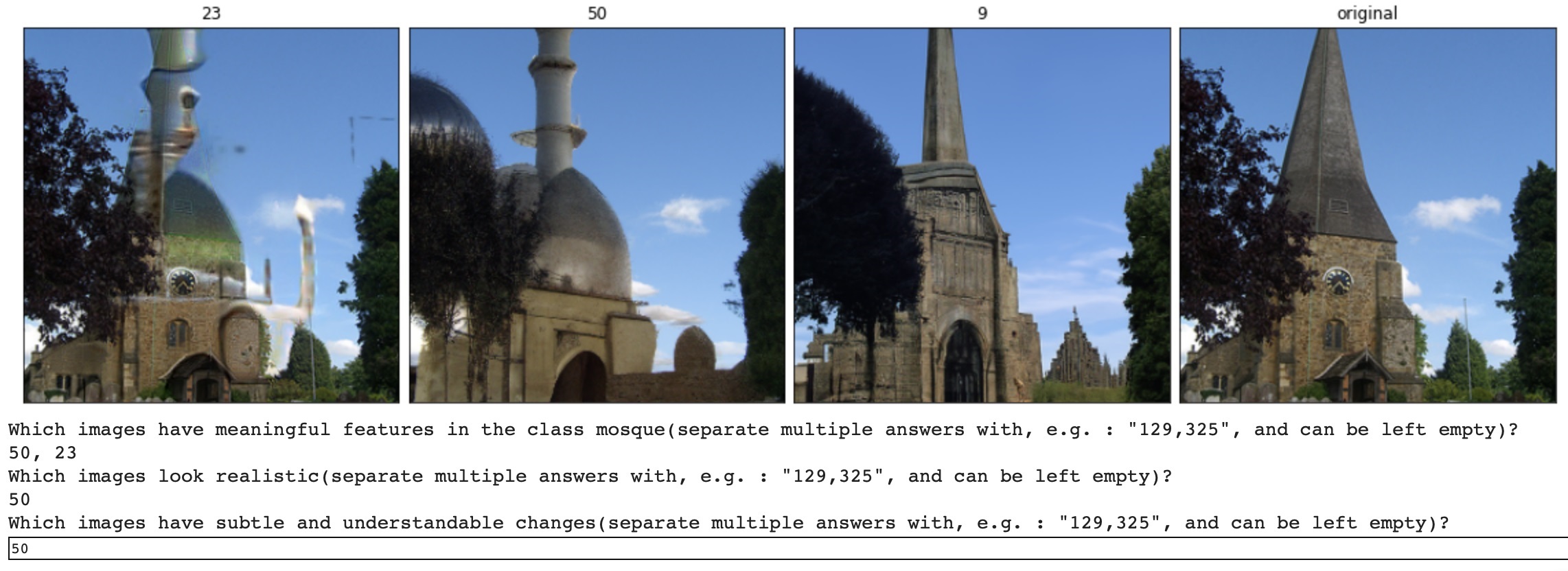}
      \caption{\label{fig:user_study_screenshot} Screenshot of the instructions for the user study. The original image is shown on the right, then the users are shown three VCEs and are asked three questions about the changes concerning the desired target class.}
     \end{figure}

In this study, the users, who participated voluntarily and without payment, were shown the original images and the VCEs of the three different methods in random order. The users were researchers in machine learning and related areas but none of them is working on VCEs themself or had seen the generated VCEs before. Following \cite{boreiko2022sparse}, we asked $20$ users to rate $44$ VCE, if the following three properties are satisfied for a given VCE (no or multiple
     answers are allowed): i) ``Which images have meaningful features in the target class?'' (\textbf{meaningful}), ii) ``Which images look realistic?'' \textbf{realism}, iii) ``Which images show subtle, yet understandable changes?'' (\textbf{subtle}). The screenshot with instructions and the shown images can be seen in Fig. \ref{fig:user_study_screenshot}.
     In the following we report the percentages of the images for DVCEs, $l_{1.5}$-SVCEs, and BDVCEs which were considered to have one of the three different properties: \textbf{meaningful} - \textbf{62.0\%}, 48.4\%, 38.7\%; \textbf{realism} - 34.7\%, 24.6\%, \textbf{52.2\%}; \textbf{subtle} -  45.0\%, \textbf{50.6\%}, 31.0\%. 
In Fig. \ref{fig:user_study_all} we report additionally all original images together with their changes and provide the percentages for each image individually.

Our DVCEs achieve with $62.0\%$ the highest percentage for meaningful changes, whereas BDVCEs achieve only $38.7\%$. However, BDVCEs are considered to be realistic in $52.2\%$ of the cases whereas our DVCEs in $34.7\%$. The reason for this seemingly contradictory result is that the BDVCEs are often not able to realize a meaningful class change, in the sense that the generated images do not show the corresponding class-specific feature e.g. in Fig. \ref{fig:user_study_all} for the change siamang $\rightarrow$ gorilla, the BDVCEs look very good (realism $80\%$) but are considered to be only $20\%$ meaningful (in fact they did not change the original image) or for jellyfish $\rightarrow$ brain coral the shown image looks realistic but shows no features of the target class. The reason is that it despite BD has the advantage that we select the image with the highest confidence from six parameter settings whereas for our method we only have one fixed parameter setting, it is sometimes not able to reach high confidence in the target class and does either too little changes (siamang $\rightarrow$ gorilla) or quite significant changes but meaningless ones (jellyfish $\rightarrow$ brain coral). Regarding $l_{1.5}$-SVCEs - they have the most subtle changes $50.6\%$ vs. $45.0\%$ for DVCEs but the images show often artefacts (please use zoom into images) which is the reason why they are considered the least realistic ones.  In total, the user study shows that DVCE performs best among the three methods if one considers all three categories. 

\section{Resources and hardware used}\label{app:hardware}
All the experiments were done on Tesla V100 GPUs, and for the generation of a batch of $6$ DVCEs without cone projection and blended VCEs $3$ minutes were required. For the batch of $6$ DVCEs with cone projections, $5$ minutes were required, as one additional model was loaded and the gradient with respect to it was calculated.

\begin{figure*}[hbt!]%
     \centering
     \small
     \begin{tabular}{c|ccc||c|ccc}
     \hline
      \scriptsize Original &
     \scriptsize DVCEs (ours) &  \scriptsize $l_{1.5}$-SVCEs \cite{boreiko2022sparse}&
       \scriptsize BDVCEs \cite{avrahami2021blended} &
     \scriptsize Original &
     \scriptsize DVCEs (ours) &  \scriptsize $l_{1.5}$-SVCEs \cite{boreiko2022sparse}&
       \scriptsize BDVCEs \cite{avrahami2021blended} \\
 \hline
      \begin{subfigure}{0.118\textwidth}\centering
     \caption*{\scriptsize puffer\\ $\rightarrow$goldfish}
     \includegraphics[width=1\textwidth]{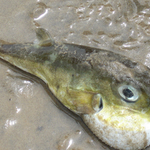}
     \end{subfigure} &
      \begin{subfigure}{0.118\textwidth}\centering
     \caption*{\scriptsize \\ m:0.6,r:0.1,s:0.5}
     \includegraphics[width=1\textwidth]{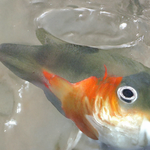}
     \end{subfigure} &
           \begin{subfigure}{0.118\textwidth}\centering
     \caption*{\scriptsize \\ m:0.6,r:0.2,s:0.5}
     \includegraphics[width=1\textwidth]{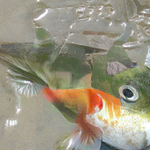}
     \end{subfigure} &
      \begin{subfigure}{0.118\textwidth}\centering
     \caption*{\scriptsize \\ m:0.4,r:0.1,s:0.2}
     \includegraphics[width=1\textwidth]{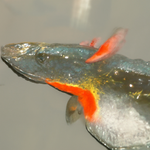}
     \end{subfigure} &
     
      \begin{subfigure}{0.118\textwidth}\centering
     \caption*{\scriptsize puffer\\ $\rightarrow$lionfish}
     \includegraphics[width=1\textwidth]{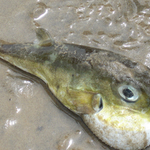}
     \end{subfigure} &
      \begin{subfigure}{0.118\textwidth}\centering
     \caption*{\scriptsize \\ m:0.8,r:0.3,s:0.6}
     \includegraphics[width=1\textwidth]{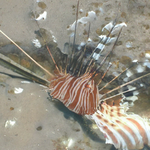}
     \end{subfigure} &
           \begin{subfigure}{0.118\textwidth}\centering
     \caption*{\scriptsize \\ m:0.7,r:0.1,s:0.8}
     \includegraphics[width=1\textwidth]{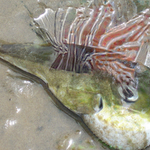}
     \end{subfigure} &
      \begin{subfigure}{0.118\textwidth}\centering
     \caption*{\scriptsize \\ m:0.1,r:0.3,s:0.2}
     \includegraphics[width=1\textwidth]{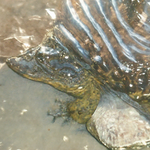}
     \end{subfigure} \\\hline
      \begin{subfigure}{0.118\textwidth}\centering
     \caption*{\scriptsize hartebeest\\ $\rightarrow$gazelle}
     \includegraphics[width=1\textwidth]{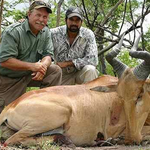}
     \end{subfigure} &
      \begin{subfigure}{0.118\textwidth}\centering
     \caption*{\scriptsize \\ m:0.2,r:0.1,s:0.1}
     \includegraphics[width=1\textwidth]{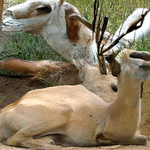}
     \end{subfigure} &
           \begin{subfigure}{0.118\textwidth}\centering
     \caption*{\scriptsize \\ m:0.1,r:0.1,s:0.3}
     \includegraphics[width=1\textwidth]{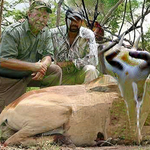}
     \end{subfigure} &
      \begin{subfigure}{0.118\textwidth}\centering
     \caption*{\scriptsize \\ m:0.0,r:0.1,s:0.1}
     \includegraphics[width=1\textwidth]{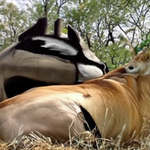}
     \end{subfigure} &
     
      \begin{subfigure}{0.118\textwidth}\centering
     \caption*{\scriptsize hartebeest\\ $\rightarrow$bison}
     \includegraphics[width=1\textwidth]{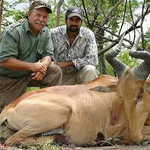}
     \end{subfigure} &
      \begin{subfigure}{0.118\textwidth}\centering
     \caption*{\scriptsize \\ m:0.1,r:0.0,s:0.1}
     \includegraphics[width=1\textwidth]{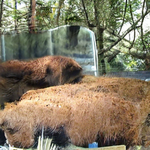}
     \end{subfigure} &
           \begin{subfigure}{0.118\textwidth}\centering
     \caption*{\scriptsize \\ m:0.1,r:0.1,s:0.2}
     \includegraphics[width=1\textwidth]{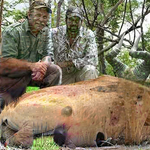}
     \end{subfigure} &
      \begin{subfigure}{0.118\textwidth}\centering
     \caption*{\scriptsize \\ m:0.3,r:0.8,s:0.3}
     \includegraphics[width=1\textwidth]{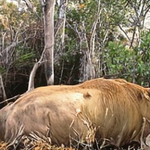}
     \end{subfigure} \\\hline
      \begin{subfigure}{0.118\textwidth}\centering
     \caption*{\scriptsize valley\\ $\rightarrow$volcano}
     \includegraphics[width=1\textwidth]{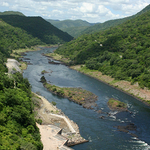}
     \end{subfigure} &
      \begin{subfigure}{0.118\textwidth}\centering
     \caption*{\scriptsize \\ m:0.7,r:0.3,s:0.3}
     \includegraphics[width=1\textwidth]{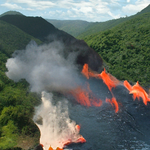}
     \end{subfigure} &
           \begin{subfigure}{0.118\textwidth}\centering
     \caption*{\scriptsize \\ m:0.8,r:0.4,s:0.6}
     \includegraphics[width=1\textwidth]{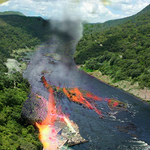}
     \end{subfigure} &
      \begin{subfigure}{0.118\textwidth}\centering
     \caption*{\scriptsize \\ m:0.3,r:0.5,s:0.5}
     \includegraphics[width=1\textwidth]{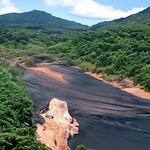}
     \end{subfigure} &
     
      \begin{subfigure}{0.118\textwidth}\centering
     \caption*{\scriptsize valley\\ $\rightarrow$alp}
     \includegraphics[width=1\textwidth]{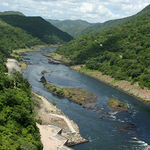}
     \end{subfigure} &
      \begin{subfigure}{0.118\textwidth}\centering
     \caption*{\scriptsize \\ m:0.8,r:0.5,s:0.4}
     \includegraphics[width=1\textwidth]{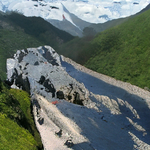}
     \end{subfigure} &
           \begin{subfigure}{0.118\textwidth}\centering
     \caption*{\scriptsize \\ m:0.3,r:0.3,s:0.2}
     \includegraphics[width=1\textwidth]{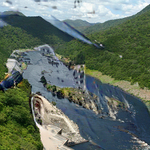}
     \end{subfigure} &
      \begin{subfigure}{0.118\textwidth}\centering
     \caption*{\scriptsize \\ m:0.2,r:0.6,s:0.5}
     \includegraphics[width=1\textwidth]{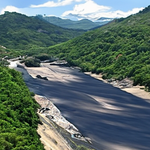}
     \end{subfigure} \\\hline
      \begin{subfigure}{0.118\textwidth}\centering
     \caption*{\scriptsize strawberry\\ $\rightarrow$pineapple}
     \includegraphics[width=1\textwidth]{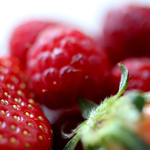}
     \end{subfigure} &
      \begin{subfigure}{0.118\textwidth}\centering
     \caption*{\scriptsize \\ m:0.7,r:0.4,s:0.6}
     \includegraphics[width=1\textwidth]{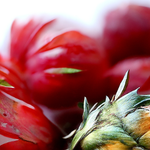}
     \end{subfigure} &
           \begin{subfigure}{0.118\textwidth}\centering
     \caption*{\scriptsize \\ m:0.6,r:0.1,s:0.6}
     \includegraphics[width=1\textwidth]{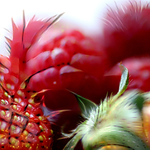}
     \end{subfigure} &
      \begin{subfigure}{0.118\textwidth}\centering
     \caption*{\scriptsize \\ m:0.0,r:0.6,s:0.2}
     \includegraphics[width=1\textwidth]{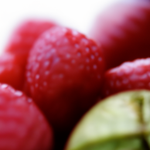}
     \end{subfigure} &
     
      \begin{subfigure}{0.118\textwidth}\centering
     \caption*{\scriptsize strawberry\\ $\rightarrow$pomegranate}
     \includegraphics[width=1\textwidth]{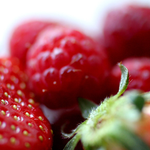}
     \end{subfigure} &
      \begin{subfigure}{0.118\textwidth}\centering
     \caption*{\scriptsize \\ m:0.3,r:0.1,s:0.4}
     \includegraphics[width=1\textwidth]{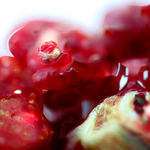}
     \end{subfigure} &
           \begin{subfigure}{0.118\textwidth}\centering
     \caption*{\scriptsize \\ m:0.5,r:0.1,s:0.4}
     \includegraphics[width=1\textwidth]{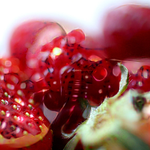}
     \end{subfigure} &
      \begin{subfigure}{0.118\textwidth}\centering
     \caption*{\scriptsize \\ m:0.1,r:0.6,s:0.3}
     \includegraphics[width=1\textwidth]{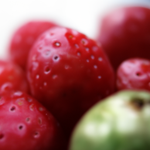}
     \end{subfigure} \\\hline
      \begin{subfigure}{0.118\textwidth}\centering
     \caption*{\scriptsize harvestman\\ $\rightarrow$b., gold spider}
     \includegraphics[width=1\textwidth]{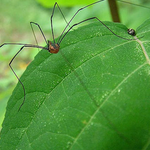}
     \end{subfigure} &
      \begin{subfigure}{0.118\textwidth}\centering
     \caption*{\scriptsize \\ m:0.2,r:0.1,s:0.2}
     \includegraphics[width=1\textwidth]{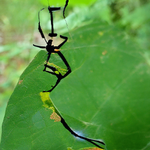}
     \end{subfigure} &
           \begin{subfigure}{0.118\textwidth}\centering
     \caption*{\scriptsize \\ m:0.4,r:0.4,s:0.6}
     \includegraphics[width=1\textwidth]{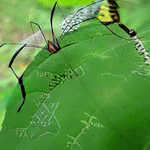}
     \end{subfigure} &
      \begin{subfigure}{0.118\textwidth}\centering
     \caption*{\scriptsize \\ m:0.1,r:0.5,s:0.1}
     \includegraphics[width=1\textwidth]{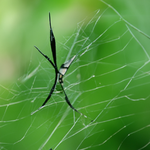}
     \end{subfigure} &
     
      \begin{subfigure}{0.118\textwidth}\centering
     \caption*{\scriptsize harvestman\\ $\rightarrow$tick}
     \includegraphics[width=1\textwidth]{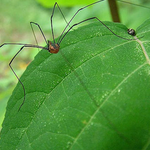}
     \end{subfigure} &
      \begin{subfigure}{0.118\textwidth}\centering
     \caption*{\scriptsize \\ m:0.0,r:0.1,s:0.1}
     \includegraphics[width=1\textwidth]{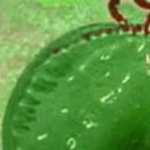}
     \end{subfigure} &
           \begin{subfigure}{0.118\textwidth}\centering
     \caption*{\scriptsize \\ m:0.1,r:0.1,s:0.3}
     \includegraphics[width=1\textwidth]{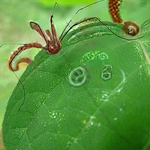}
     \end{subfigure} &
      \begin{subfigure}{0.118\textwidth}\centering
     \caption*{\scriptsize \\ m:0.6,r:0.6,s:0.2}
     \includegraphics[width=1\textwidth]{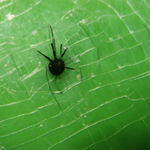}
     \end{subfigure} \\\hline
      \begin{subfigure}{0.118\textwidth}\centering
     \caption*{\scriptsize rugby ball\\ $\rightarrow$golf ball}
     \includegraphics[width=1\textwidth]{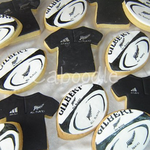}
     \end{subfigure} &
      \begin{subfigure}{0.118\textwidth}\centering
     \caption*{\scriptsize \\ m:0.1,r:0.1,s:0.2}
     \includegraphics[width=1\textwidth]{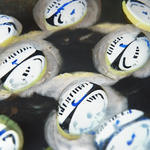}
     \end{subfigure} &
           \begin{subfigure}{0.118\textwidth}\centering
     \caption*{\scriptsize \\ m:0.1,r:0.3,s:0.3}
     \includegraphics[width=1\textwidth]{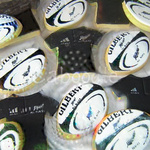}
     \end{subfigure} &
      \begin{subfigure}{0.118\textwidth}\centering
     \caption*{\scriptsize \\ m:0.1,r:0.4,s:0.2}
     \includegraphics[width=1\textwidth]{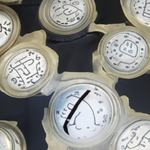}
     \end{subfigure} &
     
      \begin{subfigure}{0.118\textwidth}\centering
     \caption*{\scriptsize rugby ball\\ $\rightarrow$baseball}
     \includegraphics[width=1\textwidth]{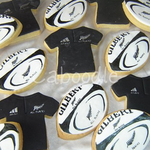}
     \end{subfigure} &
      \begin{subfigure}{0.118\textwidth}\centering
     \caption*{\scriptsize \\ m:0.5,r:0.2,s:0.6}
     \includegraphics[width=1\textwidth]{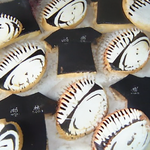}
     \end{subfigure} &
           \begin{subfigure}{0.118\textwidth}\centering
     \caption*{\scriptsize \\ m:0.5,r:0.1,s:0.6}
     \includegraphics[width=1\textwidth]{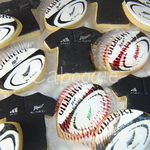}
     \end{subfigure} &
      \begin{subfigure}{0.118\textwidth}\centering
     \caption*{\scriptsize \\ m:0.5,r:0.6,s:0.2}
     \includegraphics[width=1\textwidth]{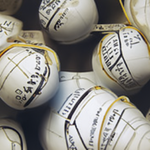}
     \end{subfigure} \\\hline
      \begin{subfigure}{0.118\textwidth}\centering
     \caption*{\scriptsize thunder snake\\ $\rightarrow$night snake}
     \includegraphics[width=1\textwidth]{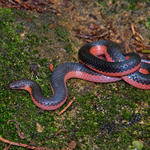}
     \end{subfigure} &
      \begin{subfigure}{0.118\textwidth}\centering
     \caption*{\scriptsize \\ m:0.8,r:0.8,s:0.4}
     \includegraphics[width=1\textwidth]{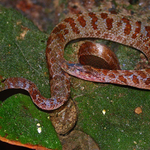}
     \end{subfigure} &
           \begin{subfigure}{0.118\textwidth}\centering
     \caption*{\scriptsize \\ m:0.4,r:0.4,s:0.6}
     \includegraphics[width=1\textwidth]{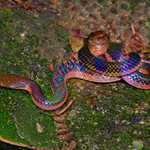}
     \end{subfigure} &
      \begin{subfigure}{0.118\textwidth}\centering
     \caption*{\scriptsize \\ m:1.0,r:0.9,s:0.2}
     \includegraphics[width=1\textwidth]{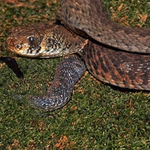}
     \end{subfigure} &
     
      \begin{subfigure}{0.118\textwidth}\centering
     \caption*{\scriptsize thunder snake\\ $\rightarrow$king snake}
     \includegraphics[width=1\textwidth]{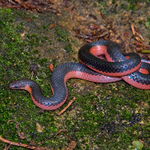}
     \end{subfigure} &
      \begin{subfigure}{0.118\textwidth}\centering
     \caption*{\scriptsize \\ m:0.8,r:0.3,s:0.3}
     \includegraphics[width=1\textwidth]{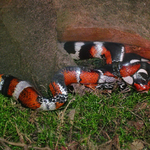}
     \end{subfigure} &
           \begin{subfigure}{0.118\textwidth}\centering
     \caption*{\scriptsize \\ m:0.8,r:0.4,s:0.7}
     \includegraphics[width=1\textwidth]{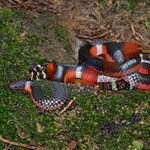}
     \end{subfigure} &
      \begin{subfigure}{0.118\textwidth}\centering
     \caption*{\scriptsize \\ m:0.5,r:1.0,s:0.3}
     \includegraphics[width=1\textwidth]{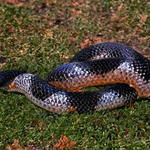}
     \end{subfigure} \\\hline
      \begin{subfigure}{0.118\textwidth}\centering
     \caption*{\scriptsize eft\\ $\rightarrow$tailed frog}
     \includegraphics[width=1\textwidth]{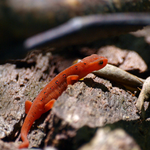}
     \end{subfigure} &
      \begin{subfigure}{0.118\textwidth}\centering
     \caption*{\scriptsize \\ m:0.6,r:0.3,s:0.3}
     \includegraphics[width=1\textwidth]{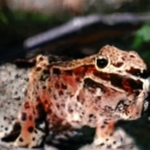}
     \end{subfigure} &
           \begin{subfigure}{0.118\textwidth}\centering
     \caption*{\scriptsize \\ m:0.5,r:0.1,s:0.5}
     \includegraphics[width=1\textwidth]{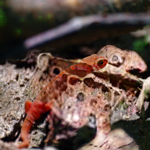}
     \end{subfigure} &
      \begin{subfigure}{0.118\textwidth}\centering
     \caption*{\scriptsize \\ m:0.3,r:0.2,s:0.0}
     \includegraphics[width=1\textwidth]{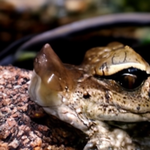}
     \end{subfigure} &
     
      \begin{subfigure}{0.118\textwidth}\centering
     \caption*{\scriptsize eft\\ $\rightarrow$axolotl}
     \includegraphics[width=1\textwidth]{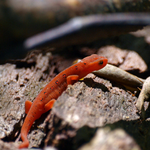}
     \end{subfigure} &
      \begin{subfigure}{0.118\textwidth}\centering
     \caption*{\scriptsize \\ m:0.5,r:0.1,s:0.4}
     \includegraphics[width=1\textwidth]{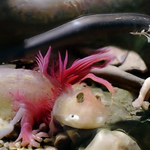}
     \end{subfigure} &
           \begin{subfigure}{0.118\textwidth}\centering
     \caption*{\scriptsize \\ m:0.6,r:0.4,s:0.5}
     \includegraphics[width=1\textwidth]{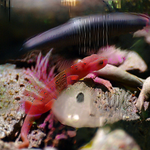}
     \end{subfigure} &
      \begin{subfigure}{0.118\textwidth}\centering
     \caption*{\scriptsize \\ m:0.1,r:0.3,s:0.2}
     \includegraphics[width=1\textwidth]{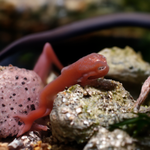}
     \end{subfigure} \\
    \end{tabular}
      
     \end{figure*}
     
     \begin{figure*}[ht]\ContinuedFloat
      \centering
     \small
     \begin{tabular}{c|ccc||c|ccc}
     \hline
      \scriptsize Original &
     \scriptsize DVCEs (ours) &  \scriptsize $l_{1.5}$-SVCEs \cite{boreiko2022sparse}&
       \scriptsize BDVCEs \cite{avrahami2021blended} &
     \scriptsize Original &
     \scriptsize DVCEs (ours) &  \scriptsize $l_{1.5}$-SVCEs \cite{boreiko2022sparse}&
       \scriptsize BDVCEs \cite{avrahami2021blended} \\
  \hline
      \begin{subfigure}{0.118\textwidth}\centering
     \caption*{\scriptsize leopard\\ $\rightarrow$snow leopard}
     \includegraphics[width=1\textwidth]{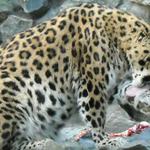}
     \end{subfigure} &
      \begin{subfigure}{0.118\textwidth}\centering
     \caption*{\scriptsize \\ m:0.8,r:0.1,s:0.4}
     \includegraphics[width=1\textwidth]{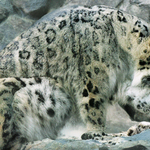}
     \end{subfigure} &
           \begin{subfigure}{0.118\textwidth}\centering
     \caption*{\scriptsize \\ m:0.8,r:0.5,s:0.6}
     \includegraphics[width=1\textwidth]{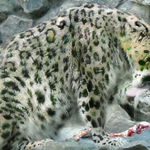}
     \end{subfigure} &
      \begin{subfigure}{0.118\textwidth}\centering
     \caption*{\scriptsize \\ m:0.9,r:0.8,s:0.9}
     \includegraphics[width=1\textwidth]{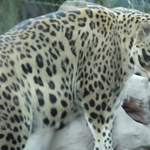}
     \end{subfigure} &
     
      \begin{subfigure}{0.118\textwidth}\centering
     \caption*{\scriptsize leopard\\ $\rightarrow$cheetah}
     \includegraphics[width=1\textwidth]{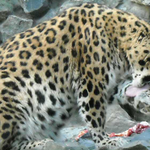}
     \end{subfigure} &
      \begin{subfigure}{0.118\textwidth}\centering
     \caption*{\scriptsize \\ m:0.9,r:0.2,s:0.7}
     \includegraphics[width=1\textwidth]{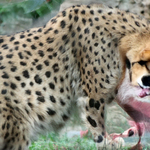}
     \end{subfigure} &
           \begin{subfigure}{0.118\textwidth}\centering
     \caption*{\scriptsize \\ m:0.4,r:0.1,s:0.2}
     \includegraphics[width=1\textwidth]{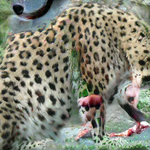}
     \end{subfigure} &
      \begin{subfigure}{0.118\textwidth}\centering
     \caption*{\scriptsize \\ m:0.7,r:0.7,s:0.4}
     \includegraphics[width=1\textwidth]{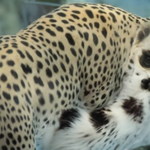}
     \end{subfigure} \\\hline
      \begin{subfigure}{0.118\textwidth}\centering
     \caption*{\scriptsize dung beetle\\ $\rightarrow$tiger beetle}
     \includegraphics[width=1\textwidth]{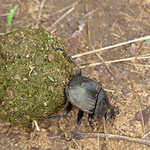}
     \end{subfigure} &
      \begin{subfigure}{0.118\textwidth}\centering
     \caption*{\scriptsize \\ m:0.7,r:0.8,s:0.5}
     \includegraphics[width=1\textwidth]{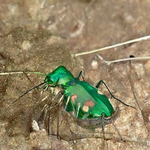}
     \end{subfigure} &
           \begin{subfigure}{0.118\textwidth}\centering
     \caption*{\scriptsize \\ m:0.5,r:0.2,s:0.5}
     \includegraphics[width=1\textwidth]{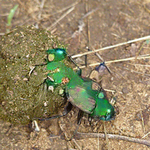}
     \end{subfigure} &
      \begin{subfigure}{0.118\textwidth}\centering
     \caption*{\scriptsize \\ m:0.8,r:0.8,s:0.5}
     \includegraphics[width=1\textwidth]{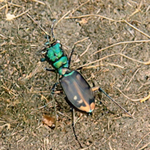}
     \end{subfigure} &
     
      \begin{subfigure}{0.118\textwidth}\centering
     \caption*{\scriptsize dung beetle\\ $\rightarrow$rhino. beetle}
     \includegraphics[width=1\textwidth]{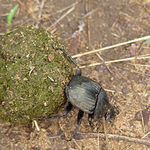}
     \end{subfigure} &
      \begin{subfigure}{0.118\textwidth}\centering
     \caption*{\scriptsize \\ m:0.5,r:0.1,s:0.2}
     \includegraphics[width=1\textwidth]{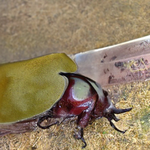}
     \end{subfigure} &
           \begin{subfigure}{0.118\textwidth}\centering
     \caption*{\scriptsize \\ m:0.8,r:0.5,s:0.8}
     \includegraphics[width=1\textwidth]{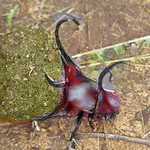}
     \end{subfigure} &
      \begin{subfigure}{0.118\textwidth}\centering
     \caption*{\scriptsize \\ m:0.2,r:0.2,s:0.0}
     \includegraphics[width=1\textwidth]{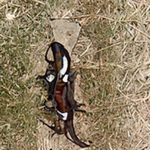}
     \end{subfigure} \\\hline
      \begin{subfigure}{0.118\textwidth}\centering
     \caption*{\scriptsize guinea pig\\ $\rightarrow$marmot}
     \includegraphics[width=1\textwidth]{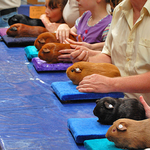}
     \end{subfigure} &
      \begin{subfigure}{0.118\textwidth}\centering
     \caption*{\scriptsize \\ m:0.6,r:0.2,s:0.3}
     \includegraphics[width=1\textwidth]{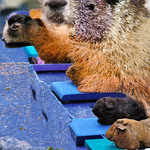}
     \end{subfigure} &
           \begin{subfigure}{0.118\textwidth}\centering
     \caption*{\scriptsize \\ m:0.2,r:0.1,s:0.3}
     \includegraphics[width=1\textwidth]{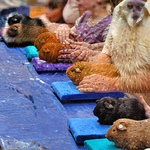}
     \end{subfigure} &
      \begin{subfigure}{0.118\textwidth}\centering
     \caption*{\scriptsize \\ m:0.8,r:0.6,s:0.2}
     \includegraphics[width=1\textwidth]{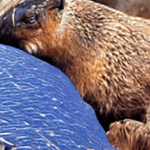}
     \end{subfigure} &
     
      \begin{subfigure}{0.118\textwidth}\centering
     \caption*{\scriptsize guinea pig\\ $\rightarrow$beaver}
     \includegraphics[width=1\textwidth]{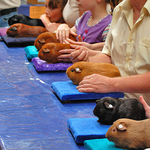}
     \end{subfigure} &
      \begin{subfigure}{0.118\textwidth}\centering
     \caption*{\scriptsize \\ m:0.5,r:0.0,s:0.1}
     \includegraphics[width=1\textwidth]{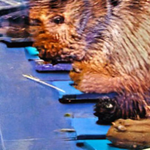}
     \end{subfigure} &
           \begin{subfigure}{0.118\textwidth}\centering
     \caption*{\scriptsize \\ m:0.2,r:0.1,s:0.3}
     \includegraphics[width=1\textwidth]{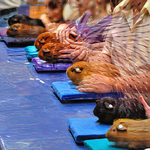}
     \end{subfigure} &
      \begin{subfigure}{0.118\textwidth}\centering
     \caption*{\scriptsize \\ m:0.3,r:0.0,s:0.1}
     \includegraphics[width=1\textwidth]{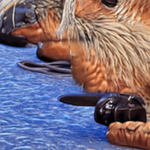}
     \end{subfigure} \\\hline
      \begin{subfigure}{0.118\textwidth}\centering
     \caption*{\scriptsize siamang\\ $\rightarrow$orangutan}
     \includegraphics[width=1\textwidth]{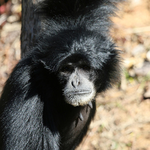}
     \end{subfigure} &
      \begin{subfigure}{0.118\textwidth}\centering
     \caption*{\scriptsize \\ m:0.7,r:0.1,s:0.4}
     \includegraphics[width=1\textwidth]{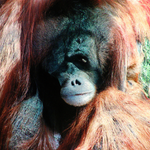}
     \end{subfigure} &
           \begin{subfigure}{0.118\textwidth}\centering
     \caption*{\scriptsize \\ m:0.4,r:0.1,s:0.3}
     \includegraphics[width=1\textwidth]{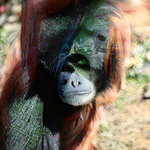}
     \end{subfigure} &
      \begin{subfigure}{0.118\textwidth}\centering
     \caption*{\scriptsize \\ m:0.6,r:0.8,s:0.6}
     \includegraphics[width=1\textwidth]{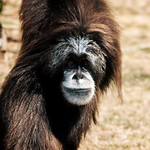}
     \end{subfigure} &
     
      \begin{subfigure}{0.118\textwidth}\centering
     \caption*{\scriptsize siamang\\ $\rightarrow$gorilla}
     \includegraphics[width=1\textwidth]{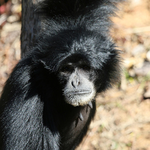}
     \end{subfigure} &
      \begin{subfigure}{0.118\textwidth}\centering
     \caption*{\scriptsize \\ m:0.7,r:0.1,s:0.5}
     \includegraphics[width=1\textwidth]{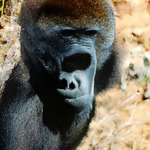}
     \end{subfigure} &
           \begin{subfigure}{0.118\textwidth}\centering
     \caption*{\scriptsize \\ m:0.7,r:0.0,s:0.3}
     \includegraphics[width=1\textwidth]{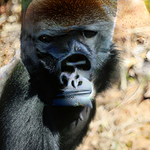}
     \end{subfigure} &
      \begin{subfigure}{0.118\textwidth}\centering
     \caption*{\scriptsize \\ m:0.2,r:0.8,s:0.5}
     \includegraphics[width=1\textwidth]{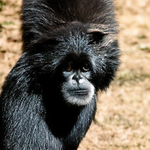}
     \end{subfigure} \\\hline
      \begin{subfigure}{0.118\textwidth}\centering
     \caption*{\scriptsize kit fox\\ $\rightarrow$red fox}
     \includegraphics[width=1\textwidth]{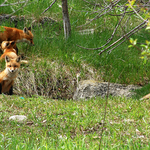}
     \end{subfigure} &
      \begin{subfigure}{0.118\textwidth}\centering
     \caption*{\scriptsize \\ m:0.6,r:0.1,s:0.2}
     \includegraphics[width=1\textwidth]{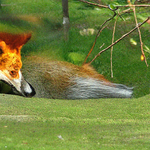}
     \end{subfigure} &
           \begin{subfigure}{0.118\textwidth}\centering
     \caption*{\scriptsize \\ m:0.3,r:0.0,s:0.2}
     \includegraphics[width=1\textwidth]{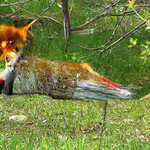}
     \end{subfigure} &
      \begin{subfigure}{0.118\textwidth}\centering
     \caption*{\scriptsize \\ m:0.8,r:0.7,s:0.1}
     \includegraphics[width=1\textwidth]{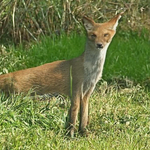}
     \end{subfigure} &
     
      \begin{subfigure}{0.118\textwidth}\centering
     \caption*{\scriptsize kit fox\\ $\rightarrow$Arctic fox}
     \includegraphics[width=1\textwidth]{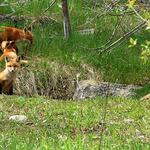}
     \end{subfigure} &
      \begin{subfigure}{0.118\textwidth}\centering
     \caption*{\scriptsize \\ m:0.5,r:0.1,s:0.2}
     \includegraphics[width=1\textwidth]{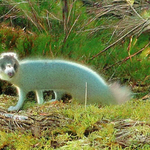}
     \end{subfigure} &
           \begin{subfigure}{0.118\textwidth}\centering
     \caption*{\scriptsize \\ m:0.8,r:0.5,s:0.8}
     \includegraphics[width=1\textwidth]{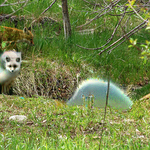}
     \end{subfigure} &
      \begin{subfigure}{0.118\textwidth}\centering
     \caption*{\scriptsize \\ m:0.0,r:0.0,s:0.0}
     \includegraphics[width=1\textwidth]{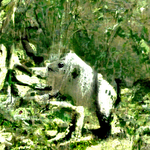}
     \end{subfigure} \\\hline
      \begin{subfigure}{0.118\textwidth}\centering
     \caption*{\scriptsize thatch\\ $\rightarrow$dome}
     \includegraphics[width=1\textwidth]{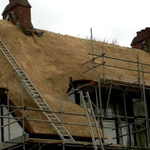}
     \end{subfigure} &
      \begin{subfigure}{0.118\textwidth}\centering
     \caption*{\scriptsize \\ m:0.8,r:0.6,s:0.7}
     \includegraphics[width=1\textwidth]{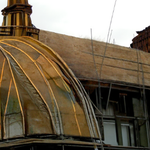}
     \end{subfigure} &
           \begin{subfigure}{0.118\textwidth}\centering
     \caption*{\scriptsize \\ m:0.2,r:0.2,s:0.4}
     \includegraphics[width=1\textwidth]{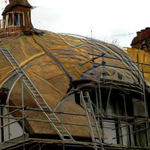}
     \end{subfigure} &
      \begin{subfigure}{0.118\textwidth}\centering
     \caption*{\scriptsize \\ m:0.8,r:0.8,s:0.7}
     \includegraphics[width=1\textwidth]{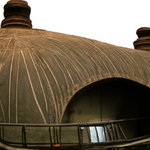}
     \end{subfigure} &
     
      \begin{subfigure}{0.118\textwidth}\centering
     \caption*{\scriptsize thatch\\ $\rightarrow$tile roof}
     \includegraphics[width=1\textwidth]{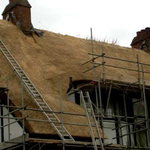}
     \end{subfigure} &
      \begin{subfigure}{0.118\textwidth}\centering
     \caption*{\scriptsize \\ m:0.8,r:0.3,s:0.6}
     \includegraphics[width=1\textwidth]{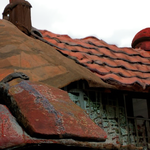}
     \end{subfigure} &
           \begin{subfigure}{0.118\textwidth}\centering
     \caption*{\scriptsize \\ m:0.7,r:0.1,s:0.6}
     \includegraphics[width=1\textwidth]{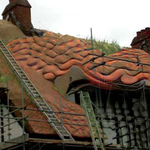}
     \end{subfigure} &
      \begin{subfigure}{0.118\textwidth}\centering
     \caption*{\scriptsize \\ m:0.9,r:0.8,s:0.8}
     \includegraphics[width=1\textwidth]{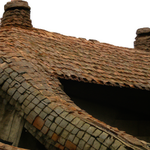}
     \end{subfigure} \\\hline
      \begin{subfigure}{0.118\textwidth}\centering
     \caption*{\scriptsize jellyfish\\ $\rightarrow$sea anemone}
     \includegraphics[width=1\textwidth]{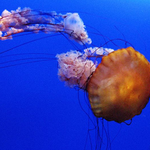}
     \end{subfigure} &
      \begin{subfigure}{0.118\textwidth}\centering
     \caption*{\scriptsize \\ m:0.8,r:0.5,s:0.7}
     \includegraphics[width=1\textwidth]{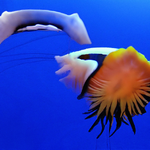}
     \end{subfigure} &
           \begin{subfigure}{0.118\textwidth}\centering
     \caption*{\scriptsize \\ m:0.5,r:0.5,s:0.6}
     \includegraphics[width=1\textwidth]{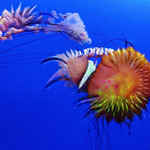}
     \end{subfigure} &
      \begin{subfigure}{0.118\textwidth}\centering
     \caption*{\scriptsize \\ m:0.3,r:0.4,s:0.2}
     \includegraphics[width=1\textwidth]{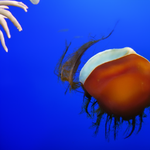}
     \end{subfigure} &
     
      \begin{subfigure}{0.118\textwidth}\centering
     \caption*{\scriptsize jellyfish\\ $\rightarrow$brain coral}
     \includegraphics[width=1\textwidth]{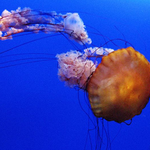}
     \end{subfigure} &
      \begin{subfigure}{0.118\textwidth}\centering
     \caption*{\scriptsize \\ m:1.0,r:0.7,s:0.8}
     \includegraphics[width=1\textwidth]{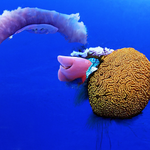}
     \end{subfigure} &
           \begin{subfigure}{0.118\textwidth}\centering
     \caption*{\scriptsize \\ m:0.5,r:0.4,s:0.8}
     \includegraphics[width=1\textwidth]{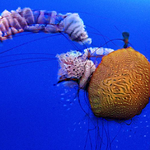}
     \end{subfigure} &
      \begin{subfigure}{0.118\textwidth}\centering
     \caption*{\scriptsize \\ m:0.1,r:0.6,s:0.1}
     \includegraphics[width=1\textwidth]{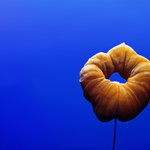}
     \end{subfigure} \\
     \end{tabular}
     \end{figure*}
     
     \begin{figure*}[ht]\ContinuedFloat
      \centering
     \small
     \begin{tabular}{c|ccc||c|ccc}
     \hline
      \scriptsize Original &
     \scriptsize DVCEs (ours) &  \scriptsize $l_{1.5}$-SVCEs \cite{boreiko2022sparse}&
       \scriptsize BDVCEs \cite{avrahami2021blended} &
     \scriptsize Original &
     \scriptsize DVCEs (ours) &  \scriptsize $l_{1.5}$-SVCEs \cite{boreiko2022sparse}&
       \scriptsize BDVCEs \cite{avrahami2021blended} \\
    \hline
      \begin{subfigure}{0.118\textwidth}\centering
     \caption*{\scriptsize church\\ $\rightarrow$mosque}
     \includegraphics[width=1\textwidth]{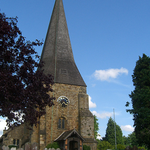}
     \end{subfigure} &
      \begin{subfigure}{0.118\textwidth}\centering
     \caption*{\scriptsize \\ m:0.9,r:0.5,s:0.5}
     \includegraphics[width=1\textwidth]{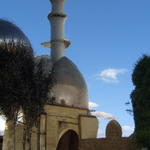}
     \end{subfigure} &
           \begin{subfigure}{0.118\textwidth}\centering
     \caption*{\scriptsize \\ m:0.3,r:0.1,s:0.3}
     \includegraphics[width=1\textwidth]{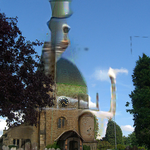}
     \end{subfigure} &
      \begin{subfigure}{0.118\textwidth}\centering
     \caption*{\scriptsize \\ m:0.1,r:0.5,s:0.5}
     \includegraphics[width=1\textwidth]{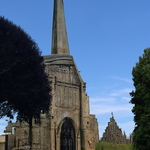}
     \end{subfigure} &
     
      \begin{subfigure}{0.118\textwidth}\centering
     \caption*{\scriptsize church\\ $\rightarrow$stupa}
     \includegraphics[width=1\textwidth]{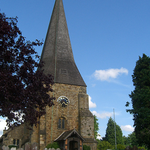}
     \end{subfigure} &
      \begin{subfigure}{0.118\textwidth}\centering
     \caption*{\scriptsize \\ m:0.6,r:0.3,s:0.5}
     \includegraphics[width=1\textwidth]{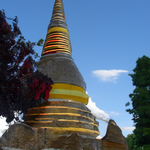}
     \end{subfigure} &
           \begin{subfigure}{0.118\textwidth}\centering
     \caption*{\scriptsize \\ m:0.3,r:0.0,s:0.3}
     \includegraphics[width=1\textwidth]{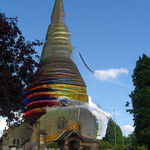}
     \end{subfigure} &
      \begin{subfigure}{0.118\textwidth}\centering
     \caption*{\scriptsize \\ m:0.5,r:0.8,s:0.6}
     \includegraphics[width=1\textwidth]{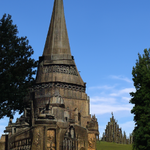}
     \end{subfigure} \\\hline
      \begin{subfigure}{0.118\textwidth}\centering
     \caption*{\scriptsize broccoli\\ $\rightarrow$cauliflower}
     \includegraphics[width=1\textwidth]{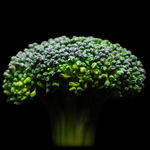}
     \end{subfigure} &
      \begin{subfigure}{0.118\textwidth}\centering
     \caption*{\scriptsize \\ m:0.9,r:0.9,s:0.7}
     \includegraphics[width=1\textwidth]{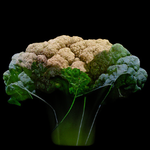}
     \end{subfigure} &
           \begin{subfigure}{0.118\textwidth}\centering
     \caption*{\scriptsize \\ m:0.3,r:0.5,s:0.6}
     \includegraphics[width=1\textwidth]{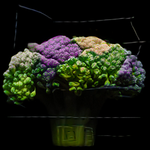}
     \end{subfigure} &
      \begin{subfigure}{0.118\textwidth}\centering
     \caption*{\scriptsize \\ m:0.5,r:0.7,s:0.6}
     \includegraphics[width=1\textwidth]{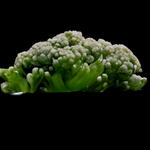}
     \end{subfigure} &
     
      \begin{subfigure}{0.118\textwidth}\centering
     \caption*{\scriptsize broccoli\\ $\rightarrow$head cabbage}
     \includegraphics[width=1\textwidth]{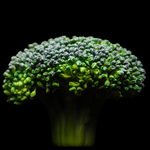}
     \end{subfigure} &
      \begin{subfigure}{0.118\textwidth}\centering
     \caption*{\scriptsize \\ m:1.0,r:0.8,s:0.8}
     \includegraphics[width=1\textwidth]{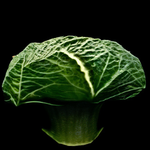}
     \end{subfigure} &
           \begin{subfigure}{0.118\textwidth}\centering
     \caption*{\scriptsize \\ m:0.8,r:0.6,s:0.8}
     \includegraphics[width=1\textwidth]{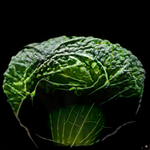}
     \end{subfigure} &
      \begin{subfigure}{0.118\textwidth}\centering
     \caption*{\scriptsize \\ m:0.1,r:0.2,s:0.2}
     \includegraphics[width=1\textwidth]{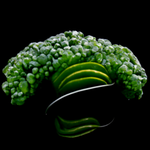}
     \end{subfigure} \\\hline
      \begin{subfigure}{0.118\textwidth}\centering
     \caption*{\scriptsize Petri dish\\ $\rightarrow$soup bowl}
     \includegraphics[width=1\textwidth]{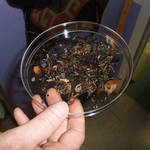}
     \end{subfigure} &
      \begin{subfigure}{0.118\textwidth}\centering
     \caption*{\scriptsize \\ m:0.2,r:0.1,s:0.2}
     \includegraphics[width=1\textwidth]{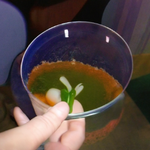}
     \end{subfigure} &
           \begin{subfigure}{0.118\textwidth}\centering
     \caption*{\scriptsize \\ m:0.6,r:0.5,s:0.8}
     \includegraphics[width=1\textwidth]{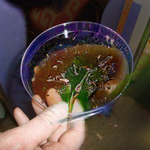}
     \end{subfigure} &
      \begin{subfigure}{0.118\textwidth}\centering
     \caption*{\scriptsize \\ m:0.4,r:0.3,s:0.1}
     \includegraphics[width=1\textwidth]{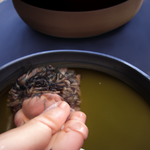}
     \end{subfigure} &
     
      \begin{subfigure}{0.118\textwidth}\centering
     \caption*{\scriptsize meat loaf\\ $\rightarrow$cheeseburger}
     \includegraphics[width=1\textwidth]{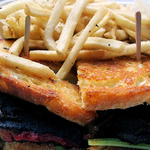}
     \end{subfigure} &
      \begin{subfigure}{0.118\textwidth}\centering
     \caption*{\scriptsize \\ m:0.6,r:0.8,s:0.8}
     \includegraphics[width=1\textwidth]{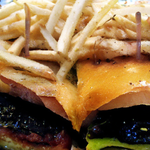}
     \end{subfigure} &
           \begin{subfigure}{0.118\textwidth}\centering
     \caption*{\scriptsize \\ m:0.6,r:0.7,s:0.8}
     \includegraphics[width=1\textwidth]{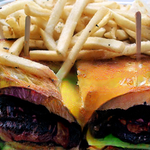}
     \end{subfigure} &
      \begin{subfigure}{0.118\textwidth}\centering
     \caption*{\scriptsize \\ m:0.8,r:0.8,s:0.6}
     \includegraphics[width=1\textwidth]{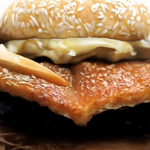}
     \end{subfigure} \\\hline
      \begin{subfigure}{0.118\textwidth}\centering
     \caption*{\scriptsize ptarmigan\\ $\rightarrow$peacock}
     \includegraphics[width=1\textwidth]{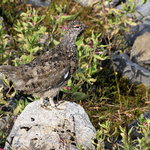}
     \end{subfigure} &
      \begin{subfigure}{0.118\textwidth}\centering
     \caption*{\scriptsize \\ m:1.0,r:0.8,s:0.7}
     \includegraphics[width=1\textwidth]{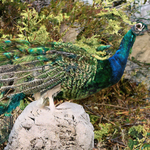}
     \end{subfigure} &
           \begin{subfigure}{0.118\textwidth}\centering
     \caption*{\scriptsize \\ m:0.8,r:0.3,s:0.6}
     \includegraphics[width=1\textwidth]{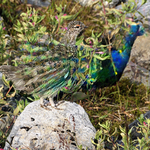}
     \end{subfigure} &
      \begin{subfigure}{0.118\textwidth}\centering
     \caption*{\scriptsize \\ m:0.1,r:0.4,s:0.1}
     \includegraphics[width=1\textwidth]{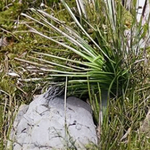}
     \end{subfigure} &
     
      \begin{subfigure}{0.118\textwidth}\centering
     \caption*{\scriptsize bolete\\ $\rightarrow$coral fungus}
     \includegraphics[width=1\textwidth]{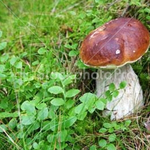}
     \end{subfigure} &
      \begin{subfigure}{0.118\textwidth}\centering
     \caption*{\scriptsize \\ m:0.8,r:0.8,s:0.6}
     \includegraphics[width=1\textwidth]{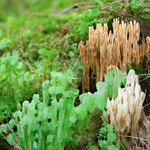}
     \end{subfigure} &
           \begin{subfigure}{0.118\textwidth}\centering
     \caption*{\scriptsize \\ m:0.6,r:0.6,s:0.7}
     \includegraphics[width=1\textwidth]{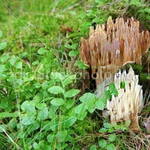}
     \end{subfigure} &
      \begin{subfigure}{0.118\textwidth}\centering
     \caption*{\scriptsize \\ m:0.8,r:0.5,s:0.6}
     \includegraphics[width=1\textwidth]{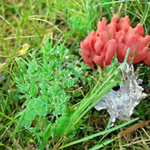}
     \end{subfigure} \\\hline
      \begin{subfigure}{0.118\textwidth}\centering
     \caption*{\scriptsize box turtle\\ $\rightarrow$leather. turtle}
     \includegraphics[width=1\textwidth]{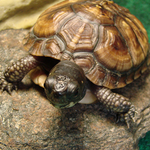}
     \end{subfigure} &
      \begin{subfigure}{0.118\textwidth}\centering
     \caption*{\scriptsize \\ m:0.4,r:0.2,s:0.5}
     \includegraphics[width=1\textwidth]{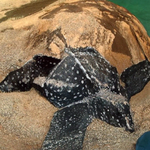}
     \end{subfigure} &
           \begin{subfigure}{0.118\textwidth}\centering
     \caption*{\scriptsize \\ m:0.3,r:0.1,s:0.3}
     \includegraphics[width=1\textwidth]{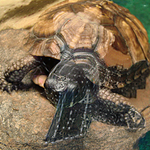}
     \end{subfigure} &
      \begin{subfigure}{0.118\textwidth}\centering
     \caption*{\scriptsize \\ m:0.8,r:0.8,s:0.4}
     \includegraphics[width=1\textwidth]{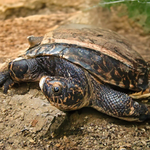}
     \end{subfigure} &
     
      \begin{subfigure}{0.118\textwidth}\centering
     \caption*{\scriptsize Egyptian cat\\ $\rightarrow$tiger cat}
     \includegraphics[width=1\textwidth]{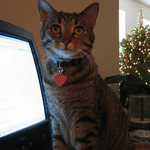}
     \end{subfigure} &
      \begin{subfigure}{0.118\textwidth}\centering
     \caption*{\scriptsize \\ m:0.4,r:0.1,s:0.2}
     \includegraphics[width=1\textwidth]{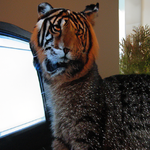}
     \end{subfigure} &
           \begin{subfigure}{0.118\textwidth}\centering
     \caption*{\scriptsize \\ m:0.2,r:0.0,s:0.4}
     \includegraphics[width=1\textwidth]{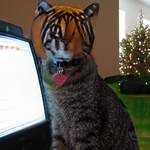}
     \end{subfigure} &
      \begin{subfigure}{0.118\textwidth}\centering
     \caption*{\scriptsize \\ m:0.8,r:0.8,s:0.3}
     \includegraphics[width=1\textwidth]{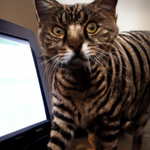}
     \end{subfigure} \\\hline
      \begin{subfigure}{0.118\textwidth}\centering
     \caption*{\scriptsize sea lion\\ $\rightarrow$grey whale}
     \includegraphics[width=1\textwidth]{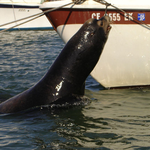}
     \end{subfigure} &
      \begin{subfigure}{0.118\textwidth}\centering
     \caption*{\scriptsize \\ m:0.9,r:0.8,s:0.7}
     \includegraphics[width=1\textwidth]{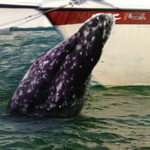}
     \end{subfigure} &
           \begin{subfigure}{0.118\textwidth}\centering
     \caption*{\scriptsize \\ m:0.9,r:0.7,s:0.8}
     \includegraphics[width=1\textwidth]{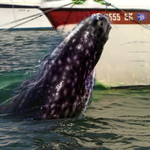}
     \end{subfigure} &
      \begin{subfigure}{0.118\textwidth}\centering
     \caption*{\scriptsize \\ m:0.2,r:0.6,s:0.6}
     \includegraphics[width=1\textwidth]{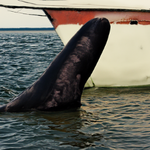}
     \end{subfigure} &
     
      \begin{subfigure}{0.118\textwidth}\centering
     \caption*{\scriptsize kite\\ $\rightarrow$g. grey owl}
     \includegraphics[width=1\textwidth]{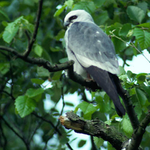}
     \end{subfigure} &
      \begin{subfigure}{0.118\textwidth}\centering
     \caption*{\scriptsize \\ m:1.0,r:0.7,s:0.8}
     \includegraphics[width=1\textwidth]{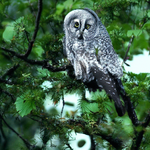}
     \end{subfigure} &
           \begin{subfigure}{0.118\textwidth}\centering
     \caption*{\scriptsize \\ m:0.4,r:0.1,s:0.6}
     \includegraphics[width=1\textwidth]{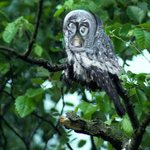}
     \end{subfigure} &
      \begin{subfigure}{0.118\textwidth}\centering
     \caption*{\scriptsize \\ m:0.1,r:0.1,s:0.1}
     \includegraphics[width=1\textwidth]{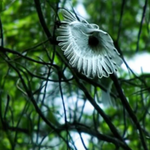}
     \end{subfigure} \\
       \end{tabular}
       \caption{\label{fig:user_study_all}Comparison of different VCE methods for the robust \madryft (taken from \cite{boreiko2022sparse}) for all images used for the user study from App. \ref{app:user_study}. We show our DVCEs (left column), the $l_{1.5}$-SVCEs of \cite{boreiko2022sparse} (middle), and BDVCEs \cite{avrahami2021blended} (right) with the respective percentage of time the respective metric (introduced in App. \ref{app:user_study}) - \textbf{meaningful} (\textbf{m}),  \textbf{realism} (\textbf{r}), \textbf{subtle} (\textbf{s}) - is chosen by the users. Here we show both the target class and the ground truth label for all the VCEs.}
     \end{figure*}

\clearpage

\section{Quantitative evaluation}\label{app:quant-eval}

In this section, we discuss the quantitative evaluation presented in Tab. \ref{tab:quantitative_eval} to complement the user study and the qualitative evaluation of VCEs. The images for DVCEs were generated the same as in App. \ref{app:user_study}. For BDVCEs, because generating images for the FID evaluation is costly, we have chosen one of the settings of the hyperparameters (that achieves high confidence and such that the resulting images look similar to the original ones on average) discussed in Sec. \ref{subsec:method_comp}. For $l_{1.5}$-SVCEs we have selected the highest radius, $r=150$ for the same reasons as discussed in App. \ref{app:user_study}.

\textbf{Realism} is assessed using Fréchet Inception Distance (FID) \cite{FID_score}, which was also used in \cite{boreiko2022sparse}. However, in \cite{boreiko2022sparse}, the authors noticed that there is a risk of having low FID scores for the VCEs that don't change the starting images significantly. To overcome this problem, we propose the following simple ``crossover'' evaluation:
\begin{enumerate}
    \item divide the classes in the WordNet clusters into two disjoint sets A and B with resp. $5504$ and $4576$ images so that one gets a roughly balanced split of  ImageNet classes.
    \item compute VCEs for the test set images of classes A and B with targets ($425$ different classes) in B resp. A ($352
$ different classes) (crossover). More precisely, as targets we use classes from the same WordNet cluster (see the first step) but which are in the other set. This ensures subtle changes of semantically similar classes and rules out meaningless class changes like ``granny smith $\rightarrow$ container ship''.
    \item determine two subsets of the ImageNet trainset for the classes in A resp. B, such that in each subset the distribution of the labels corresponds to the one of A resp. B. Then we compute FID scores once between the training set corresponding to A and the VCEs generated with a target in A and for the training set corresponding to B and the VCEs with targets in B and report the average of the two FID scores.
\end{enumerate}

This way, we make sure that the original images for VCEs don't come from the same distribution as the images from the subset of the training set of ImageNet. As a sanity check, we evaluated the FID scores of the training set of classes in A to the original images of the classes in B and vice versa which yields an average of 41.5. As all methods achieve a smaller FID score (lower is better), they are all able to produce features of the target classes. However, DVCE stands out with an FID score of 17.6 compared to 27.9 (BDVCEs) and  25.6 ($l_{1.5}$-SVCEs).

\textbf{Validity} is evaluated as the mean confidence of the model achieved on the VCEs for the selected target classes. The mean confidence of DVCEs is almost the same as that of $l_{1.5}$-SVCEs which maximize the confidence over the $l_{1.5}$-ball and thus are expected to have high mean confidence. So there is little difference in validity, whereas BDVCEs have significantly lower confidence and thus have worse performance regarding validity. This is again due to the problem that the parameters leading to high confidence of the classifier but still leading to an image related to the original class are extremely difficult to find (if they exist at all) as they are image-specific.

\textbf{Closeness} is assessed with the set of metrics - $l_1, l_{1.5}, l_2,$ as well as the perceptual LPIPS metric \cite{zhang2018unreasonable} for the AlexNet model. As $l_{1.5}$-SVCEs directly manipulate the image without the modification by a generative model, it is not surprising that in terms of $l_p$-distances they outperform DVCEs and BDVCEs. %

\textbf{Summary} From the qualitative and the quantitative results we deduce that DVCEs and $l_{1.5}$-SVCEs are equally valid but DVCEs outperform in terms of image quality (realism) all other approaches significantly while remaining close to the original images.

\newpage
\section{Spurious features}\label{app:spurious-features}
In this section, we want to show how DVCEs can be used as a ``debugging'' tool for ImageNet classifiers (resp. the training set). First, we show that DVCE finds the same spurious features as discovered in \cite{boreiko2022sparse} and we show also more spurious features found by our method.

\begin{figure*}[hbt!]
   \centering
     \small
     \begin{tabular}{ccccc}
     \hline
     \multicolumn{1}{C{0.2\textwidth}}{Original} & \multicolumn{1}{C{0.2\textwidth}}{$l_{1.5}$-SVCEs, \madryft} & 
     \multicolumn{1}{C{0.2\textwidth}}{DVCEs, \madryft} &
     \multicolumn{1}{C{0.2\textwidth}}{\;\;\;\; DVCEs, \newline ConvNeXt} &
     \multicolumn{1}{C{0.2\textwidth}}{\;\;\;\; DVCEs, \newline Swin-TF}\\
     \hline
       
     \begin{subfigure}{0.2\textwidth}\centering
     \caption*{bell pepper}
     \includegraphics[width=1\textwidth]{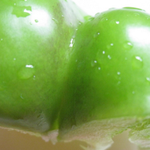}
     \end{subfigure} &
     
     \begin{subfigure}{00.2\textwidth}\centering
     \caption*{$\rightarrow$granny sm.: 1.00}
     \includegraphics[width=1\textwidth]{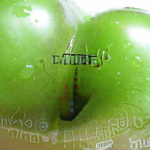}
     \end{subfigure} &

     \begin{subfigure}{00.2\textwidth}\centering
     \caption*{$\rightarrow$granny sm.: 1.00}
     \includegraphics[width=1\textwidth]{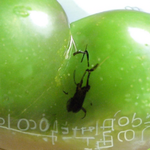}
     \end{subfigure} &

     \begin{subfigure}{00.2\textwidth}\centering
     \caption*{$\rightarrow$granny sm.: 0.99}
     \includegraphics[width=1\textwidth]{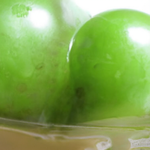}
     
     \end{subfigure}
     
     &

     \begin{subfigure}{00.2\textwidth}\centering
     \caption*{$\rightarrow$granny sm.: 0.99}
     \includegraphics[width=1\textwidth]{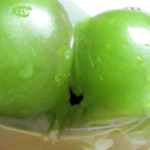}
     
     \end{subfigure}\\

      \begin{subfigure}{00.2\textwidth}\centering
     \caption*{coral reef}
     \includegraphics[width=1\textwidth]{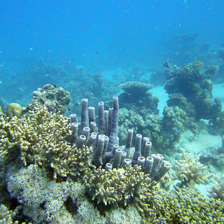}
     \end{subfigure} &
     
     \begin{subfigure}{00.2\textwidth}\centering
     \caption*{$\rightarrow$w. shark: 1.00}
     \includegraphics[width=1\textwidth]{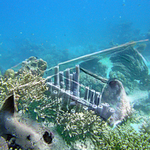}
     \end{subfigure} &

     \begin{subfigure}{00.2\textwidth}\centering
     \caption*{$\rightarrow$w. shark: 1.00}
     \includegraphics[width=1\textwidth]{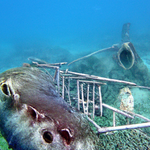}
     \end{subfigure} &

     \begin{subfigure}{00.2\textwidth}\centering
     \caption*{$\rightarrow$w. shark: 0.90}
     \includegraphics[width=1\textwidth]{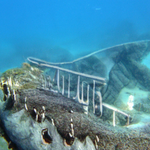}
     \end{subfigure} &

     \begin{subfigure}{00.2\textwidth}\centering
     \caption*{$\rightarrow$w. shark: 0.98}
     \includegraphics[width=1\textwidth]{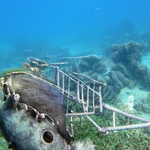}
     \end{subfigure}\\

      \begin{subfigure}{00.2\textwidth}\centering
     \caption*{coral reef}
     \includegraphics[width=1\textwidth]{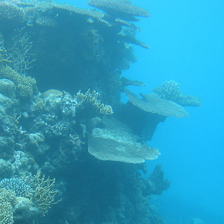}
     \end{subfigure} &
     
     \begin{subfigure}{00.2\textwidth}\centering
     \caption*{$\rightarrow$w. shark: 1.00}
     \includegraphics[width=1\textwidth]{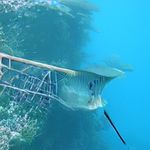}
     \end{subfigure} &

     \begin{subfigure}{00.2\textwidth}\centering
     \caption*{$\rightarrow$w. shark: 1.00}
     \includegraphics[width=1\textwidth]{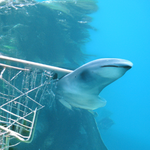}
     \end{subfigure} &

     \begin{subfigure}{00.2\textwidth}\centering
     \caption*{$\rightarrow$w. shark: 0.98}
     \includegraphics[width=1\textwidth]{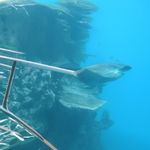}
     \end{subfigure} &

     \begin{subfigure}{00.2\textwidth}\centering
     \caption*{$\rightarrow$w. shark: 0.99}
     \includegraphics[width=1\textwidth]{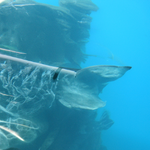}
     \end{subfigure}\\

      \begin{subfigure}{00.2\textwidth}\centering
     \caption*{buckeye}
     \includegraphics[width=1\textwidth]{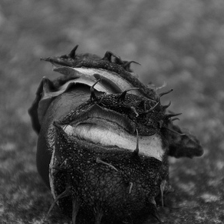}
     \end{subfigure} &
     \begin{subfigure}{00.2\textwidth}\centering
     \caption*{$\rightarrow$tench: 1.00}
     \includegraphics[width=1\textwidth]{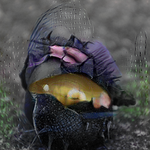}
     \end{subfigure} &

     \begin{subfigure}{00.2\textwidth}\centering
     \caption*{$\rightarrow$tench: 1.00}
    \includegraphics[width=1\textwidth]{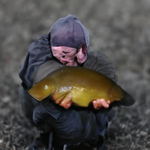}
     \end{subfigure} &

     \begin{subfigure}{00.2\textwidth}\centering
     \caption*{$\rightarrow$tench: 0.98}
     \includegraphics[width=1\textwidth]{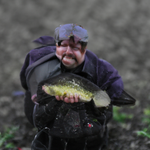}
    \end{subfigure} &

     \begin{subfigure}{00.2\textwidth}\centering
     \caption*{$\rightarrow$tench: 0.98}
     \includegraphics[width=1\textwidth]{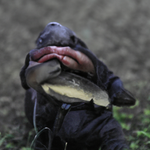}
    \end{subfigure}\\
    
      \begin{subfigure}{00.2\textwidth}\centering
     \caption*{goldfish}
     \includegraphics[width=1\textwidth]{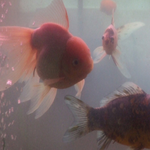}
     \end{subfigure} &
     \begin{subfigure}{00.2\textwidth}\centering
     \caption*{$\rightarrow$tench: 0.99}
     \includegraphics[width=1\textwidth]{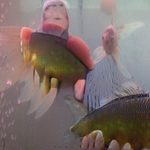}
     \end{subfigure} &

     \begin{subfigure}{00.2\textwidth}\centering
     \caption*{$\rightarrow$tench: 0.99}
    \includegraphics[width=1\textwidth]{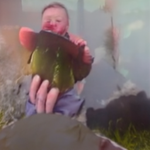}
     \end{subfigure} &

     \begin{subfigure}{00.2\textwidth}\centering
     \caption*{$\rightarrow$tench: 0.97}
     \includegraphics[width=1\textwidth]{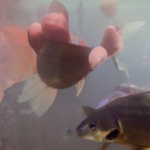}
    \end{subfigure} &

     \begin{subfigure}{00.2\textwidth}\centering
     \caption*{$\rightarrow$tench: 0.97}
     \includegraphics[width=1\textwidth]{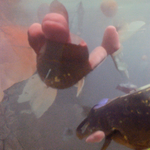}
    \end{subfigure}\\
     \end{tabular}
      \caption{\label{fig:spurious_reproduce_old}\textbf{Reproducing spurious features of \protect\cite{boreiko2022sparse} for the target classes ``granny smith'', ``white shark'', and ``tench''.} $l_{1.5}$-SVCEs from \protect\cite{boreiko2022sparse} for $\epsilon_{1.5}=150$ and DVCEs for \madryft. DVCEs for ConvNeXt and Swin-TF have spurious features only in some cases, see last row, or second row for Swin-TF.
    }
     \end{figure*}

\textbf{Reproducing spurious features found in \cite{boreiko2022sparse}.} In Fig. \ref{fig:spurious_reproduce_old}, we first reproduce the spurious features found by \cite{boreiko2022sparse} with $l_{1.5}$-SVCE for $\epsilon_{1.5}=150$ for the \madryft model. %
We generate the DVCEs for \madryft, the non-robust Swin-TF \cite{liu2021swin}, and ConvNeXt \cite{liu2022convnet}. We find similar spurious features for %
the target classes ``white shark'' and ``tench''. %
As discussed in  \cite{boreiko2022sparse}, these spurious features are a consequence of the selection of the images in the training set. For ``white shark'' it contains a lot of images containing cages to protect the diver and thus the models pick up this ``co-occurrence''. Even more extreme for ``tench'' where a large fraction of images show the angler holding the tench in their hands leading to the fact that counterfactuals for tench contain human faces and hands. 
The spurious text feature for the class ``granny smith'' shown by the \madryft model is not visible for Swin-TF, and ConvNeXt. One key difference is that they have been trained on ImageNet-22k and then fine-tuned to ImageNet which might have reduced this spurious feature but it requires a more detailed analysis to be sure about this.   %

\begin{figure*}[hbt!]
   \centering
     \small
     \begin{tabular}{ccccc}
     \hline
     \multicolumn{1}{C{0.2\textwidth}}{Original} &
     \multicolumn{1}{C{0.2\textwidth}}{DVCEs, \madryft} &
     \multicolumn{1}{C{0.2\textwidth}}{\;\;\;\;\; DVCEs, \newline ConvNeXt} &
     \multicolumn{1}{C{0.2\textwidth}}{\;\;\;\;\;\; DVCEs, \newline Swin-TF} & \multicolumn{1}{C{0.2\textwidth}}{Trainset}\\
     \hline
                   
      \begin{subfigure}{00.2\textwidth}\centering
     \caption*{toyshop}
     \includegraphics[width=1\textwidth]{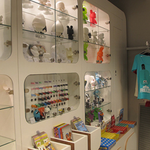}
     \end{subfigure} &
     
     \begin{subfigure}{00.2\textwidth}\centering
     \caption*{$\rightarrow$bee: 1.00}
     \includegraphics[width=1\textwidth]{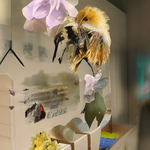}
     \end{subfigure} &
     \begin{subfigure}{00.2\textwidth}\centering
     \caption*{$\rightarrow$bee: 0.95}
     \includegraphics[width=1\textwidth]{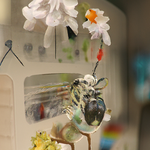}
     \end{subfigure} &

     \begin{subfigure}{00.2\textwidth}\centering
     \caption*{$\rightarrow$bee: 0.51}
     \includegraphics[width=1\textwidth]{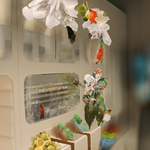}
     \end{subfigure} &

     \begin{subfigure}{00.2\textwidth}\centering
     \caption*{bee}
     \includegraphics[width=1\textwidth]{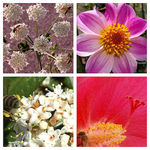}
     \end{subfigure} \\
     
      \begin{subfigure}{00.2\textwidth}\centering
     \caption*{bubble}
     \includegraphics[width=1\textwidth]{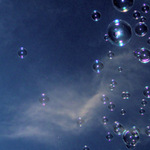}
     \end{subfigure} &
     
     \begin{subfigure}{00.2\textwidth}\centering
     \caption*{$\rightarrow$bee: 0.79}
     \includegraphics[width=1\textwidth]{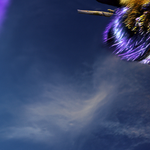}
     \end{subfigure} &

     \begin{subfigure}{00.2\textwidth}\centering
     \caption*{$\rightarrow$bee: 0.64}
     \includegraphics[width=1\textwidth]{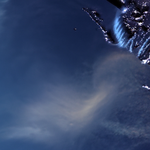}
     \end{subfigure} &

     \begin{subfigure}{00.2\textwidth}\centering
     \caption*{$\rightarrow$bee: 0.99}
     \includegraphics[width=1\textwidth]{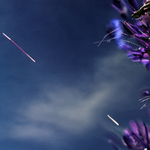}
     \end{subfigure} &

     \begin{subfigure}{00.2\textwidth}\centering
     \caption*{bee}
     \includegraphics[width=1\textwidth]{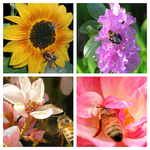}
     \end{subfigure} \\

    \hline
    \hline
      \begin{subfigure}{00.2\textwidth}\centering
     \caption*{tabby}
     \includegraphics[width=1\textwidth]{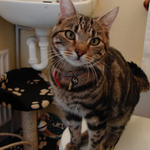}
     \end{subfigure} &
     
     \begin{subfigure}{00.2\textwidth}\centering
     \caption*{$\rightarrow$tiger cat: 1.00}
     \includegraphics[width=1\textwidth]{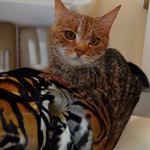}
     \end{subfigure} &

     \begin{subfigure}{00.2\textwidth}\centering
     \caption*{$\rightarrow$tiger cat: 0.97}
     \includegraphics[width=1\textwidth]{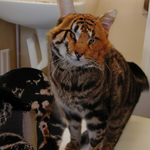}
     \end{subfigure} &

     \begin{subfigure}{00.2\textwidth}\centering
     \caption*{$\rightarrow$tiger cat: 0.98}
    \includegraphics[width=1\textwidth]{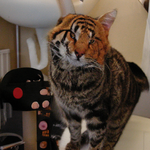}
     \end{subfigure} &

     \begin{subfigure}{00.2\textwidth}\centering
     \caption*{tiger cat}
     \includegraphics[width=1\textwidth]{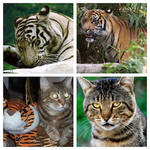}
    \end{subfigure} \\
    
      \begin{subfigure}{00.2\textwidth}\centering
     \caption*{Egyptian cat}
     \includegraphics[width=1\textwidth]{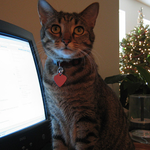}
     \end{subfigure} &
     
     \begin{subfigure}{00.2\textwidth}\centering
     \caption*{$\rightarrow$tiger cat: 0.97}
     \includegraphics[width=1\textwidth]{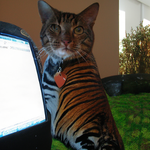}
     \end{subfigure} &
     
     \begin{subfigure}{00.2\textwidth}\centering
     \caption*{$\rightarrow$tiger cat: 0.94}
     \includegraphics[width=1\textwidth]{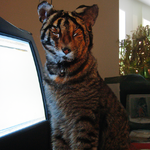}
     \end{subfigure} &

     \begin{subfigure}{00.2\textwidth}\centering
     \caption*{$\rightarrow$tiger cat: 0.95}
    \includegraphics[width=1\textwidth]{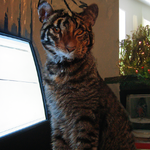}
     \end{subfigure} &

     \begin{subfigure}{00.2\textwidth}\centering
     \caption*{tiger cat}
     \includegraphics[width=1\textwidth]{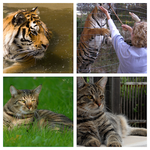}
    \end{subfigure} \\
         \end{tabular}
      \caption{\label{fig:spurious_reproduce_new}\textbf{New spurious features of \cite{boreiko2022sparse} for the target classes ``bee'', and ``tiger cat''.} DVCEs for for ConvNeXt and Swin-TF have spurious features for the respective target classes: flowers (and not the bee itself) for ``bee'', and tiger face for the  ``tiger cat''. Here, we show both the target class and the ground truth label for all the DVCEs.
    }
     \end{figure*}

\textbf{New spurious features for non-robust models.} We use our DVCEs here to find novel spurious features picked up by all classifiers (but to a mixed extent) \madryft, ConvNeXt and Swin-TF: features of flowers (and not the bee itself) increase the confidence in the target class ``bee'' and features of the tiger face increase the confidence in the target class ``tiger cat'' as can be seen in Fig. \ref{fig:spurious_reproduce_new}. Here, we show both the target class and the ground truth label for all the VCEs. We also show in the rightmost column the samples from the trainset of ImageNet from the respective target classes. In both cases, they show why the classifier has picked up these spurious features. Images of bees often show flowers, most of the time much larger than the bee itself. Whereas the DVCE for ``tiger cat'' shows that the training set of this class is completely broken as it contains images of ``tigers'' (note that ``tiger'' is a separate class in ImageNet). 

\clearpage
\section{Failure cases}\label{app:failure-cases}
From what we have seen, generally, there are two failure cases of our method: i) sometimes DVCEs can be blurry, which seems to be an artefact of the diffusion model as it happens for BDVCEs and is visible also in the original diffusion paper (see top left image in Fig. 15 of \cite{dhariwal2021diffusion}), ii) DVCEs can fail when the change is difficult to realize (i.e. the original image is not similar to the target class).

\begin{figure*}[hbt!]
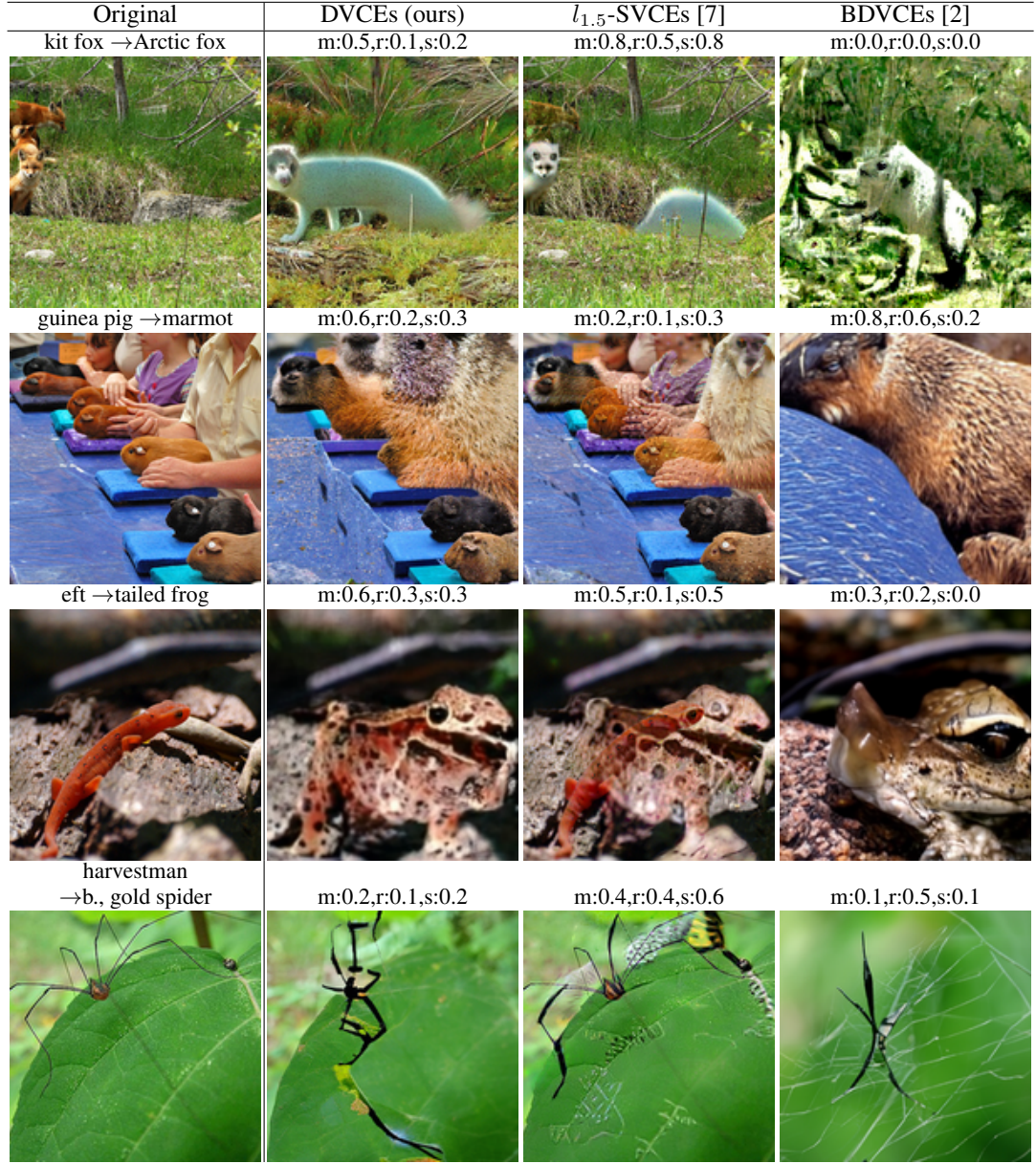
%
     \centering
     \begin{tabular}{c|ccc}
     \hline
     Original &
     DVCEs (ours) &  $l_{1.5}$-SVCEs \cite{boreiko2022sparse}&
       BDVCEs \cite{avrahami2021blended} \\
 \hline
       \begin{subfigure}{0.25\textwidth}\centering
     \caption*{kit fox $\rightarrow$Arctic fox}
     \includegraphics[width=1\textwidth]{images/user_study/55.png}
     \end{subfigure} &
      \begin{subfigure}{0.25\textwidth}\centering
     \caption*{m:0.5,r:0.1,s:0.2}
     \includegraphics[width=1\textwidth]{images/user_study/55_diffusion.png}
     \end{subfigure} &
           \begin{subfigure}{0.25\textwidth}\centering
     \caption*{m:0.8,r:0.5,s:0.8}
     \includegraphics[width=1\textwidth]{images/user_study/55_afw.png}
     \end{subfigure} &
      \begin{subfigure}{0.25\textwidth}\centering
     \caption*{m:0.0,r:0.0,s:0.0}
     \includegraphics[width=1\textwidth]{images/user_study/55_blended.png}
     \end{subfigure} \\
     
     \begin{subfigure}{0.25\textwidth}\centering
     \caption*{guinea pig $\rightarrow$marmot}
     \includegraphics[width=1\textwidth]{images/user_study/30.png}
     \end{subfigure} &
      \begin{subfigure}{0.25\textwidth}\centering
     \caption*{m:0.6,r:0.2,s:0.3}
     \includegraphics[width=1\textwidth]{images/user_study/30_diffusion.png}
     \end{subfigure} &
           \begin{subfigure}{0.25\textwidth}\centering
     \caption*{m:0.2,r:0.1,s:0.3}
     \includegraphics[width=1\textwidth]{images/user_study/30_afw.png}
     \end{subfigure} &
      \begin{subfigure}{0.25\textwidth}\centering
     \caption*{m:0.8,r:0.6,s:0.2}
     \includegraphics[width=1\textwidth]{images/user_study/30_blended.png}
     \end{subfigure} \\
     \begin{subfigure}{0.25\textwidth}\centering
     \caption*{eft $\rightarrow$tailed frog}
     \includegraphics[width=1\textwidth]{images/user_study/16.png}
     \end{subfigure} &
      \begin{subfigure}{0.25\textwidth}\centering
     \caption*{m:0.6,r:0.3,s:0.3}
     \includegraphics[width=1\textwidth]{images/user_study/16_diffusion.png}
     \end{subfigure} &
           \begin{subfigure}{0.25\textwidth}\centering
     \caption*{m:0.5,r:0.1,s:0.5}
     \includegraphics[width=1\textwidth]{images/user_study/16_afw.png}
     \end{subfigure} &
      \begin{subfigure}{0.25\textwidth}\centering
     \caption*{m:0.3,r:0.2,s:0.0}
     \includegraphics[width=1\textwidth]{images/user_study/16_blended.png}
     \end{subfigure}\\
     
     \begin{subfigure}{0.25\textwidth}\centering
     \caption*{harvestman\\ $\rightarrow$b., gold spider}
     \includegraphics[width=1\textwidth]{images/user_study/10.png}
     \end{subfigure} &
      \begin{subfigure}{0.25\textwidth}\centering
     \caption*{\\ m:0.2,r:0.1,s:0.2}
     \includegraphics[width=1\textwidth]{images/user_study/10_diffusion.png}
     \end{subfigure} &
           \begin{subfigure}{0.25\textwidth}\centering
     \caption*{\\ m:0.4,r:0.4,s:0.6}
     \includegraphics[width=1\textwidth]{images/user_study/10_afw.png}
     \end{subfigure} &
      \begin{subfigure}{0.25\textwidth}\centering
     \caption*{\\ m:0.1,r:0.5,s:0.1}
     \includegraphics[width=1\textwidth]{images/user_study/10_blended.png}
     \end{subfigure} \\
       \end{tabular}
       \caption{\label{fig:user_study_failure}\textbf{Some failure cases of DVCEs for the robust \madryft from our user study in App. \ref{app:user_study}}. The failure cases that we have seen are i) DVCEs (and  BDVCEs) can be slightly blurry, or ii) change is difficult to realize. As in Fig. \ref{fig:user_study_all}, we show our DVCEs (left column), the $l_{1.5}$-SVCEs of \protect\cite{boreiko2022sparse} (middle), and  BDVCEs \protect\cite{avrahami2021blended} (right) with the respective percentage of time the respective metric (introduced in App. \ref{app:user_study}) - \textbf{meaningful} (\textbf{m}),  \textbf{realism} (\textbf{r}), \textbf{subtle} (\textbf{s}) - is chosen by the users. Here, we show both the target class and the ground truth label for all the VCEs.}
     \end{figure*}

\clearpage 

\end{document}